\documentclass{article}

 \usepackage[main, final]{neurips_2026}

\usepackage[utf8]{inputenc} 
\usepackage[T1]{fontenc}    
\usepackage{hyperref}       
\usepackage{url}            
\usepackage{booktabs}       
\usepackage{amsfonts}       
\usepackage{nicefrac}       
\usepackage{microtype}      
\usepackage{xcolor}         

\usepackage{float}
\usepackage{amsthm}
\usepackage{amsmath}
\usepackage{amssymb}
\usepackage{caption}
\usepackage{wrapfig}
\usepackage{colortbl}
\usepackage{graphicx}
\usepackage{multirow}
\usepackage{algorithm}
\usepackage{mathtools}
\usepackage{subcaption}
\usepackage{algorithmic}
\definecolor{Gray}{gray}{0.85}

\newcommand{\up}[1]{\textcolor{teal}{\textbf{#1}}}
\newcommand{\down}[1]{\textcolor{gray}{\textbf{#1}}}

\title{CASS: Contribution-Aware Structured Sparsity for Model Merging}

\author{%
Yan Li$^{1}$ \quad
Guiping Cao$^{1}$ \quad
Meng Xu$^{2}$ \quad
Tao Jiang$^{1,3}$ \quad \\
\textbf{Yaguang Song$^{1}$} \quad
\textbf{Ming Tao$^{1}$} \quad
\textbf{Yaowei Wang$^{1,4}$} \quad
\textbf{Dongmei Jiang$^{1,5}$}\thanks{Corresponding author} \\
$^{1}$Pengcheng Laboratory, Shenzhen, China \\
$^{2}$City University of Hong Kong, Hong Kong, China \\
$^{3}$Southern University of Science and Technology, Shenzhen, China \\
$^{4}$Harbin Institute of Technology, Shenzhen, China \\
$^{5}$Northwestern Polytechnical University, Xi'an, China \\
{\small liyan4ai@gmail.com,\; jiangdm@pcl.ac.cn}
}

\begin{document}

\maketitle

\begin{abstract}
Model merging integrates task-specific fine-tuned models into a single multi-task model, but often suffers from parameter interference caused by conflicting task-vector updates. Existing methods typically mitigate conflicts by pruning task vectors based on weight magnitude or random heuristics, treating Transformers as unstructured ``bags of parameters'' and overlooking their inherent modularity. In this paper, we propose \textbf{C}ontribution-\textbf{A}ware \textbf{S}tructured \textbf{S}parsity (CASS), a unified framework that reduces parameter interference by identifying and preserving task-specific components. At the core of CASS is a contribution-aware structured mask that identifies task-relevant attention heads and FFN neurons. We instantiate this mask in two settings: CASS-Merging, the primary post-hoc setting where masks serve as a plug-and-play denoising filter for existing merging operators, and CASS-Tuning, an extension for scenarios with fine-tuning access where masks constrain gradients to reduce structural overlap between task vectors. Our analysis shows that task-relevant components are sparse and partially disjoint, supporting structured component-level filtering as an effective way to reduce merging interference. Extensive experiments across vision (ViT, 20 tasks) and language (RoBERTa, 8 tasks; Qwen2.5, 4 tasks) benchmarks demonstrate that CASS improves a range of representative merging baselines.
\end{abstract}

\section{Introduction}

Fine-tuning a foundation model for each downstream task produces a collection of task-specific expert models, but deploying all experts independently incurs substantial storage and memory costs. Model merging addresses this problem by integrating multiple fine-tuned models into a single multi-task model, typically by combining their task vectors, i.e., the parameter differences between fine-tuned and pre-trained models~\cite{wortsman2022model,ilharco2022editing}. This paradigm enables capability integration without retraining a joint model from scratch and has become a practical tool for reusing task-specific checkpoints~\cite{tang2024fusionbench,he2025mergebench}.

\begin{wrapfigure}{r}{0.4\columnwidth}
    \vspace{-0.1cm}
    \centering
    \includegraphics[width=0.35\columnwidth]{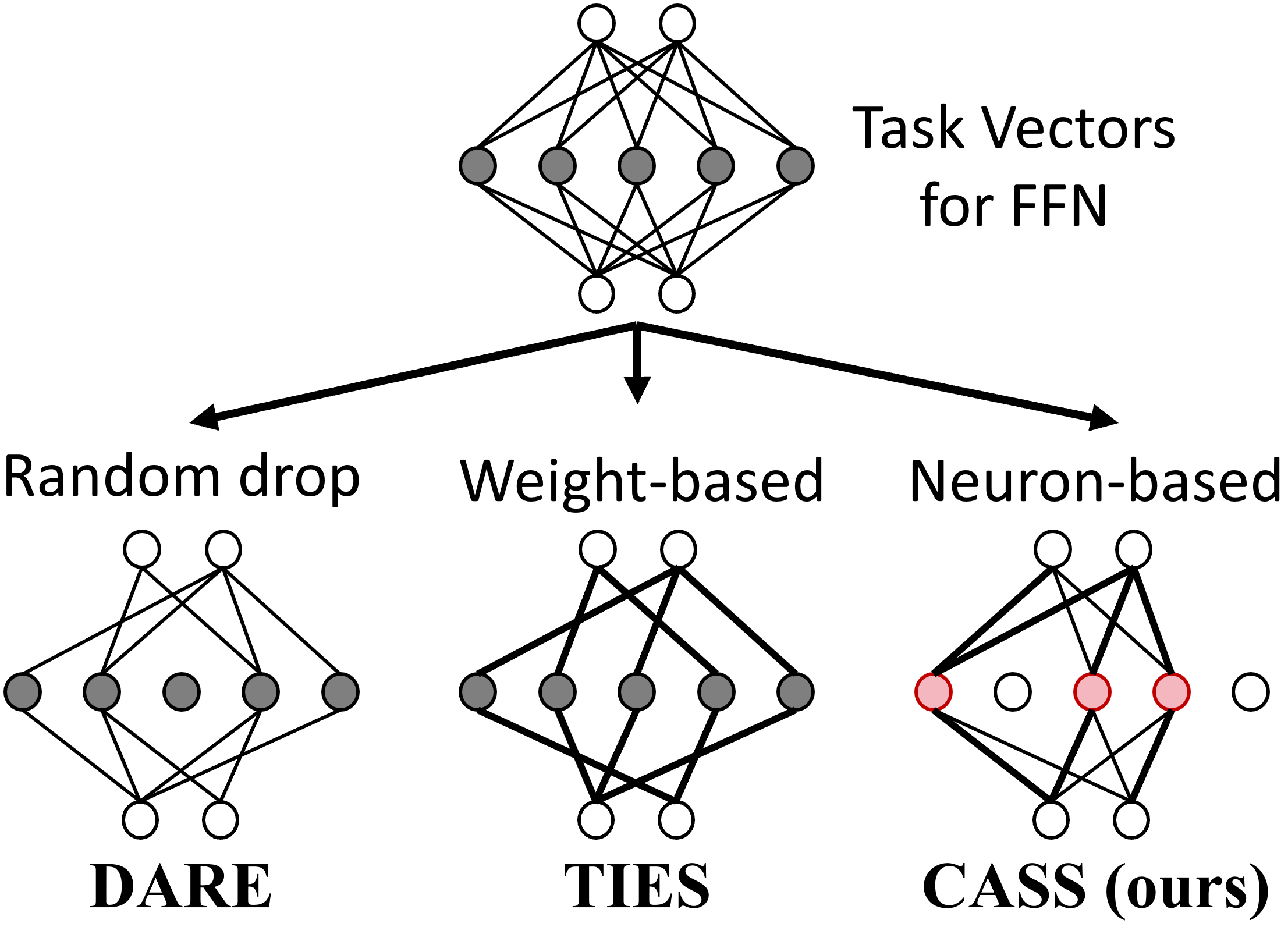}
    \caption{Sparsification-based model merging methods.}
    \label{fig:1motivation}
    \vspace{-0.1cm}
\end{wrapfigure}

Despite its efficiency, model merging remains limited by parameter interference: task vectors obtained from different tasks may contain incompatible updates, and directly aggregating them can degrade the resulting multi-task model. Existing methods mitigate this issue from different perspectives. Optimization-based and subspace-based methods use second-order statistics, inner-product structure, or singular value decomposition to reweight or transform task vectors~\cite{matena2022merging,yang2023adamerging,gargiulo2025task,marczak2025no}. These approaches are often effective but may introduce additional computational or optimization overhead. In contrast, sparsification-based methods improve efficiency by pruning task-vector updates according to magnitude or random criteria~\cite{yadav2023ties,yu2024language,wang2024localizing}. However, such methods usually operate at the individual-parameter level, effectively treating a Transformer as an unstructured ``bag of parameters.'' As illustrated in Fig.~\ref{fig:1motivation}, they do not explicitly account for the fact that Transformer computations are organized into structured functional components such as attention heads and feed-forward neurons.

This motivates a different view of task-vector filtering. In Transformers, both multi-head attention and feed-forward networks can be decomposed into additive contributions to the residual stream. Attention heads perform distinct information-routing functions~\cite{michel2019sixteen,olsson2022context}, while FFN neurons often exhibit sparse and task-dependent activation patterns~\cite{geva2021transformer,xiao2024configurable}. Therefore, task-specific knowledge is unlikely to be uniformly distributed over all parameters; instead, it may be concentrated in a sparse subset of activation-prominent components. From this perspective, the importance of a task-vector update should not be determined solely by its static weight magnitude, but by the functional contribution of the corresponding component under task-specific inputs.

In this paper, we propose \textbf{C}ontribution-\textbf{A}ware \textbf{S}tructured \textbf{S}parsity (CASS), a framework for reducing model-merging interference by filtering task vectors at the level of attention heads and FFN neurons. CASS first estimates the contribution of each component using a small unlabeled task-specific probe set, combining activation statistics with the component's output projection strength. It then constructs structured binary masks that retain task-relevant components and suppress task-irrelevant updates. Importantly, CASS removes only task-specific parameter updates on inactive components, thereby reverting them to the pre-trained model rather than deleting the components themselves.

We use CASS to refer to the general contribution-aware structured masking framework, and distinguish between two instantiations depending on whether fine-tuned task experts are already available. In the standard post-hoc merging setting, CASS-Merging serves as the primary instantiation: it masks each task vector before applying an existing merging operator such as Task Arithmetic, optimization-based merging, or subspace-based merging. When fine-tuning access is available, CASS-Tuning provides a proactive extension that applies the same structured masks to gradients during task-specific fine-tuning, encouraging different task vectors to occupy structurally separated subspaces by construction.

Our contributions are summarized as follows:
\begin{itemize}
    \item We formulate Transformer attention heads and FFN neurons as structured additive contributors to the residual stream, providing an algebraic basis for component-level task-vector filtering.
    \item We propose CASS, a contribution-aware structured sparsity framework that identifies task-relevant heads and neurons using activation-weighted importance metrics and applies the resulting masks to reduce task-vector interference.
    \item We instantiate CASS primarily as CASS-Merging, a post-hoc pre-filter compatible with existing merging operators, and further extend it to CASS-Tuning, a training-time gradient masking strategy that proactively induces structural separation among task vectors.
    \item We conduct extensive experiments across vision encoders (ViT), encoder-only (RoBERTa) and decoder-only (Qwen2.5) language models, spanning 20, 8, and 4 tasks respectively, to demonstrate the effectiveness of the proposed CASS.
\end{itemize}
\section{Related Work}

\subsection{Model Merging}

Model merging aims to integrate the capabilities of multiple task-specific models into a single multi-task model without retraining. We categorize existing methods into four groups. Here, we omit a detailed discussion of MoE-based methods~\cite{lu2024twin,tang2024merging,li2026duetmerging} because the resulting merged model no longer preserves the original model architecture.

\textbf{Linear merging} methods combine task vectors via simple arithmetic operations. Task Arithmetic defines task vectors as the weight difference between fine-tuned and pre-trained models, and aggregates them via scaled summation~\cite{ilharco2022editing}. Model Soups averages the weights of multiple fine-tuned checkpoints directly~\cite{wortsman2022model}.

\textbf{Sparsification-based methods} mitigate interference by pruning conflicting updates. TIES-Merging resolves sign conflicts by retaining only updates with larger magnitudes~\cite{yadav2023ties}. DARE randomly drops parameter updates and rescales the remaining updates~\cite{yu2024language}. PCB-Merging introduces intra-balancing and inter-balancing mechanisms to decide the retention and pruning of parameters~\cite{du2024parameter}. TALL-Masks constructs task-specific binary masks based on weight magnitude differences~\cite{wang2024localizing}. While these methods improve upon simple averaging, they operate at the individual parameter level, treating the model as an unstructured ``bag of parameters'' and ignoring the inherent modular structure of Transformer architectures.

\textbf{Subspace-based methods} exploit the geometric structure of task matrices. For example, TSV-Merging applies singular value decomposition to per-layer task matrices and uses a whitening transformation to decorrelate singular vectors across tasks~\cite{gargiulo2025task}. Iso-C enforces an isotropic singular value spectrum on the merged matrix to improve subspace alignment~\cite{marczak2025no,li2025alignmamba}. These methods aim to identify low-interference directions in the static weight space.

\textbf{Optimization-based methods} resolve interference via iterative optimization~\cite{jiang2026evogm}. Fisher Merging~\cite{matena2022merging} and RegMean~\cite{jin2022dataless} employ second-order information (Fisher matrix or Gram matrix) to perform weighted parameter averaging, but incur significant computational overhead. WUDI-Merging minimizes interference within the linear subspace spanned by task vectors via gradient descent, achieving strong performance without requiring additional data~\cite{cheng2025whoever}. AdaMerging optimizes per-layer merging coefficients via entropy minimization on unlabeled test data, but is limited to classification settings where output entropy is well-defined~\cite{yang2023adamerging}.

In contrast, CASS identifies task-specific components via contribution-based metrics, operating at the structured level of attention heads and FFN neurons rather than individual parameters or weight-space decompositions. This approach is complementary to all four categories above, as demonstrated by our experiments applying CASS as a pre-filter on top of representative methods from each group.

\subsection{Mechanistic Interpretability}

Our work is inspired by the growing field of mechanistic interpretability~\cite{gao2025weight,li2025multimodal,sofroniew2026emotion}, which views Transformers not as monolithic blocks but as collections of specialized functional components. Extensive research suggests that Feed-Forward Networks (FFNs) act as Key-Value memories~\cite{geva2021transformer}, where specific neurons store discrete knowledge patterns. These neurons often exhibit high activation sparsity~\cite{xiao2024configurable}. Similarly, Attention Heads have been shown to perform distinct roles~\cite{michel2019sixteen}, such as induction heads for in-context learning~\cite{olsson2022context} or retrieval heads for factual recall~\cite{wu2024retrieval}. 

CASS leverages these functional specialization properties directly. Unlike previous merging methods that rely on static weights, we identify task-specific components using contribution-based statistics, aligning the merging process with the model's intrinsic modular structure.
\section{Motivation and Analysis}
\label{sec:motivation}

\subsection{Unified Summation Form of Transformer}
\label{sec:summation}

Common merging methods often treat model parameters as unstructured vectors, ignoring the modular architecture of Transformers. To provide an algebraic grounding for structured masking, we reformulate both MHA and FFN into a unified summation form. Consider a vanilla Transformer layer~\cite{vaswani2017attention} with input $\mathbf{x} \in \mathbb{R}^{d}$. We demonstrate that the output of both sub-layers can be decomposed into a sum of independent additive terms.

\begin{figure}[t]
    \centering
    \begin{subfigure}{0.48\columnwidth}
        \centering
        \includegraphics[width=\linewidth]{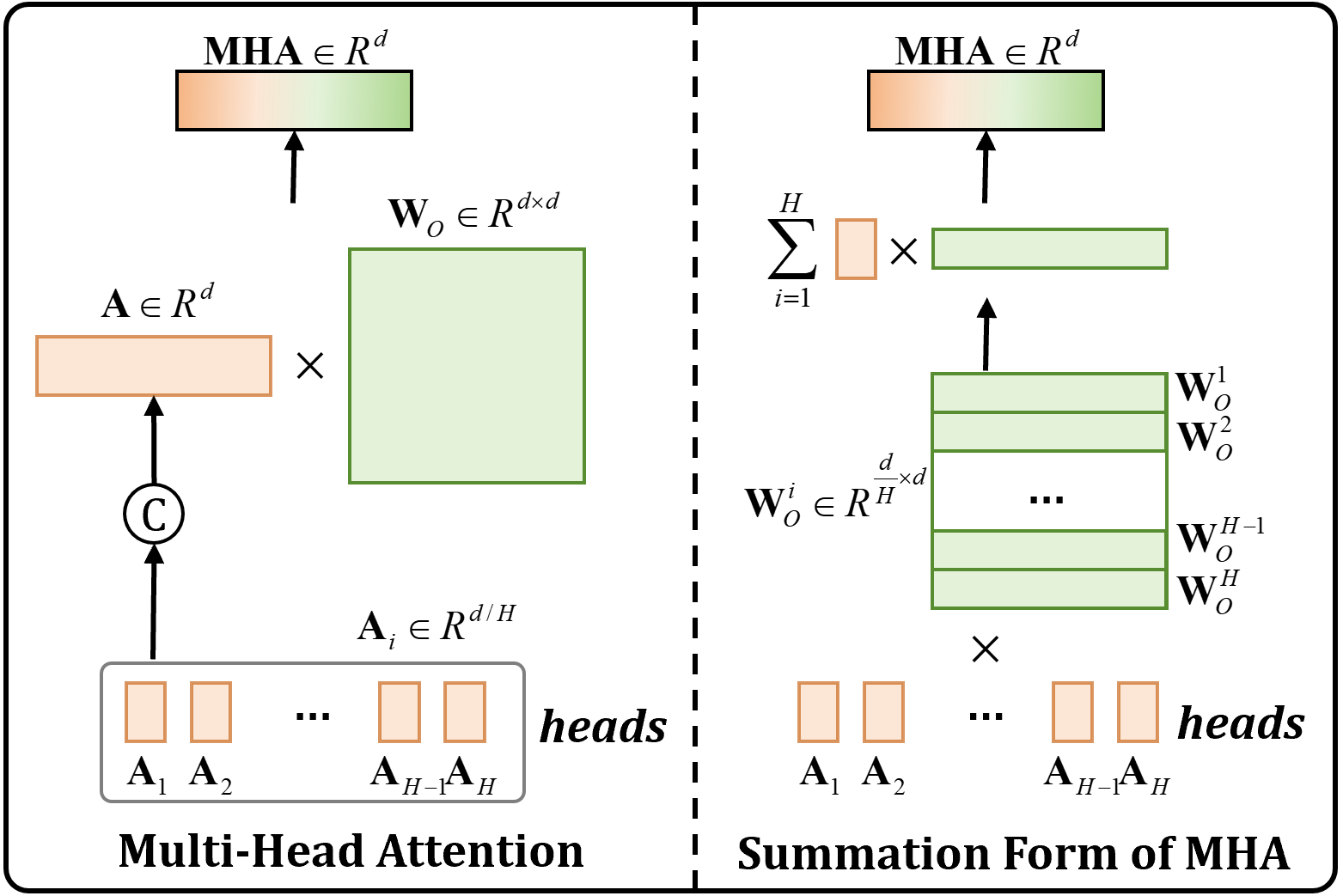}
        \caption{Unified summation form of MHA}
        \label{fig:summation_MHA}
    \end{subfigure}
    \hfill
    \begin{subfigure}{0.48\columnwidth}
        \centering
        \includegraphics[width=\linewidth]{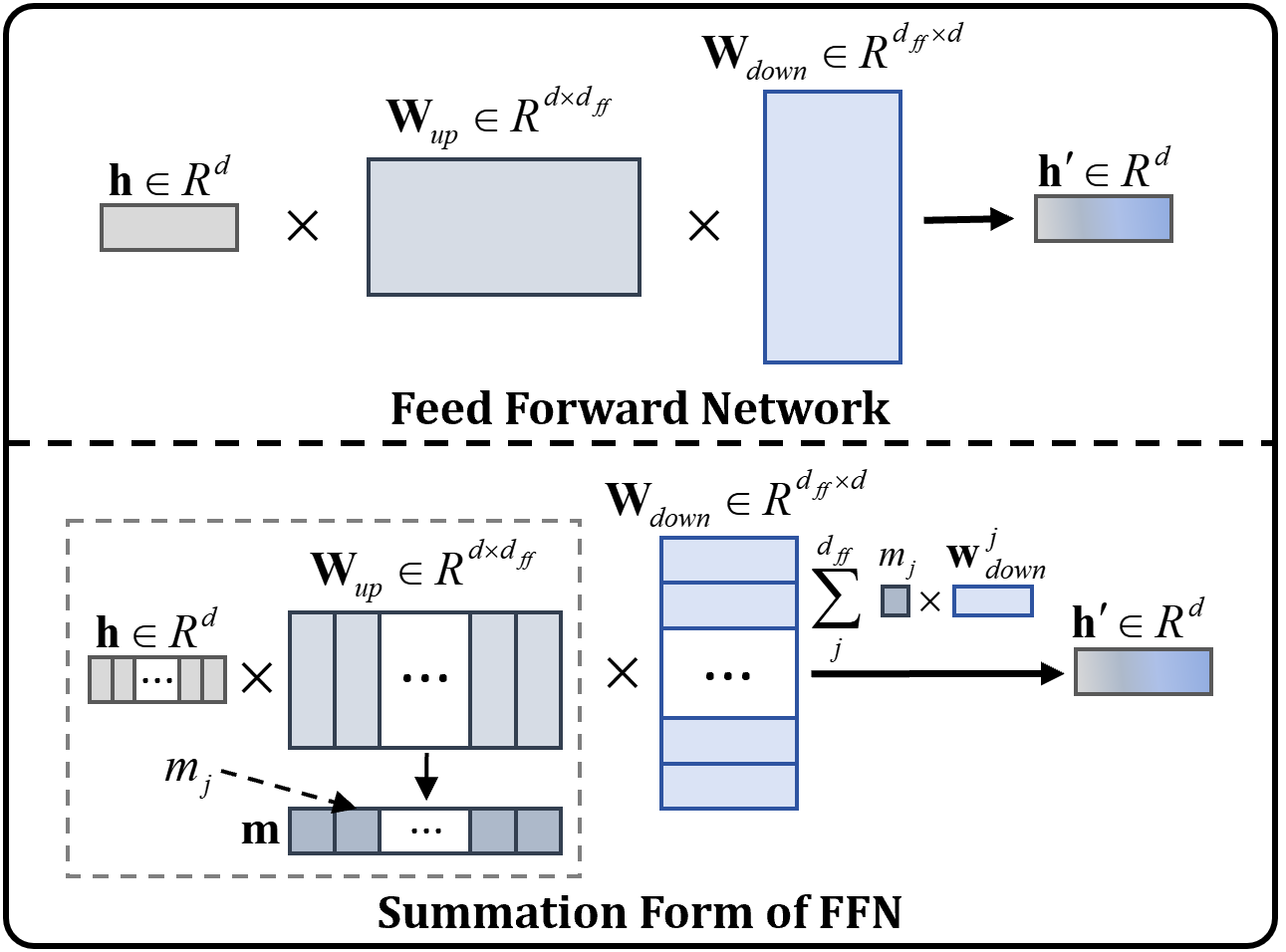}
        \caption{Unified summation form of FFN}
        \label{fig:summation_FFN}
    \end{subfigure}
    \caption{Unified summation form of Transformer.}
    \label{fig:summation_form}
    \vspace{-0.4cm}
\end{figure}

\textbf{MHA as a Sum of Heads.} Let an MHA layer have $H$ heads. By decomposing the output projection $\mathbf{W}_O \in \mathbb{R}^{d \times d}$ into $H$ sub-matrices $\mathbf{W}_O^{(i)} \in \mathbb{R}^{(d/H) \times d}$, as illustrated in Fig.~\ref{fig:summation_form}, the MHA output becomes:
\begin{equation}
    \text{MHA}(\mathbf{x}) = \sum_{i=1}^{H} 
    \mathbf{A}_i(\mathbf{x})\mathbf{W}_O^{(i)}
    \label{eq:mha_sum}
\end{equation}
$\mathbf{A}_i(\mathbf{x})$ is the output of the $i$-th head. Therefore, each term $\mathbf{A}_i(\mathbf{x})\mathbf{W}_O^{(i)}$ is an independent additive update to the residual stream. Masking head $i$ in the task vector is thus algebraically equivalent to zeroing out the task-specific parameter updates associated with that head.

\textbf{FFN as a Sum of Neurons.} Let $\mathbf{m} = \sigma(\mathbf{x}\mathbf{W}_{up}) \in \mathbb{R}^{d_{ff}}$ be the intermediate activation, where $m_j$ is the scalar activation of neuron $j$. Decomposing $\mathbf{W}_{down}$ into row vectors $\mathbf{w}_{down}^{(j)} \in \mathbb{R}^{d}$:
\begin{equation}
    \text{FFN}(\mathbf{x}) = \sum_{j=1}^{d_{ff}} m_j \cdot 
    \mathbf{w}_{down}^{(j)}
    \label{eq:ffn_sum}
\end{equation}
Therefore, each neuron $j$ contributes an additive vector scaled by its activation $m_j$; masking neuron $j$ in the task vector is algebraically equivalent to zeroing out the task-specific updates associated with that intermediate channel.
Detailed derivations are provided in Appendix~\ref{app:derivation}.

\subsection{Importance Metric via Contribution-Awareness}
\label{sec:metric}

Having established the summation form, the challenge lies in identifying which components are critical for a specific task. Traditional sparsification-based model merging methods rely on weight magnitude alone. Here, we propose a \textit{Contribution Score} $\mathcal{I}$ that measures the expected magnitude of \textbf{a component's additive contribution to the residual stream} over a small unlabeled probe set $\mathcal{D}_{\text{probe}}$. For \textbf{FFN neurons}, the absolute contribution magnitude serves as an effective task-specificity proxy:
\begin{equation}
\mathcal{I}_{\text{neuron}}^{(j)} = \mathbb{E}_{\mathbf{x} \sim 
\mathcal{D}_{\text{probe}}} \left[ |m_j| \cdot \left\| 
\mathbf{w}_{down}^{(j)} \right\|_2 \right]
\label{eq:metric}
\end{equation}
For \textbf{attention heads}, however, many heads serve as general-purpose mechanisms active across all tasks, making absolute magnitude an unreliable discriminator. We therefore employ a task-selectivity score that captures relative deviation from the global average contribution (see Appendix~\ref{app:mha_analysis} for details).

\subsection{Empirical Analysis}
\label{sec:analysis_motivation}

To validate our motivation, we analyze the distribution of functional components on \textbf{Qwen2.5-1.5B-Instruct}~\cite{qwen2025qwen25technicalreport} across four tasks: Coding~\cite{wei2023magicoder}, Instruction Following~\cite{lambert2024tulu}, Mathematics~\cite{tong2024dart}, and Safety~\cite{han2024wildguard}.

\begin{wrapfigure}{r}{0.6\columnwidth}
    \centering
    \vspace{-0.2cm}
    \begin{subfigure}{0.29\columnwidth}
        \centering
        \includegraphics[width=\linewidth]{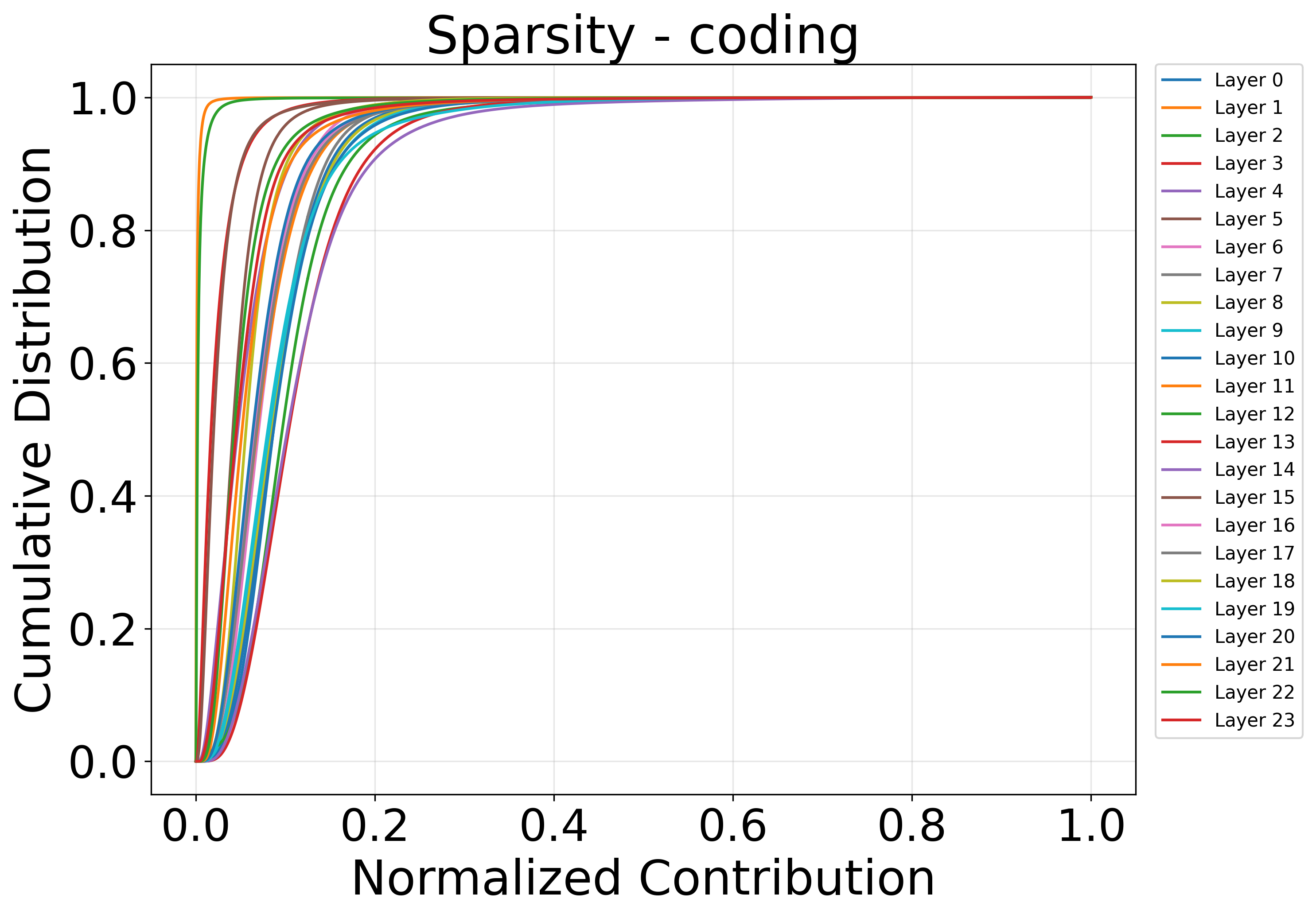}
    \end{subfigure}
    \hfill
    \begin{subfigure}{0.29\columnwidth}
        \centering
        \includegraphics[width=\linewidth]
        {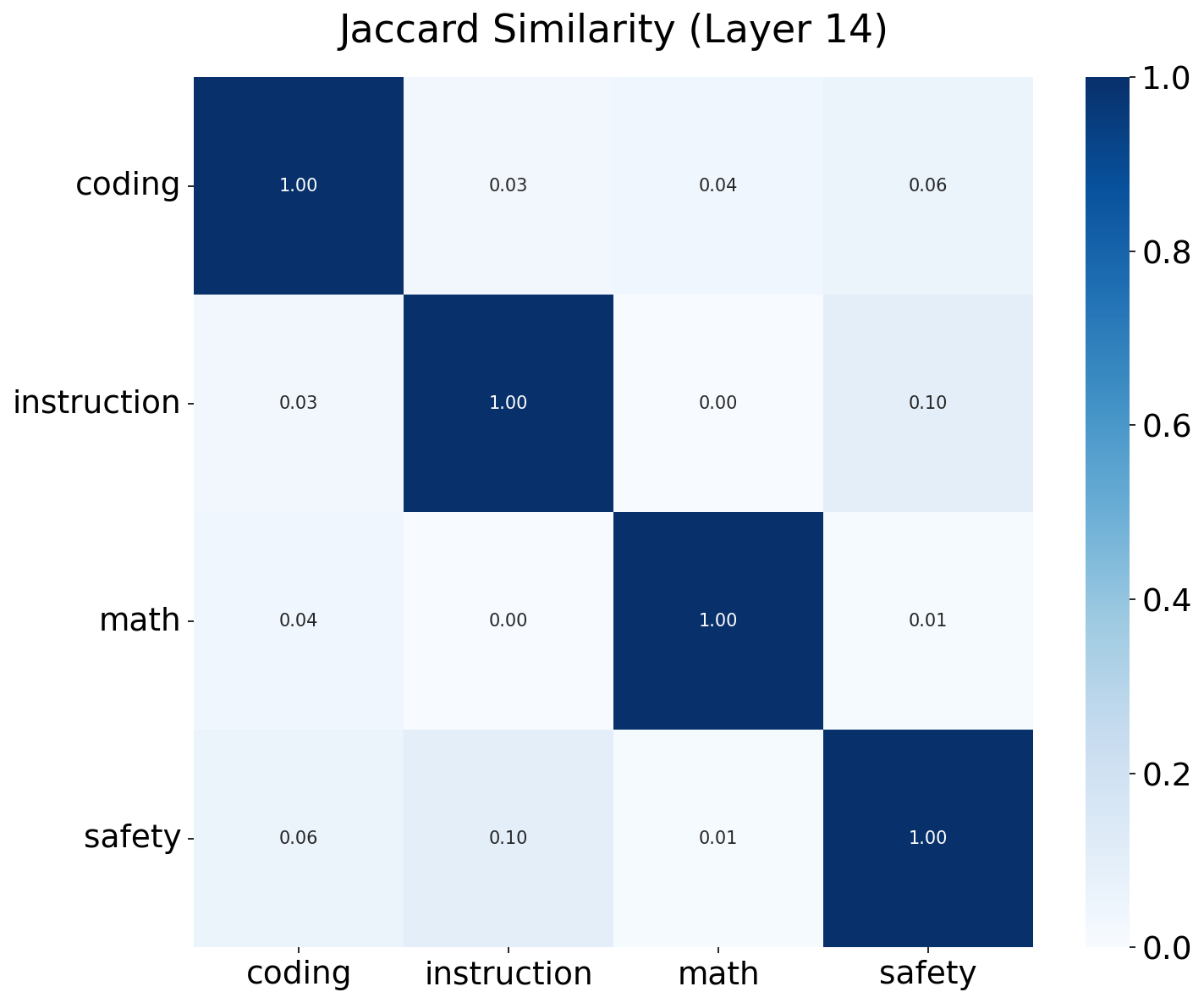}
    \end{subfigure}
    \caption{\textbf{(Left)} CDF of neuron contribution scores (Coding task). \textbf{(Right)} Jaccard similarity of top-20\% neurons across tasks (Layer 14).}
    \label{fig:analysis}
    \vspace{-0.2cm}
\end{wrapfigure}

The left of Fig.~\ref{fig:analysis} plots the Cumulative Distribution Function (CDF) of normalized neuron contribution scores for the Coding task. The sharp rise near zero suggests that the vast majority ($>$80\%) of neurons have negligible contributions to any specific task, which means task-relevant functional activity is concentrated in a small subset of ``loud'' components.

The right of Fig.~\ref{fig:analysis} shows the Jaccard similarity of the top-20\% most active neurons across tasks (Layer 14). The near-zero off-diagonal values reveal a striking pattern: \textit{the highly active components of different tasks are largely disjoint}. This structural non-overlap suggests that parameter interference in standard merging primarily arises from forcing the ``loud'' components of one task to mix with the ``silent'' background of another. By explicitly identifying and preserving these task-specific activation-prominent components via structured masks, our method aims to reduce such conflicts. More analysis on different layers and tasks is provided in Appendix~\ref{app:ffn_analysis} and~\ref{app:mha_analysis}.
\section{Methodology: CASS}
\label{sec:method}

\begin{figure*}[htbp]
    \centering
    \includegraphics[width=\textwidth]{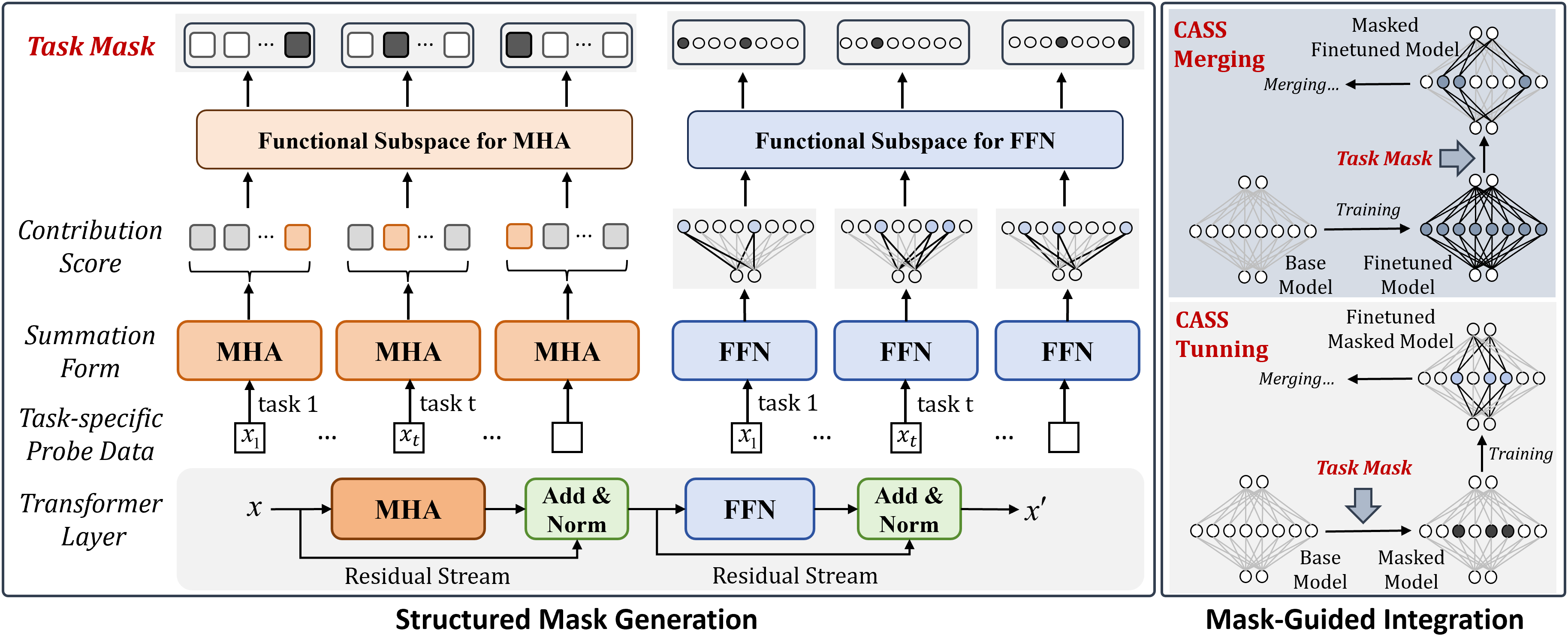}
    \caption{The proposed \textbf{C}ontribution-\textbf{A}ware 
    \textbf{S}tructured \textbf{S}parsity (CASS) framework.}
    \label{fig:4framework}
    \vspace{-0.3cm}
\end{figure*}

Based on the insights in Sec.~\ref{sec:motivation}, we propose \textbf{CASS}, a framework to mitigate model merging interference by identifying and preserving task-specific components. As illustrated in Fig.~\ref{fig:4framework}, CASS consists of a shared mask-generation stage followed by a setting-specific mask-application stage. The shared stage estimates contribution-aware structured masks over attention heads and FFN neurons. The application stage then depends on the available access: in CASS-Merging, masks are applied to task vectors after fine-tuning; in CASS-Tuning, masks are applied to gradients during fine-tuning. This organization emphasizes that both variants share the same structured sparsity principle, while differing only in when the mask is applied.

\subsection{Shared Stage: Structured Mask Generation}
\label{sec:phase1}

CASS operationalizes the contribution scores in Sec.~\ref{sec:metric} by selecting task-relevant Transformer components and translating them into structured binary masks that can be applied to task-vector updates or gradients. For a given task $t$ with probe data $\mathcal{D}_t$, we perform a forward pass on a reference model to compute the importance score $\mathcal{I}_t^{(c)}$ for each component $c$, where $c$ denotes either an attention head or an FFN neuron. The choice of reference model and probe data depends on the application scenario and is detailed in Sec.~\ref{sec:cass_merging} and Sec.~\ref{sec:cass_tuning}, respectively.

Given the set of functional components $\Theta$, we retain the top-$k\%$ components with the highest scores as the task-specific active set. The binary mask is then defined as $\mathbf{M}_t^{(c)}=1$ if component $c$ is selected and $\mathbf{M}_t^{(c)}=0$ otherwise. To reflect the different granularities of attention heads and FFN neurons, we rank attention heads globally across layers, while ranking FFN neurons within each layer. This choice is motivated by the relatively small number of attention heads per layer, for which global ranking yields more stable selection, and the much larger number of FFN neurons, for which layer-wise ranking provides a more balanced and discriminative allocation of active components.

\subsection{Post-hoc Application: CASS-Merging}
\label{sec:cass_merging}

CASS-Merging addresses the standard model merging setting, where $K$ independently fine-tuned expert models $\{\theta_t\}_{t=1}^K$ are given and the goal is to integrate them into a single multi-task model. Standard methods merge task vectors $\tau_t = \theta_t - \theta_{base}$ directly, which accumulate updates from components irrelevant to task $t$, leading to parameter interference.

\textbf{Mask Computation.} In this setting, the fine-tuned checkpoints $\{\theta_t\}$ are available. We therefore compute the structural masks using each fine-tuned model $\theta_t$ as the reference, running a forward pass on unlabeled probe data $\mathcal{D}_t$. This ensures that the identified components faithfully reflect the functional distribution of the fine-tuned model.

\textbf{Task Vector Filtering.} We utilize the structural masks as a functional denoising filter. Before merging, we project each raw task vector onto its identified active subspace:
\begin{equation}
    \tilde{\tau}_t = \mathbf{M}_t \odot \tau_t
    \label{eq:denoising}
\end{equation}
where $\odot$ denotes element-wise structured broadcasting multiplication. This operation zeros out the task vector at inactive component positions, effectively reverting non-essential heads and neurons back to the pre-trained state. The filtered vectors $\tilde{\tau}_t$ are then aggregated using standard Merge Operators (e.g., TIES, TSV, or simple summation):
\begin{equation}
    \theta_{merged} = \theta_{base} + 
    \text{MergeOp}
    \left(
        \{\tilde{\tau}_t\}_{t=1}^{K}
    \right),
    \label{eq:merge}
\end{equation}
By filtering out updates from components that are not prominently activated by task $t$, CASS-Merging reduces the probability of destructive collisions between task-specific updates.

\subsection{Training-time Extension: CASS-Tuning}
\label{sec:cass_tuning}

While CASS-Merging operates \textit{reactively}---filtering interference after fine-tuning has occurred---we further propose \textbf{CASS-Tuning} as a \textit{proactive} extension that mitigates interference at the source by constraining the fine-tuning process itself.

\textbf{Mask Computation.} In this setting, the fine-tuned checkpoints are not yet available. We therefore compute the structural masks using the pre-trained base model $\theta_{base}$ as the reference, running a forward pass on unlabeled training data $\mathcal{D}_t$. This is motivated by the latent capability hypothesis: fine-tuning primarily amplifies functional pathways already present in the pre-trained model, rather than creating entirely new circuits~\cite{aghajanyan2021intrinsic,zhou2023lima}. The base model's activation response to task-specific data is therefore sufficient to identify the relevant subspace.

\textbf{Gradient Masking.} We use the mask $\mathbf{M}_t$ to constrain the optimization trajectory during fine-tuning of task $t$ by masking gradients for inactive components:
\begin{equation}
    \mathbf{g}'_t = \mathbf{M}_t \odot \nabla_\theta \mathcal{L}_t
    \label{eq:grad_mask}
\end{equation}
This ensures that parameter updates are strictly confined to the pre-identified active subspace. Since masks restrict updates to task-relevant components and empirically these active sets tend to have low overlap (as shown in Sec.~\ref{sec:analysis_motivation}), the resulting task vectors are encouraged to occupy more structurally separated regions of the parameter space. This structural separation reduces the need for complex conflict resolution during subsequent merging.
\section{Experiments}
\label{sec:experiments}

\subsection{Experimental Setup}
\label{sec:setup}

\textbf{Datasets and Models.} We evaluate CASS across three benchmarks spanning vision and language modalities. For \textbf{Vision}, we adopt FusionBench~\cite{tang2024fusionbench} with ViT-B/32 and ViT-B/16 architectures pre-trained on CLIP~\cite{radford2021learning}, evaluating on 20 diverse image classification tasks. For \textbf{Language (Encoder-only)}, we evaluate on RoBERTa~\cite{liu2019roberta} across 8 natural language understanding tasks~\cite{wang2018glue}. For \textbf{Language (Decoder-only)}, we follow the MergeBench protocol~\cite{he2025mergebench} with Qwen2.5-0.5B-Instruct and Qwen2.5-1.5B-Instruct~\cite{qwen2025qwen25technicalreport}, covering four capabilities: Instruction Following, Mathematics, Coding, and Safety. For brevity, \textit{Instruct} is omitted from the model names in the subsequent tables. More details are provided in Appendix~\ref{app:settings}.

\textbf{Baselines.} We evaluate CASS as a plug-and-play filter applied on existing model merging methods, spanning four categories: \textit{(1) Linear merging}: Task Arithmetic (TA)~\cite{ilharco2022editing}; \textit{(2) Sparsification-based}: TIES-Merging (TIES)~\cite{yadav2023ties}, DARE~\cite{yu2024language}, PCB-Merging (PCB)~\cite{du2024parameter}; \textit{(3) Subspace-based}: TSV-Merging (TSV)~\cite{gargiulo2025task}, Iso-C~\cite{marczak2025no}; \textit{(4) Optimization-based}: WUDI-Merging (WUDI)~\cite{cheng2025whoever}, AdaMerging (Ada)~\cite{yang2023adamerging}.
AdaMerging is evaluated on vision tasks only, as its entropy-minimization objective is not well-defined for generative and regression tasks.

\subsection{Main Results: CASS-Merging}
\label{sec:exp_merging}

\begin{wraptable}{r}{0.55\columnwidth}
    \centering
    \vspace{-0.3cm}
    \caption{\textbf{Results on Vision Tasks.} Average accuracy (\%) on 20 tasks.}
    \label{tab:vit}
    \setlength{\tabcolsep}{3pt}
    \begin{tabular}{lcccc}
        \toprule
        \multirow{2}{*}{\textbf{Method}} 
        & \multicolumn{2}{c}{\textbf{ViT-B/32}} 
        & \multicolumn{2}{c}{\textbf{ViT-B/16}} \\
        \cmidrule(lr){2-3} \cmidrule(lr){4-5}
        & Base & w/ CASS-M
        & Base & w/ CASS-M \\
        \midrule
        TA
        & 61.37 & 65.29 {\scriptsize\up{(+3.92)}}
        & 65.02 & 69.99 {\scriptsize\up{(+4.97)}} \\
        TIES
        & 63.88 & 65.52 {\scriptsize\up{(+1.64)}}
        & 66.54 & 69.34 {\scriptsize\up{(+2.80)}} \\
        DARE
        & 61.40 & 65.24 {\scriptsize\up{(+3.84)}}
        & 65.03 & 69.90 {\scriptsize\up{(+4.87)}} \\
        PCB
        & 63.72 & 65.75 {\scriptsize\up{(+2.03)}}
        & 67.74 & 69.78 {\scriptsize\up{(+2.04)}} \\
        TSV
        & 73.67 & 74.91 {\scriptsize\up{(+1.24)}}
        & 78.37 & 78.98 {\scriptsize\up{(+0.61)}} \\
        Iso-C
        & 71.61 & 71.88 {\scriptsize\up{(+0.27)}}
        & 76.10 & 76.62 {\scriptsize\up{(+0.52)}} \\
        Ada
        & 68.24 & 69.34 {\scriptsize\up{(+1.10)}}
        & 72.41 & 74.01 {\scriptsize\up{(+1.60)}} \\
        WUDI
        & 72.29 & 73.51 {\scriptsize\up{(+1.22)}}
        & 75.73 & 77.13 {\scriptsize\up{(+1.40)}} \\
        \bottomrule
    \end{tabular}
    \vspace{-0.3cm}
\end{wraptable}

\textbf{Vision Tasks.}
Table~\ref{tab:vit} reports average accuracy across 20 tasks. CASS-Merging consistently improves all eight baselines on both architectures. Specifically, TA and DARE benefit most significantly, with gains up to $+4.97\%$ (TA, ViT-B/16) and $+4.87\%$ (DARE, ViT-B/16), as these methods lack structural awareness and CASS compensates by preserving functionally coherent units. Notably, CASS-Merging also improves strong baselines including subspace-based TSV-Merging ($+1.24\%$ on ViT-B/32) and optimization-based WUDI-Merging ($+1.40\%$ on ViT-B/16), suggesting that activation-based structured sparsity captures functional information complementary to weight-space decomposition. These results indicate that CASS is particularly effective for merge operators that do not explicitly model Transformer component structure, while remaining beneficial when combined with more advanced weight-space conflict-resolution methods.

\begin{table}[htbp]
\vspace{-0.5cm}
    \centering
    \caption{\textbf{Results on RoBERTa and Qwen2.5.} Average performance on 8 tasks for RoBERTa and average normalized performance on 4 tasks for Qwen2.5.}
    \vspace{0.1cm}
    \label{tab:roberta_qwen}
    \setlength{\tabcolsep}{3pt}
    \begin{tabular}{lcc|cccc}
        \toprule
        \multirow{2}{*}{\textbf{Method}}
        & \multicolumn{2}{c|}{\textbf{RoBERTa}}
        & \multicolumn{2}{c}{\textbf{Qwen2.5-0.5B}}
        & \multicolumn{2}{c}{\textbf{Qwen2.5-1.5B}} \\
        \cmidrule(lr){2-3} \cmidrule(lr){4-5} \cmidrule(lr){6-7}
        & Base & w/ CASS-M
        & Base & w/ CASS-M
        & Base & w/ CASS-M \\
        \midrule
        Base Model
        & 47.00 & --
        & 0.720 & --
        & 0.768 & -- \\
        Fine-Tuned
        & 85.91 & --
        & 1.000 & --
        & 1.000 & -- \\
        Post-Mask
        & 79.33 & --
        & 0.910 & --
        & 0.975 & -- \\
        \midrule
        TA
        & 66.36 & 70.12 {\scriptsize\up{(+3.76)}}
        & 0.743 & 0.776 {\scriptsize\up{(+.034)}}
        & 0.789 & 0.837 {\scriptsize\up{(+.048)}} \\
        TIES
        & 70.05 & 71.15 {\scriptsize\up{(+1.10)}}
        & 0.695 & 0.773 {\scriptsize\up{(+.078)}}
        & 0.803 & 0.829 {\scriptsize\up{(+.026)}} \\
        DARE
        & 67.11 & 67.79 {\scriptsize\up{(+0.68)}}
        & 0.797 & 0.807 {\scriptsize\up{(+.010)}}
        & 0.842 & 0.871 {\scriptsize\up{(+.029)}} \\
        PCB
        & 70.28 & 72.26 {\scriptsize\up{(+1.98)}} 
        & 0.693 & 0.745 {\scriptsize\up{(+.052)}}
        & 0.791 & 0.823 {\scriptsize\up{(+.032)}}\\
        TSV
        & 73.38 & 74.19 {\scriptsize\up{(+0.81)}}
        & 0.776 & 0.777 {\scriptsize\up{(+.001)}}
        & 0.831 & 0.847 {\scriptsize\up{(+.016)}} \\
        Iso-C
        & 75.72 & 75.98 {\scriptsize\up{(+0.26)}}
        & 0.738 & 0.729 {\scriptsize\down{(-.009)}}
        & 0.809 & 0.814 {\scriptsize\up{(+.005)}} \\
        WUDI
        & 73.65 & 74.21 {\scriptsize\up{(+0.56)}}
        & 0.717 & 0.776 {\scriptsize\up{(+.059)}}
        & 0.862 & 0.873 {\scriptsize\up{(+.011)}} \\
        \bottomrule
    \end{tabular}
\vspace{-0.2cm}
\end{table}

\textbf{Language Tasks (RoBERTa).}
Table~\ref{tab:roberta_qwen} reports results on the 8-task RoBERTa benchmark. The ``Post-Mask'' row reports the single-task performance of fine-tuned experts after applying CASS masks, serving as a diagnostic reference for how much task-specific capability is preserved by the selected subspace. CASS-Merging yields consistent improvements across all methods, with the largest gains on TA ($+3.76\%$) and PCB-Merging ($+1.98\%$). The complementary gains over the subspace-based method (TSV: $+0.81\%$) and the optimization-based method (WUDI: $+0.56\%$) confirm the cross-architecture generalization of structured activation-based filtering.

\textbf{Language Tasks (Qwen2.5).}
Table~\ref{tab:roberta_qwen} also reports results on the 4-task decoder-only benchmark. The high Post-Mask performance (0.910 / 0.975) suggests that task-specific knowledge is concentrated in a small subset of activation-prominent components, and that the remaining parameters are largely redundant. CASS-Merging improves most baselines across both model scales, with the only exception of Iso-C on Qwen2.5-0.5B. This suggests that CASS is broadly complementary to existing merging operators, while its interaction with spectrum-regularized subspace methods may depend on model scale and hyperparameter choices. Detailed per-task results are provided in Appendix~\ref{app:qwen_detail_merging}.

\subsection{Extension Results: CASS-Tuning}
\label{sec:exp_tuning}

CASS-Tuning constrains the fine-tuning process itself and is evaluated in the decoder-only setting where training access is available. As shown in Table~\ref{tab:qwen_tuning}, the ``Masked FT'' row reports the single-task performance when fine-tuning is restricted to the identified subspace, suggesting that the active components are sufficient to preserve most task-specific capabilities (0.954 / 0.924).

\begin{wraptable}{r}{0.52\columnwidth}
    \centering
    \vspace{-0.3cm}
    \caption{\textbf{Results of CASS-Tuning} on Qwen2.5. Average normalized performance on 4 tasks.}
    \label{tab:qwen_tuning}
    \small
    \setlength{\tabcolsep}{2.5pt}
    \begin{tabular}{lcccc}
        \toprule
        \multirow{2}{*}{\textbf{Method}}
        & \multicolumn{2}{c}{\textbf{Qwen2.5-0.5B}}
        & \multicolumn{2}{c}{\textbf{Qwen2.5-1.5B}} \\
        \cmidrule(lr){2-3} \cmidrule(lr){4-5}
        & Base & w/ CASS-T
        & Base & w/ CASS-T \\
        \midrule
        Base Model
        & 0.720 & --
        & 0.768 & -- \\
        Fine-Tuned
        & 1.000 & --
        & 1.000 & -- \\
        Masked FT
        & 0.954 & --
        & 0.924 & -- \\
        \midrule
        TA
        & 0.743 & 0.811 {\scriptsize\up{(+.069)}}
        & 0.789 & 0.820 {\scriptsize\up{(+.031)}} \\
        TIES
        & 0.695 & 0.816 {\scriptsize\up{(+.121)}}
        & 0.803 & 0.867 {\scriptsize\up{(+.064)}} \\
        DARE
        & 0.797 & 0.810 {\scriptsize\up{(+.013)}}
        & 0.842 & 0.850 {\scriptsize\up{(+.008)}} \\
        PCB 
        & 0.693 & 0.739 {\scriptsize\up{(+.046)}}
        & 0.791 & 0.826 {\scriptsize\up{(+.035)}} \\
        TSV
        & 0.776 & 0.806 {\scriptsize\up{(+.030)}}
        & 0.831 & 0.859 {\scriptsize\up{(+.028)}} \\
        Iso-C
        & 0.738 & 0.757 {\scriptsize\up{(+.019)}}
        & 0.809 & 0.825 {\scriptsize\up{(+.016)}} \\
        WUDI
        & 0.717 & 0.790 {\scriptsize\up{(+.073)}}
        & 0.862 & 0.883 {\scriptsize\up{(+.021)}} \\
        \bottomrule
    \end{tabular}
    \vspace{-0.5cm}
\end{wraptable}

CASS-Tuning consistently improves all baselines, with the most notable gains for TIES ($+0.121$ on Qwen2.5-0.5B, $+0.064$ on Qwen2.5-1.5B). This is because gradient masking ensures any non-zero parameter update is almost task-relevant, converting TIES's heuristic assumption (the parameter magnitude is approximately equivalent to its importance) into a structural property. TA and WUDI-Merging both yield improvements exceeding 0.06 on Qwen2.5-0.5B. For Iso-C on 0.5B, the improvement is modest (+0.019), as the isotropic spectrum adjustment operates in the weight space and its interaction with subspace-constrained fine-tuning is limited at smaller model scales. Detailed per-task results are in Appendix~\ref{app:qwen_detail_tuning}.

\subsection{Analysis}
\label{sec:analysis}

\subsubsection{OOD Generalization}
\label{sec:ood}

Beyond in-distribution task performance, we evaluate whether CASS improves the general capabilities of the fusion model. Table~\ref{tab:ood} reports OOD results on two settings.

\begin{table*}[htbp]
    \vspace{-0.2cm}
    \centering
    \caption{\textbf{OOD Generalization} on ViT and Qwen.}
    \label{tab:ood}
    \begin{minipage}[t]{0.48\textwidth}
        \centering
        \small
        \setlength{\tabcolsep}{4pt}
        \begin{tabular}{lcc}
            \toprule
            \textbf{Method} & \textbf{12-task ID} & \textbf{8-task OOD} \\
            \midrule
            TA (ViT-B/32) & 0.663 & 0.570 \\
            \quad w/ CASS-M & 0.715 \up{(+.052)} & 0.588 \up{(+.018)} \\
            \midrule
            TA (ViT-B/16) & 0.711 & 0.576 \\
            \quad w/ CASS-M & 0.780 \up{(+.069)} & 0.597 \up{(+.021)} \\
            \bottomrule
        \end{tabular}
        \vspace{0.3em}
        \subcaption{Vision OOD (ViT). 12 tasks merged; 
        remaining 8 tasks as OOD.}
        \label{tab:ood_vit}
    \end{minipage}
    \hfill
    \begin{minipage}[t]{0.48\textwidth}
        \centering
        \small
        \setlength{\tabcolsep}{4pt}
        \begin{tabular}{lcc}
            \toprule
            \textbf{Method} & \textbf{MMLU} & $\uparrow$ \\
            \midrule
            TA (Qwen-1.5B)       & 0.4531 & - \\
            \quad w/ CASS-M   & 0.4563 & \up{+0.71\%} \\
            \quad w/ CASS-T    & 0.4603 & \up{+1.59\%}\\
            \midrule
            TA (Qwen-0.5B)      & 0.4251 & - \\
            \quad w/ CASS-M   & 0.4388 & \up{+3.22\%} \\
            \quad w/ CASS-T   & 0.4387 & \up{+3.20\%} \\
            \bottomrule
        \end{tabular}
        \vspace{0.3em}
        \subcaption{Language OOD (Qwen2.5).}
        \label{tab:ood_qwen}
    \end{minipage}
    \vspace{-0.4cm}
\end{table*}

For \textbf{ViT}, we merge 12 in-distribution tasks and evaluate the resulting model on the remaining 8 held-out tasks. CASS-Merging improves both in-distribution and OOD performance on ViT-B/32 and ViT-B/16. This suggests that structured task-vector filtering reduces interference among the merged tasks while better preserving transferable visual representations.

For \textbf{Qwen2.5}, we evaluate the merged model on MMLU~\cite{hendrycks2020measuring}, which measures general knowledge and reasoning ability beyond the four task-specific fine-tuning domains. Both CASS-Merging and CASS-Tuning improve MMLU performance over standard Task Arithmetic on Qwen2.5-0.5B and Qwen2.5-1.5B. The gains are modest but consistent, indicating that restricting task-specific updates to activation-prominent components does not degrade general capability and may help preserve general-purpose knowledge during merging and mask-guided fine-tuning.

\subsubsection{Probe Sample Size Sensitivity}
\label{sec:probe}

\begin{wrapfigure}{r}{0.42\columnwidth}
    \centering
    \vspace{-1.5cm}
    \includegraphics[width=0.4\columnwidth]{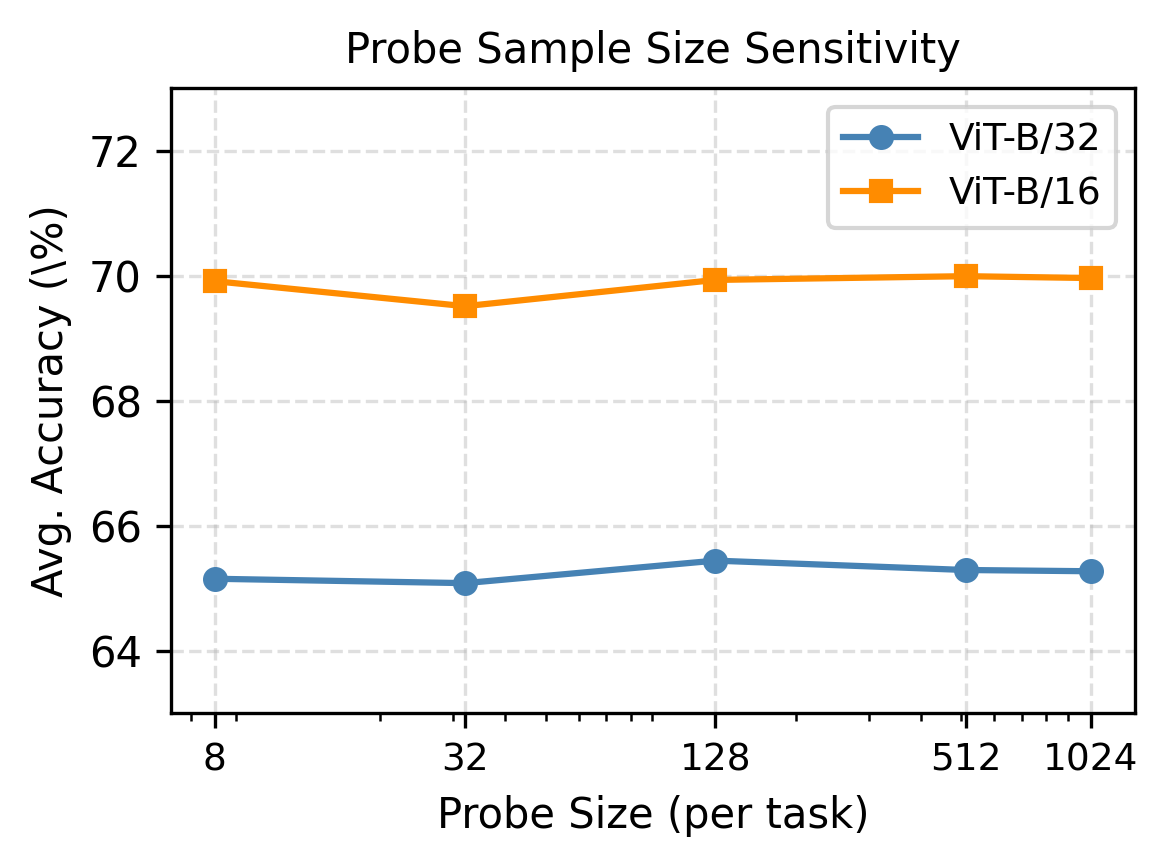}
    \vspace{-0.3cm}
    \caption{Probe sample size sensitivity.}
    \label{fig:probe}
    \vspace{-1cm}
\end{wrapfigure}

CASS relies on a small unlabeled probe set to compute contribution statistics. Fig.~\ref{fig:probe} shows the average accuracy on ViT-B/32 and ViT-B/16 (TA method with CASS-Merging) as the probe size varies from 8 to 1024 samples per task. Performance remains remarkably stable across this range, with less than 0.5\% variation on both architectures. This confirms that CASS is robust to the probe set size and that even a minimal probe of 8 samples per task is sufficient for reliable mask generation.

\textbf{Computational Overhead.} The mask generation step of CASS introduces negligible computational overhead. Taking 8 unlabeled samples per task as an example, computing the activation statistics via a single forward pass and generating the binary masks requires only a few seconds on a single NVIDIA A100 GPU. This one-time cost is amortized over the merging process and is orders of magnitude smaller than the fine-tuning cost. Moreover, since CASS is applied as a preprocessing step, it does not introduce additional memory overhead during the subsequent model merging procedure. Once generated, the masks are reusable across different merging operators (e.g., TA, TIES, DARE) without recomputation. This makes CASS a lightweight and practical pre-filter that can be seamlessly integrated into a broad range of existing merging pipelines with minimal additional cost.

\subsubsection{Ablation Study}
\label{sec:ablation}

\begin{wrapfigure}{r}{0.4\columnwidth}
    \centering
    \vspace{-0.5cm}
    \includegraphics[width=0.38\columnwidth]{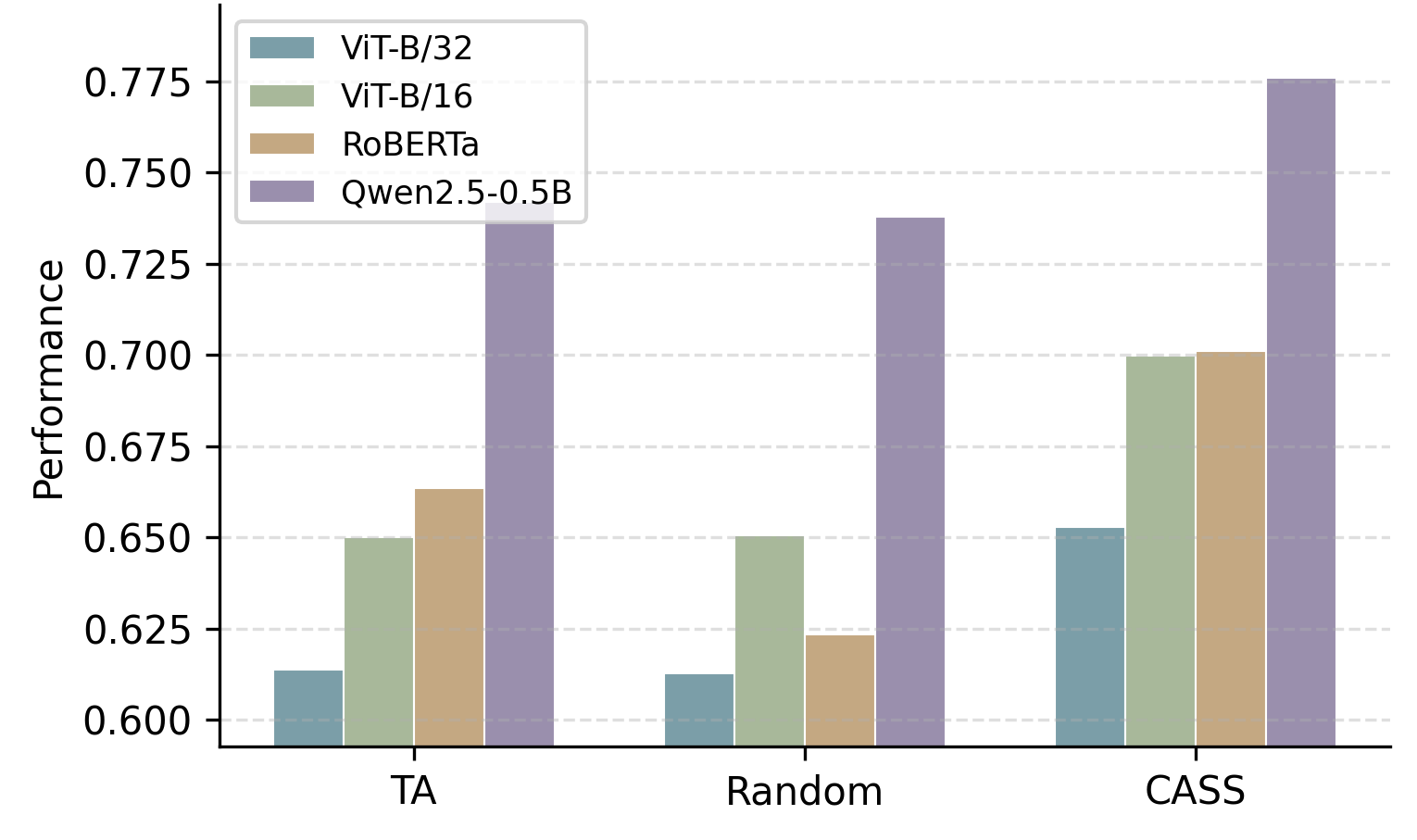}
    \caption{Comparison between TA, random structured masks, and CASS masks. Random masks use the same sparsity budget as CASS.}
    \label{fig:random_vs_cass}
    \vspace{-0.5cm}
\end{wrapfigure}

\paragraph{Contribution-aware masks vs. random masks.}
We first compare CASS with random structured masks under the same masking granularity and sparsity budget. As shown in Fig.~\ref{fig:random_vs_cass}, random masks do not consistently improve over the TA baseline. In particular, on RoBERTa, random masking even leads to a performance drop, indicating that indiscriminately removing task-vector components can discard task-relevant updates and amplify merging instability. In contrast, CASS consistently improves TA across all evaluated benchmarks. This demonstrates that the gains of CASS are not merely caused by reducing the number of merged parameters, but by selectively preserving activation-prominent components that are more likely to carry task-specific functionality. The results support our central hypothesis that contribution-aware structured filtering can reduce inter-task conflicts more effectively than uninformative sparsity.

\begin{wrapfigure}{r}{0.4\columnwidth}
    \centering
    \vspace{-0.5cm}
    \includegraphics[width=0.38\columnwidth]{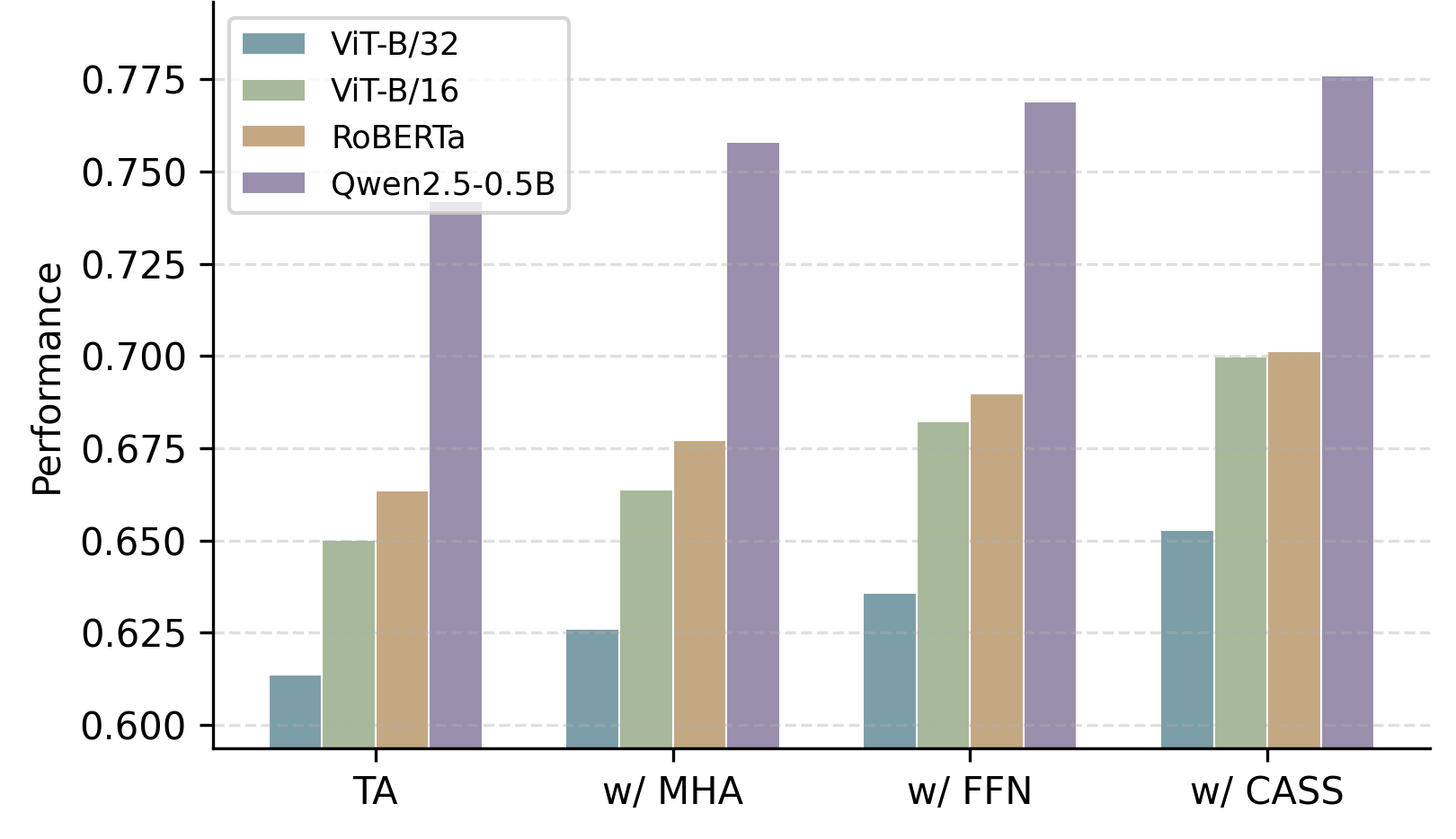}
    \caption{Component-wise ablation study. ``w/ MHA'' and ``w/ FFN'' denote applying CASS masks only to attention heads or FFN neurons, respectively.}
    \label{fig:ablation}
    \vspace{-0.5cm}
\end{wrapfigure}

\paragraph{Component-wise ablation.}
We further isolate the effect of the two types of Transformer components by applying CASS masks only to attention heads (w/ MHA) or only to FFN neurons (w/ FFN). As shown in Fig.~\ref{fig:ablation}, both variants improve over TA, suggesting that task-specific activation patterns exist in both attention and FFN components.

Across the four benchmarks, FFN-only masking generally yields larger gains than MHA-only masking. For example, on ViT-B/32, w/ FFN improves TA by +2.20\%, whereas w/ MHA improves it by +1.24\%. This observation is consistent with the view that FFN neurons often act as sparse key-value memories and therefore contain more task-specific updates. Attention heads, in contrast, tend to serve broader information-routing roles, so head-level filtering provides smaller but complementary improvements. Finally, combining MHA and FFN masks achieves the best performance across all benchmarks. This indicates that attention heads and FFN neurons capture complementary aspects of task-specific functional organization, and that CASS benefits from filtering both types of structured components.
\section{Conclusion}

This paper presents CASS, a framework that mitigates parameter interference in model merging by identifying and preserving task-specific components. Grounded in the algebraic decomposition of Transformer layers, CASS generates structured binary masks via contribution-aware importance metrics, filtering out components irrelevant to each task before merging. CASS operates in two paradigms: CASS-Merging as a lightweight post-hoc filter applicable to any existing merging pipeline, and CASS-Tuning as a proactive training-time constraint designed to inherently reduce structural overlap between task vectors. Extensive experiments show that CASS improves most evaluated baselines across vision and language benchmarks, with particularly consistent gains for sparsification-based and linear merging methods. Limitations are discussed in Appendix~\ref{app:limitations}.

\bibliography{example_paper}
\bibliographystyle{abbrv}


\appendix
\newpage
\clearpage
\begin{center}
    \vspace*{1em}
    \begin{minipage}{0.9\linewidth}
        \hrule height 1.2pt
        \vspace{1em}
        \centering
        {\Large \textbf{Appendix Table of Contents}}
        \vspace{0.8em}
        
        \begin{itemize}
            \setlength{\itemsep}{0.45em}
            \setlength{\leftmargin}{1.5em}
            
            \item \textbf{Appendix~\ref{app:derivation}: Detailed Derivation of the Unified Summation Form} \\
            \small{Provides the detailed derivation of the additive summation forms for MHA and FFN, establishing the algebraic basis for component-level structured masking.}
            
            \item \textbf{Appendix~\ref{app:algorithms}: Algorithms} \\
            \small{Summarizes the complete procedures of CASS-Merging and CASS-Tuning, including mask estimation, task-vector filtering, and mask-guided fine-tuning.}
            
            \item \textbf{Appendix~\ref{app:mask_broadcast}: Mask Construction and Broadcasting} \\
            \small{Specifies how component-level masks over attention heads and FFN neurons are broadcast to parameter-level task-vector updates or gradients.}
            
            \item \textbf{Appendix~\ref{app:ffn_analysis}: Extended Analysis on FFN} \\
            \small{Presents additional analyses of FFN neuron contribution sparsity and cross-task overlap across different task domains and layers.}
            
            \item \textbf{Appendix~\ref{app:mha_analysis}: Extended Analysis on MHA} \\
            \small{Analyzes task-selective attention heads, including selectivity-score distributions, cross-task overlap, and depth-wise localization.}
            
            \item \textbf{Appendix~\ref{app:settings}: Experimental Settings} \\
            \small{Describes the evaluated tasks, models, datasets, metrics, fine-tuning hyperparameters, normalization protocol, and random seed configuration.}
            
            \item \textbf{Appendix~\ref{app:qwen_detail}: Detailed Qwen2.5 Results} \\
            \small{Reports detailed per-capability results for CASS-Merging and CASS-Tuning on Qwen2.5-0.5B-Instruct and Qwen2.5-1.5B-Instruct.}
            
            \item \textbf{Appendix~\ref{app:limitations}: Limitations} \\
            \small{Discusses current limitations of CASS, including fixed sparsity ratios and evaluation scale.}
        \end{itemize}
        
        \vspace{0.8em}
        \hrule height 1.2pt
    \end{minipage}
\end{center}
\clearpage

\section{Detailed Derivation of the Unified Summation Form}
\label{app:derivation}

In Section~\ref{sec:summation}, we introduced a unified summation view of Transformer sub-layers to motivate structured component-level filtering. Here, we provide a more detailed derivation for Multi-Head Attention (MHA) and Feed-Forward Networks (FFNs), explicitly accounting for tensor dimensions and bias terms.

This appendix has two purposes. First, we show that, under a fixed set of model parameters, attention heads and FFN neurons contribute additive terms to the residual stream. This provides the basis for defining contribution-based component scores. Second, we clarify how CASS applies the resulting component masks to task-specific parameter updates. Importantly, CASS does not perform output-level component deletion. Instead, it masks task-vector updates or gradients at the parameter level, so that inactive components retain their pre-trained parameters.

For clarity, we present the derivation using a sequence input $\mathbf{X} \in \mathbb{R}^{L \times d}$, where $L$ is the sequence length and $d$ is the hidden dimension. The single-token formulation used in the main text can be viewed as the special case $L=1$.

\subsection{Multi-Head Attention}
\label{app:derivation_mha}

Consider an MHA layer with $H$ attention heads and head dimension $d_h=d/H$. For the $i$-th head, the query, key, and value projections are
$\mathbf{W}_Q^{(i)}, \mathbf{W}_K^{(i)}, \mathbf{W}_V^{(i)} \in \mathbb{R}^{d \times d_h}$.
The output of head $i$ is given by
\begin{equation}
    \mathbf{H}_i(\mathbf{X})
    =
    \mathrm{Attn}
    \left(
    \mathbf{X}\mathbf{W}_Q^{(i)},
    \mathbf{X}\mathbf{W}_K^{(i)},
    \mathbf{X}\mathbf{W}_V^{(i)}
    \right)
    \in \mathbb{R}^{L \times d_h}.
\end{equation}

The standard MHA output is obtained by concatenating all head outputs and applying the output projection:
\begin{equation}
    \mathrm{MHA}(\mathbf{X})
    =
    \mathrm{Concat}
    \left(
    \mathbf{H}_1(\mathbf{X}), \ldots, \mathbf{H}_H(\mathbf{X})
    \right)
    \mathbf{W}_O
    +
    \mathbf{b}_O,
\end{equation}
where $\mathbf{W}_O \in \mathbb{R}^{d \times d}$ and $\mathbf{b}_O \in \mathbb{R}^{d}$ is broadcast along the sequence dimension.

To expose the additive structure, we decompose the output projection matrix $\mathbf{W}_O$ into $H$ row blocks:
\begin{equation}
    \mathbf{W}_O
    =
    \begin{bmatrix}
        \mathbf{W}_O^{(1)} \\
        \vdots \\
        \mathbf{W}_O^{(H)}
    \end{bmatrix},
    \quad
    \mathbf{W}_O^{(i)} \in \mathbb{R}^{d_h \times d}.
\end{equation}
Then, by block matrix multiplication,
\begin{equation}
\begin{aligned}
    &\mathrm{Concat}
    \left(
    \mathbf{H}_1(\mathbf{X}), \ldots, \mathbf{H}_H(\mathbf{X})
    \right)
    \mathbf{W}_O \\
    &=
    \begin{bmatrix}
        \mathbf{H}_1(\mathbf{X}) & \cdots & \mathbf{H}_H(\mathbf{X})
    \end{bmatrix}
    \begin{bmatrix}
        \mathbf{W}_O^{(1)} \\
        \vdots \\
        \mathbf{W}_O^{(H)}
    \end{bmatrix} \\
    &=
    \sum_{i=1}^{H}
    \mathbf{H}_i(\mathbf{X}) \mathbf{W}_O^{(i)}.
\end{aligned}
\end{equation}
Therefore, for a fixed model, the MHA output can be written as
\begin{equation}
    \mathrm{MHA}(\mathbf{X})
    =
    \sum_{i=1}^{H}
    \mathbf{H}_i(\mathbf{X}) \mathbf{W}_O^{(i)}
    +
    \mathbf{b}_O.
    \label{eq:app_mha_sum}
\end{equation}

Equation~\eqref{eq:app_mha_sum} shows that each attention head contributes an additive term
$\mathbf{H}_i(\mathbf{X})\mathbf{W}_O^{(i)}$ to the residual stream under a fixed parameterization. CASS uses this additive term to estimate the functional contribution of a head. For example, the head-level contribution can be measured by
\begin{equation}
    \mathcal{A}_{i,t}
    =
    \mathbb{E}_{\mathbf{X}\sim\mathcal{D}_t}
    \left[
    \left\|
    \mathbf{H}_i(\mathbf{X})\mathbf{W}_O^{(i)}
    \right\|_F
    \right],
    \label{eq:app_mha_contribution}
\end{equation}
where the activations and weights are taken from the reference model used for mask estimation.

\paragraph{Parameter-level masking in CASS.}
The above additive decomposition should not be confused with output-level component deletion. In CASS, a head mask is applied to task-specific parameter updates, not to the final head output. Let $\theta_0$ denote the base model and $\theta_t=\theta_0+\tau_t$ denote the fine-tuned expert for task $t$. For head $i$, the task-vector updates are
\begin{equation}
\begin{aligned}
    \Delta \mathbf{W}_{Q,t}^{(i)} &= \mathbf{W}_{Q,t}^{(i)} - \mathbf{W}_{Q,0}^{(i)}, \\
    \Delta \mathbf{W}_{K,t}^{(i)} &= \mathbf{W}_{K,t}^{(i)} - \mathbf{W}_{K,0}^{(i)}, \\
    \Delta \mathbf{W}_{V,t}^{(i)} &= \mathbf{W}_{V,t}^{(i)} - \mathbf{W}_{V,0}^{(i)}, \\
    \Delta \mathbf{W}_{O,t}^{(i)} &= \mathbf{W}_{O,t}^{(i)} - \mathbf{W}_{O,0}^{(i)}.
\end{aligned}
\end{equation}
Given a binary head mask $M_{\mathrm{head}}^{(i,t)}\in\{0,1\}$, CASS constructs the masked task-specific parameters as
\begin{equation}
\begin{aligned}
    \widetilde{\mathbf{W}}_{Q,t}^{(i)}
    &=
    \mathbf{W}_{Q,0}^{(i)}
    +
    M_{\mathrm{head}}^{(i,t)}
    \Delta \mathbf{W}_{Q,t}^{(i)}, \\
    \widetilde{\mathbf{W}}_{K,t}^{(i)}
    &=
    \mathbf{W}_{K,0}^{(i)}
    +
    M_{\mathrm{head}}^{(i,t)}
    \Delta \mathbf{W}_{K,t}^{(i)}, \\
    \widetilde{\mathbf{W}}_{V,t}^{(i)}
    &=
    \mathbf{W}_{V,0}^{(i)}
    +
    M_{\mathrm{head}}^{(i,t)}
    \Delta \mathbf{W}_{V,t}^{(i)}, \\
    \widetilde{\mathbf{W}}_{O,t}^{(i)}
    &=
    \mathbf{W}_{O,0}^{(i)}
    +
    M_{\mathrm{head}}^{(i,t)}
    \Delta \mathbf{W}_{O,t}^{(i)}.
\end{aligned}
\label{eq:app_mha_param_mask}
\end{equation}
The resulting head output is then computed using these masked parameters:
\begin{equation}
    \widetilde{\mathbf{H}}_{i,t}(\mathbf{X})
    =
    \mathrm{Attn}
    \left(
    \mathbf{X}\widetilde{\mathbf{W}}_{Q,t}^{(i)},
    \mathbf{X}\widetilde{\mathbf{W}}_{K,t}^{(i)},
    \mathbf{X}\widetilde{\mathbf{W}}_{V,t}^{(i)}
    \right).
\end{equation}
Thus, the CASS-filtered MHA layer is
\begin{equation}
    \widetilde{\mathrm{MHA}}_t(\mathbf{X})
    =
    \sum_{i=1}^{H}
    \widetilde{\mathbf{H}}_{i,t}(\mathbf{X})
    \widetilde{\mathbf{W}}_{O,t}^{(i)}
    +
    \widetilde{\mathbf{b}}_{O,t}.
    \label{eq:app_mha_cass_param}
\end{equation}

This formulation is the exact parameter-level operation used by CASS. When $M_{\mathrm{head}}^{(i,t)}=0$, the task-specific update of head $i$ is removed, and the head reverts to its pre-trained parameters. When $M_{\mathrm{head}}^{(i,t)}=1$, the task-specific update of the head is retained. Because attention is nonlinear in the query and key projections, Eq.~\eqref{eq:app_mha_cass_param} is not algebraically equivalent to multiplying the fine-tuned head output by a scalar mask. The mask is applied to parameter updates rather than to the already-computed output contribution.

\subsection{Feed-Forward Networks}
\label{app:derivation_ffn}

We next consider a standard FFN layer with an up-projection
$\mathbf{W}_{up} \in \mathbb{R}^{d \times d_{ff}}$, a down-projection
$\mathbf{W}_{down} \in \mathbb{R}^{d_{ff} \times d}$, and biases
$\mathbf{b}_{up} \in \mathbb{R}^{d_{ff}}$ and
$\mathbf{b}_{down} \in \mathbb{R}^{d}$. Given sequence input
$\mathbf{X} \in \mathbb{R}^{L \times d}$, the intermediate activation is
\begin{equation}
    \mathbf{M}
    =
    \sigma
    \left(
    \mathbf{X}\mathbf{W}_{up} + \mathbf{b}_{up}
    \right)
    \in \mathbb{R}^{L \times d_{ff}},
\end{equation}
where $\sigma(\cdot)$ denotes the element-wise nonlinearity. The FFN output is
\begin{equation}
    \mathrm{FFN}(\mathbf{X})
    =
    \mathbf{M}\mathbf{W}_{down}
    +
    \mathbf{b}_{down}.
\end{equation}

Let $\mathbf{m}_j \in \mathbb{R}^{L}$ denote the $j$-th column of $\mathbf{M}$, i.e., the activation of neuron $j$ across all sequence positions. Let
$\mathbf{w}_{down}^{(j)} \in \mathbb{R}^{d}$ denote the $j$-th row of $\mathbf{W}_{down}$. Then the matrix product can be decomposed as
\begin{equation}
    \mathbf{M}\mathbf{W}_{down}
    =
    \sum_{j=1}^{d_{ff}}
    \mathbf{m}_j
    \mathbf{w}_{down}^{(j)},
    \label{eq:app_ffn_sum}
\end{equation}
where each term $\mathbf{m}_j\mathbf{w}_{down}^{(j)} \in \mathbb{R}^{L \times d}$ is the additive residual-stream contribution of neuron $j$ across the sequence.

Therefore, for a fixed model, the FFN output can be written as
\begin{equation}
    \mathrm{FFN}(\mathbf{X})
    =
    \sum_{j=1}^{d_{ff}}
    \mathbf{m}_j
    \mathbf{w}_{down}^{(j)}
    +
    \mathbf{b}_{down}.
    \label{eq:app_ffn_fixed_sum}
\end{equation}

Equation~\eqref{eq:app_ffn_fixed_sum} motivates the FFN neuron contribution score used by CASS:
\begin{equation}
    \mathcal{I}_{\mathrm{neuron}}^{(j,t)}
    =
    \mathbb{E}_{\mathbf{X}\sim\mathcal{D}_t}
    \left[
    \left\|
    \mathbf{m}_j(\mathbf{X})
    \mathbf{w}_{down}^{(j)}
    \right\|_F
    \right].
    \label{eq:app_ffn_contribution}
\end{equation}
In practice, this can be computed using an equivalent activation-weighted output-norm form.

\paragraph{Parameter-level masking in CASS.}
As with attention heads, CASS does not delete FFN neurons from the model. Instead, it masks the task-specific updates associated with the corresponding intermediate channel. Let the task-vector updates for neuron $j$ be
\begin{equation}
\begin{aligned}
    \Delta \mathbf{W}_{up,t}[:,j]
    &=
    \mathbf{W}_{up,t}[:,j]
    -
    \mathbf{W}_{up,0}[:,j],
    \\
    \Delta \mathbf{W}_{down,t}[j,:]
    &=
    \mathbf{W}_{down,t}[j,:]
    -
    \mathbf{W}_{down,0}[j,:].
\end{aligned}
\end{equation}
Given a binary neuron mask $M_{\mathrm{neuron}}^{(j,t)}\in\{0,1\}$, CASS constructs
\begin{equation}
\begin{aligned}
    \widetilde{\mathbf{W}}_{up,t}[:,j]
    &=
    \mathbf{W}_{up,0}[:,j]
    +
    M_{\mathrm{neuron}}^{(j,t)}
    \Delta \mathbf{W}_{up,t}[:,j],
    \\
    \widetilde{\mathbf{W}}_{down,t}[j,:]
    &=
    \mathbf{W}_{down,0}[j,:]
    +
    M_{\mathrm{neuron}}^{(j,t)}
    \Delta \mathbf{W}_{down,t}[j,:].
\end{aligned}
\label{eq:app_ffn_param_mask}
\end{equation}
If the up-projection bias is present and updated, the corresponding bias channel is masked in the same way:
\begin{equation}
    \widetilde{\mathbf{b}}_{up,t}[j]
    =
    \mathbf{b}_{up,0}[j]
    +
    M_{\mathrm{neuron}}^{(j,t)}
    \Delta \mathbf{b}_{up,t}[j].
\end{equation}

The resulting CASS-filtered intermediate activation is
\begin{equation}
    \widetilde{\mathbf{m}}_{j,t}(\mathbf{X})
    =
    \sigma
    \left(
    \mathbf{X}
    \widetilde{\mathbf{W}}_{up,t}[:,j]
    +
    \widetilde{\mathbf{b}}_{up,t}[j]
    \right),
\end{equation}
and the CASS-filtered FFN output is
\begin{equation}
    \widetilde{\mathrm{FFN}}_t(\mathbf{X})
    =
    \sum_{j=1}^{d_{ff}}
    \widetilde{\mathbf{m}}_{j,t}(\mathbf{X})
    \widetilde{\mathbf{W}}_{down,t}[j,:]
    +
    \widetilde{\mathbf{b}}_{down,t}.
    \label{eq:app_ffn_cass_param}
\end{equation}

When $M_{\mathrm{neuron}}^{(j,t)}=0$, neuron $j$ is not removed. Instead, its task-specific update is suppressed, and the neuron uses its pre-trained up- and down-projection parameters. When $M_{\mathrm{neuron}}^{(j,t)}=1$, the task-specific update associated with that neuron is retained. Because the activation $\widetilde{\mathbf{m}}_{j,t}(\mathbf{X})$ depends nonlinearly on the masked up-projection parameters, this operation cannot in general be written as a scalar multiplication of the fine-tuned neuron's final contribution. The mask is applied to the parameter update that defines the neuron, not to the output contribution after the forward pass.

\paragraph{Summary.}
The summation forms in Eq.~\eqref{eq:app_mha_sum} and Eq.~\eqref{eq:app_ffn_fixed_sum} are used to define contribution scores and identify task-relevant components under a fixed reference model. The actual CASS operation is parameter-level task-vector filtering, as shown in Eq.~\eqref{eq:app_mha_param_mask} and Eq.~\eqref{eq:app_ffn_param_mask}. Thus, inactive components retain their pre-trained behavior rather than being deleted from the network.

\clearpage
\section{Algorithms}
\label{app:algorithms}

In this section, we provide the detailed procedures of CASS-Merging and CASS-Tuning. One implementation detail is worth noting. FFN neuron scores are computed independently for each task, while attention head scores use a task-selectivity criterion that depends on the global average contribution of each head across the task set. Therefore, head-mask construction is performed after collecting head contribution statistics from all tasks. This introduces a task-set-level dependency, which is natural in the standard model merging setting where all task experts or task datasets are available before merging. The global average is computed only from unlabeled probe activations and does not use task labels.

\subsection{CASS-Merging}
\label{app:alg_cass_merging}

Algorithm~\ref{alg:cass_merging} summarizes the CASS-Merging procedure. The algorithm first collects component statistics for all tasks. It then computes the global head contribution baseline and constructs task-specific masks. This two-stage structure makes explicit that attention head selectivity is defined relative to the current set of tasks being merged.

\begin{algorithm}[h]
\caption{CASS-Merging}
\label{alg:cass_merging}
\begin{algorithmic}[1]
\REQUIRE Base model $\theta_{base}$; fine-tuned experts $\{\theta_t\}_{t=1}^{K}$; unlabeled probe sets $\{\mathcal{D}_t\}_{t=1}^{K}$; retention ratios $k_{\mathrm{head}}$ and $k_{\mathrm{ffn}}$; merge operator $\mathrm{MergeOp}(\cdot)$.
\ENSURE Merged model $\theta_{merged}$.

\STATE \textbf{Stage 1: Collect task-specific component statistics.}

\FOR{each task $t = 1,\ldots,K$}
    \STATE Compute the task vector:
    \[
        \tau_t = \theta_t - \theta_{base}.
    \]
    \STATE Use the fine-tuned expert $\theta_t$ as the reference model.
    \STATE Run a forward pass on $\mathcal{D}_t$ and collect component activations.
    \STATE Compute FFN neuron contribution scores
    $\mathcal{I}_{\mathrm{neuron}}^{(\ell,j,t)}$ for each layer $\ell$ and neuron $j$.
    \STATE Compute absolute attention head contributions
    $\mathcal{A}_{h,t}$ for each attention head $h$.
\ENDFOR

\STATE Compute the global average contribution of each attention head:
\[
    \mu_{h,\mathrm{global}}
    =
    \frac{1}{K}
    \sum_{t=1}^{K}
    \mathcal{A}_{h,t}.
\]

\STATE \textbf{Stage 2: Construct masks and filter task vectors.}

\FOR{each task $t = 1,\ldots,K$}
    \STATE Compute task-selectivity scores for attention heads:
    \[
        \mathcal{S}_{h,t}
        =
        \frac{
        \mathcal{A}_{h,t}
        -
        \mu_{h,\mathrm{global}}
        }{
        \mu_{h,\mathrm{global}} + \epsilon
        }.
    \]
    \STATE Select the top-$k_{\mathrm{ffn}}$ FFN neurons within each layer according to $\mathcal{I}_{\mathrm{neuron}}^{(\ell,j,t)}$.
    \STATE Select the top-$k_{\mathrm{head}}$ attention heads across the model according to positive task-selectivity scores $\mathcal{S}_{h,t}$.
    \STATE Construct the structured binary mask $\mathbf{M}_t$ over selected attention heads and FFN neurons.
    \STATE Broadcast $\mathbf{M}_t$ to the corresponding parameter slices of $\tau_t$.
    \STATE Filter the task vector:
    \[
        \tilde{\tau}_t = \mathbf{M}_t \odot \tau_t.
    \]
\ENDFOR

\STATE Merge the filtered task vectors:
\[
    \theta_{merged}
    =
    \theta_{base}
    +
    \mathrm{MergeOp}
    \left(
    \{\tilde{\tau}_t\}_{t=1}^{K}
    \right).
\]

\RETURN $\theta_{merged}$.
\end{algorithmic}
\end{algorithm}

\subsection{CASS-Tuning}
\label{app:alg_cass_tuning}

Algorithm~\ref{alg:cass_tuning} summarizes the CASS-Tuning procedure. Since fine-tuned experts are not yet available, masks are estimated using the base model before task-specific fine-tuning. As in CASS-Merging, attention head selectivity is computed relative to the global average contribution across the task set.

\begin{algorithm}[h]
\caption{CASS-Tuning}
\label{alg:cass_tuning}
\begin{algorithmic}[1]
\REQUIRE Base model $\theta_{base}$; task training sets $\{\mathcal{D}_t\}_{t=1}^{K}$; retention ratios $k_{\mathrm{head}}$ and $k_{\mathrm{ffn}}$; learning rate $\eta$; number of training steps $S$; merge operator $\mathrm{MergeOp}(\cdot)$.
\ENSURE Merged model $\theta_{merged}$.

\STATE \textbf{Stage 1: Estimate masks from the base model.}

\FOR{each task $t = 1,\ldots,K$}
    \STATE Use the base model $\theta_{base}$ as the reference model.
    \STATE Run a forward pass on unlabeled samples from $\mathcal{D}_t$ and collect component activations.
    \STATE Compute FFN neuron contribution scores
    $\mathcal{I}_{\mathrm{neuron}}^{(\ell,j,t)}$ for each layer $\ell$ and neuron $j$.
    \STATE Compute absolute attention head contributions
    $\mathcal{A}_{h,t}$ for each attention head $h$.
\ENDFOR

\STATE Compute the global average contribution of each attention head:
\[
    \mu_{h,\mathrm{global}}
    =
    \frac{1}{K}
    \sum_{t=1}^{K}
    \mathcal{A}_{h,t}.
\]

\FOR{each task $t = 1,\ldots,K$}
    \STATE Compute task-selectivity scores for attention heads:
    \[
        \mathcal{S}_{h,t}
        =
        \frac{
        \mathcal{A}_{h,t}
        -
        \mu_{h,\mathrm{global}}
        }{
        \mu_{h,\mathrm{global}} + \epsilon
        }.
    \]
    \STATE Select the top-$k_{\mathrm{ffn}}$ FFN neurons within each layer according to $\mathcal{I}_{\mathrm{neuron}}^{(\ell,j,t)}$.
    \STATE Select the top-$k_{\mathrm{head}}$ attention heads according to task-selectivity scores $\mathcal{S}_{h,t}$.
    \STATE Construct the structured binary mask $\mathbf{M}_t$.
\ENDFOR

\STATE \textbf{Stage 2: Fine-tune each task under its structured mask.}

\FOR{each task $t = 1,\ldots,K$}
    \STATE Initialize the task-specific model:
    \[
        \theta_t^{(0)} = \theta_{base}.
    \]
    \FOR{training step $s = 0,\ldots,S-1$}
        \STATE Sample a mini-batch $\mathcal{B}_t \subset \mathcal{D}_t$.
        \STATE Compute the gradient and apply gradient masking:
        \[
            \tilde{\mathbf{g}}_t^{(s)}
            =
            \mathbf{M}_t
            \odot
            \mathbf{g}_t^{(s)}.
        \]
        \STATE Update the active subspace:
        \[
            \theta_t^{(s+1)}
            =
            \theta_t^{(s)}
            -
            \eta
            \tilde{\mathbf{g}}_t^{(s)}.
        \]
    \ENDFOR
    \STATE Obtain the CASS-tuned task vector:
    \[
        \tilde{\tau}_t
        =
        \theta_t^{(S)}
        -
        \theta_{base}.
    \]
\ENDFOR

\STATE Merge the CASS-tuned task vectors:
\[
    \theta_{merged}
    =
    \theta_{base}
    +
    \mathrm{MergeOp}
    \left(
    \{\tilde{\tau}_t\}_{t=1}^{K}
    \right).
\]

\RETURN $\theta_{merged}$.
\end{algorithmic}
\end{algorithm}

\clearpage
\section{Mask Construction and Broadcasting}
\label{app:mask_broadcast}

CASS defines binary masks at the level of functional components, i.e., attention heads and FFN neurons. However, both task-vector filtering in CASS-Merging and gradient masking in CASS-Tuning are implemented at the parameter level. This section specifies how component-level masks are broadcast to the corresponding parameter slices.

Throughout this section, let $\Delta \mathbf{W}$ denote either a task-vector update in CASS-Merging or a gradient/update tensor in CASS-Tuning. For CASS-Merging, $\Delta \mathbf{W}$ corresponds to the parameter difference between a fine-tuned expert and the base model. For CASS-Tuning, $\Delta \mathbf{W}$ corresponds to the gradient or optimizer update at a given training step. In both cases, a zero mask suppresses only the task-specific update associated with a component. It does not remove the component from the pre-trained model. Therefore, inactive components retain their pre-trained parameters after merging or masked fine-tuning.

\subsection{Attention Head Masks}
\label{app:head_mask_broadcast}

Consider a Transformer layer $\ell$ with $H$ attention heads and head dimension $d_h=d/H$. For head $i$, CASS assigns a binary mask
$M_{\mathrm{head}}^{(\ell,i)} \in \{0,1\}$.
The corresponding projection matrices are
\begin{equation}
    \mathbf{W}_Q^{(\ell,i)},
    \mathbf{W}_K^{(\ell,i)},
    \mathbf{W}_V^{(\ell,i)}
    \in \mathbb{R}^{d \times d_h},
    \quad
    \mathbf{W}_O^{(\ell,i)}
    \in \mathbb{R}^{d_h \times d},
\end{equation}
where $\mathbf{W}_O^{(\ell,i)}$ is the output-projection block associated with head $i$.

The head mask is broadcast to all parameter slices associated with this head:
\begin{equation}
\begin{aligned}
    \Delta \widetilde{\mathbf{W}}_Q^{(\ell,i)}
    &=
    M_{\mathrm{head}}^{(\ell,i)}
    \cdot
    \Delta \mathbf{W}_Q^{(\ell,i)}, \\
    \Delta \widetilde{\mathbf{W}}_K^{(\ell,i)}
    &=
    M_{\mathrm{head}}^{(\ell,i)}
    \cdot
    \Delta \mathbf{W}_K^{(\ell,i)}, \\
    \Delta \widetilde{\mathbf{W}}_V^{(\ell,i)}
    &=
    M_{\mathrm{head}}^{(\ell,i)}
    \cdot
    \Delta \mathbf{W}_V^{(\ell,i)}, \\
    \Delta \widetilde{\mathbf{W}}_O^{(\ell,i)}
    &=
    M_{\mathrm{head}}^{(\ell,i)}
    \cdot
    \Delta \mathbf{W}_O^{(\ell,i)}.
\end{aligned}
\label{eq:head_broadcast}
\end{equation}

If the attention projections contain bias terms, the same mask is applied to the head-specific slices of the query, key, and value biases. The output-projection bias is not decomposed by head and is therefore left unmasked unless otherwise specified.

This broadcasting rule ensures that when a head is inactive for task $t$, CASS suppresses the task-specific update associated with the entire attention head. In CASS-Merging, the corresponding task-vector slices are zeroed out before applying the merge operator. In CASS-Tuning, the corresponding gradient slices are zeroed out during fine-tuning.

\subsection{FFN Neuron Masks}
\label{app:ffn_mask_broadcast}

We next consider a standard FFN layer with an up-projection
$\mathbf{W}_{up}^{(\ell)} \in \mathbb{R}^{d \times d_{ff}}$
and a down-projection
$\mathbf{W}_{down}^{(\ell)} \in \mathbb{R}^{d_{ff} \times d}$.
For neuron $j$ in layer $\ell$, CASS assigns a binary mask
$M_{\mathrm{neuron}}^{(\ell,j)} \in \{0,1\}$.

Since neuron $j$ corresponds to the $j$-th intermediate channel, its parameters consist of the $j$-th column of the up-projection and the $j$-th row of the down-projection. The neuron mask is therefore broadcast as
\begin{equation}
\begin{aligned}
    \Delta \widetilde{\mathbf{W}}_{up}^{(\ell)}[:,j]
    &=
    M_{\mathrm{neuron}}^{(\ell,j)}
    \cdot
    \Delta \mathbf{W}_{up}^{(\ell)}[:,j],
    \\
    \Delta \widetilde{\mathbf{W}}_{down}^{(\ell)}[j,:]
    &=
    M_{\mathrm{neuron}}^{(\ell,j)}
    \cdot
    \Delta \mathbf{W}_{down}^{(\ell)}[j,:].
\end{aligned}
\label{eq:ffn_broadcast}
\end{equation}

If the FFN uses an up-projection bias $\mathbf{b}_{up}^{(\ell)}$, we also apply the same mask to its corresponding channel:
\begin{equation}
    \Delta \widetilde{\mathbf{b}}_{up}^{(\ell)}[j]
    =
    M_{\mathrm{neuron}}^{(\ell,j)}
    \cdot
    \Delta \mathbf{b}_{up}^{(\ell)}[j].
\end{equation}
The down-projection bias $\mathbf{b}_{down}^{(\ell)}$ is shared across all FFN neurons and is not naturally attributable to a single intermediate channel. We therefore leave it unmasked in our implementation.

This rule is consistent with the additive FFN decomposition in Appendix~\ref{app:derivation}. Masking neuron $j$ suppresses task-specific updates to the parameters that generate and project the corresponding intermediate activation channel, while preserving the base model parameters of that neuron.

\subsection{Gated FFNs}
\label{app:gated_ffn_broadcast}

Modern decoder-only language models, including Qwen and Gemma-style architectures, commonly use gated FFNs such as SwiGLU. In this case, the intermediate activation is computed from a gate projection and an up projection:
\begin{equation}
    \mathbf{M}
    =
    \mathrm{SiLU}
    \left(
    \mathbf{X}\mathbf{W}_{gate}
    \right)
    \odot
    \left(
    \mathbf{X}\mathbf{W}_{up}
    \right),
\end{equation}
where
$\mathbf{W}_{gate}, \mathbf{W}_{up} \in \mathbb{R}^{d \times d_{ff}}$.
The output is then obtained through
$\mathbf{W}_{down} \in \mathbb{R}^{d_{ff} \times d}$.

Although gated FFNs use a different mechanism to compute the intermediate activation, each intermediate channel $j$ still contributes to the residual stream through the corresponding row of $\mathbf{W}_{down}$. Therefore, the same neuron-level mask can be applied to the triplet of parameter slices associated with channel $j$:
\begin{equation}
\begin{aligned}
    \Delta \widetilde{\mathbf{W}}_{gate}^{(\ell)}[:,j]
    &=
    M_{\mathrm{neuron}}^{(\ell,j)}
    \cdot
    \Delta \mathbf{W}_{gate}^{(\ell)}[:,j],
    \\
    \Delta \widetilde{\mathbf{W}}_{up}^{(\ell)}[:,j]
    &=
    M_{\mathrm{neuron}}^{(\ell,j)}
    \cdot
    \Delta \mathbf{W}_{up}^{(\ell)}[:,j],
    \\
    \Delta \widetilde{\mathbf{W}}_{down}^{(\ell)}[j,:]
    &=
    M_{\mathrm{neuron}}^{(\ell,j)}
    \cdot
    \Delta \mathbf{W}_{down}^{(\ell)}[j,:].
\end{aligned}
\label{eq:gated_ffn_broadcast}
\end{equation}

If gate or up projection biases are present, the same mask is applied to the corresponding bias channel. This ensures that the entire gated intermediate channel is treated as one functional unit.

\subsection{LoRA Implementation}
\label{app:lora_broadcast}

CASS can also be used with parameter-efficient fine-tuning methods such as LoRA. In this case, the task-specific update to a weight matrix is represented by a low-rank update $\Delta \mathbf{W}_{\mathrm{LoRA}}$, rather than by directly fine-tuning the full-rank matrix.

The key principle is that CASS applies the structural mask to the effective update along the same channel dimension as in the full-rank case. For example, for an FFN down-projection, neuron $j$ corresponds to the $j$-th input channel of $\mathbf{W}_{down}$. Therefore, CASS masks the effective LoRA update on the same row:
\begin{equation}
    \Delta \widetilde{\mathbf{W}}_{\mathrm{LoRA},down}^{(\ell)}[j,:]
    =
    M_{\mathrm{neuron}}^{(\ell,j)}
    \cdot
    \Delta \mathbf{W}_{\mathrm{LoRA},down}^{(\ell)}[j,:].
\end{equation}
Similarly, for an up or gate projection, neuron $j$ corresponds to the $j$-th output channel, and the mask is applied to the corresponding column of the effective LoRA update:
\begin{equation}
    \Delta \widetilde{\mathbf{W}}_{\mathrm{LoRA},up}^{(\ell)}[:,j]
    =
    M_{\mathrm{neuron}}^{(\ell,j)}
    \cdot
    \Delta \mathbf{W}_{\mathrm{LoRA},up}^{(\ell)}[:,j].
\end{equation}

In practice, this can be implemented in two equivalent ways. One may explicitly apply the mask to the effective low-rank update $\Delta \mathbf{W}_{\mathrm{LoRA}}$ after composing the LoRA factors. Alternatively, for efficiency, one may apply the mask to the LoRA factor whose dimension aligns with the masked channel. Both implementations yield the same masked effective update.

For CASS-Tuning with LoRA, we mask gradients or optimizer updates associated with inactive structural channels. As with full-rank fine-tuning, inactive channels do not acquire task-specific updates. For optimizers with momentum or decoupled weight decay, we ensure that inactive effective updates remain zero so that the corresponding parameters preserve their base-model behavior.

\paragraph{Summary.}
The broadcasting rules above convert component-level masks into parameter-level masks while preserving the structural interpretation of CASS. Attention head masks are applied to the query, key, value, and output-projection slices of each head. FFN neuron masks are applied to the projection slices associated with the corresponding intermediate channel. For gated FFNs and LoRA, the same principle is applied to the effective update along the structurally aligned channel dimension.

\clearpage
\section{Extended Analysis on FFN}
\label{app:ffn_analysis}

In Section~\ref{sec:analysis_motivation}, we presented preliminary evidence that task-relevant FFN neurons are sparse and partially separated across tasks. In this section, we provide extended visualizations on Qwen2.5-1.5B-Instruct across four task domains: Coding, Instruction Following, Mathematics, and Safety. The goal is to examine whether the sparsity and cross-task separation observed in the main text persist beyond a single task and a single layer. Similar patterns are also observed on ViT and RoBERTa. To keep the appendix concise, we focus the detailed breakdown on Qwen2.5-1.5B-Instruct results.

\subsection{Universality of FFN Contribution Sparsity}
\label{app:ffn_sparsity}

We first examine the distribution of neuron contribution scores across different task domains. For visualization, we follow the same normalized contribution score used in the main text, where larger values indicate stronger additive contribution to the residual stream under task-specific inputs.

\begin{figure}[h]
    \centering
    \begin{subfigure}{0.24\textwidth}
        \includegraphics[width=\linewidth]{Pic/neuron_cdf_coding.png}
        \caption{Coding}
    \end{subfigure}
    \hfill
    \begin{subfigure}{0.24\textwidth}
        \includegraphics[width=\linewidth]{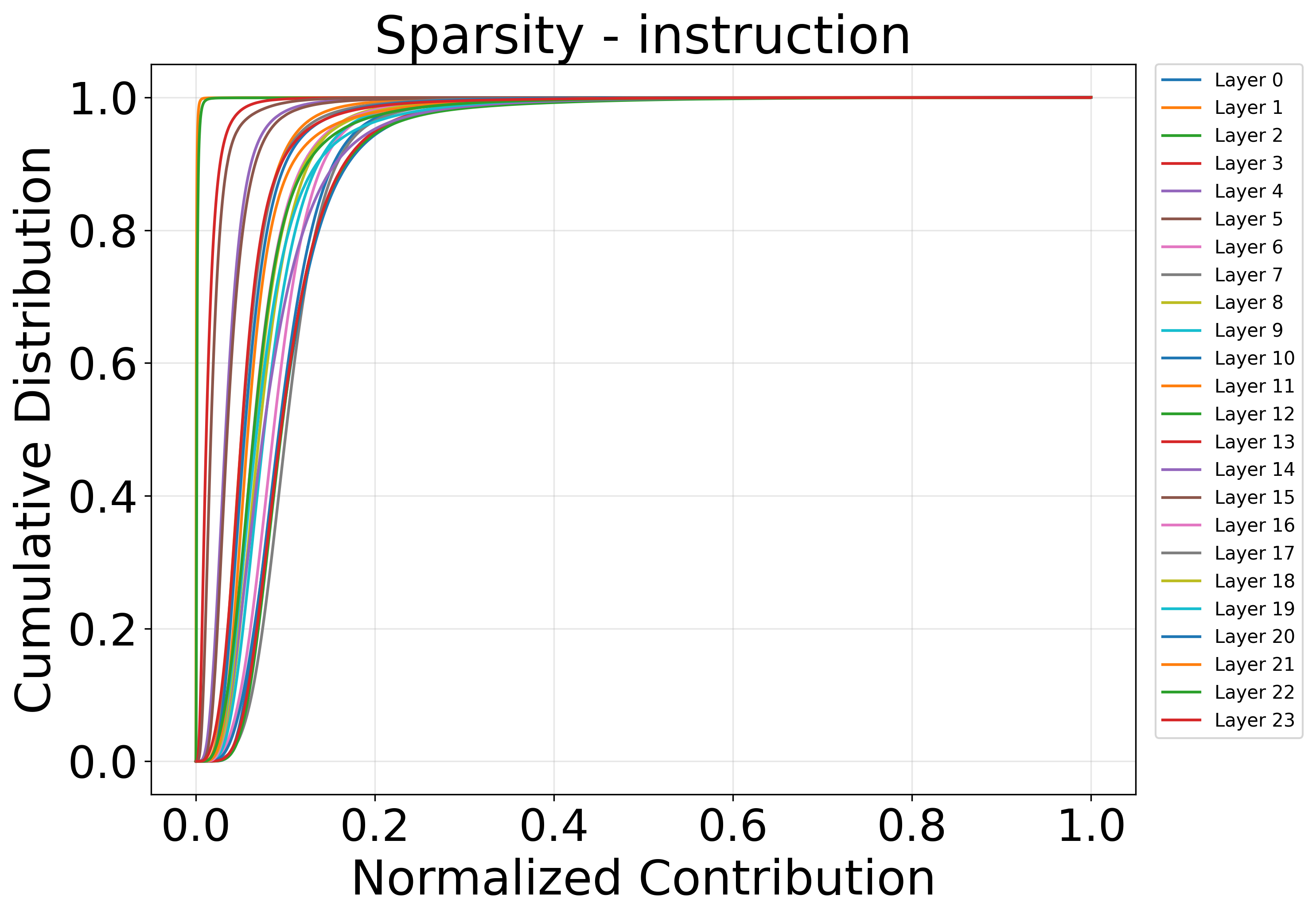}
        \caption{Instruction}
    \end{subfigure}
    \hfill
    \begin{subfigure}{0.24\textwidth}
        \includegraphics[width=\linewidth]{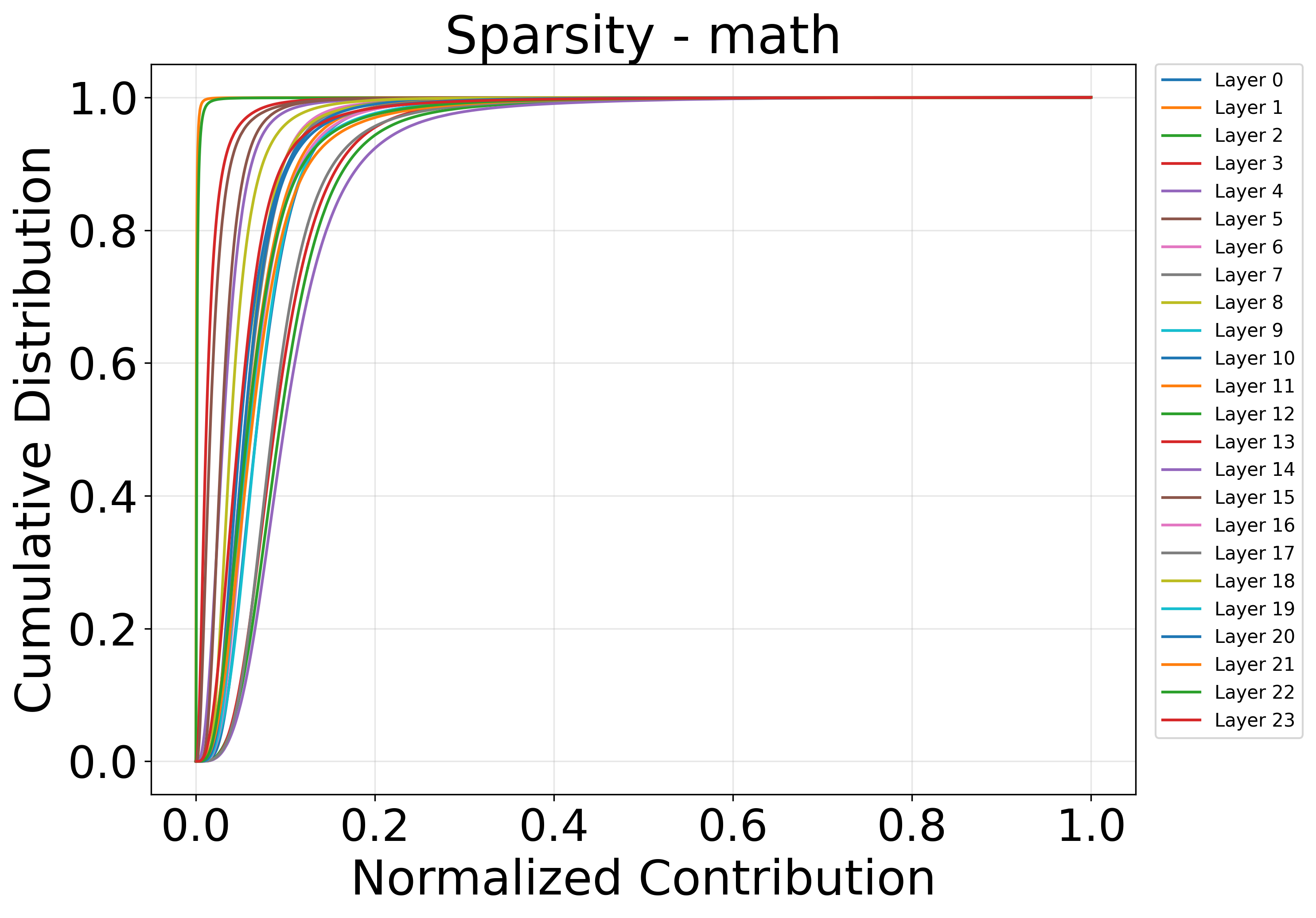}
        \caption{Mathematics}
    \end{subfigure}
    \hfill
    \begin{subfigure}{0.24\textwidth}
        \includegraphics[width=\linewidth]{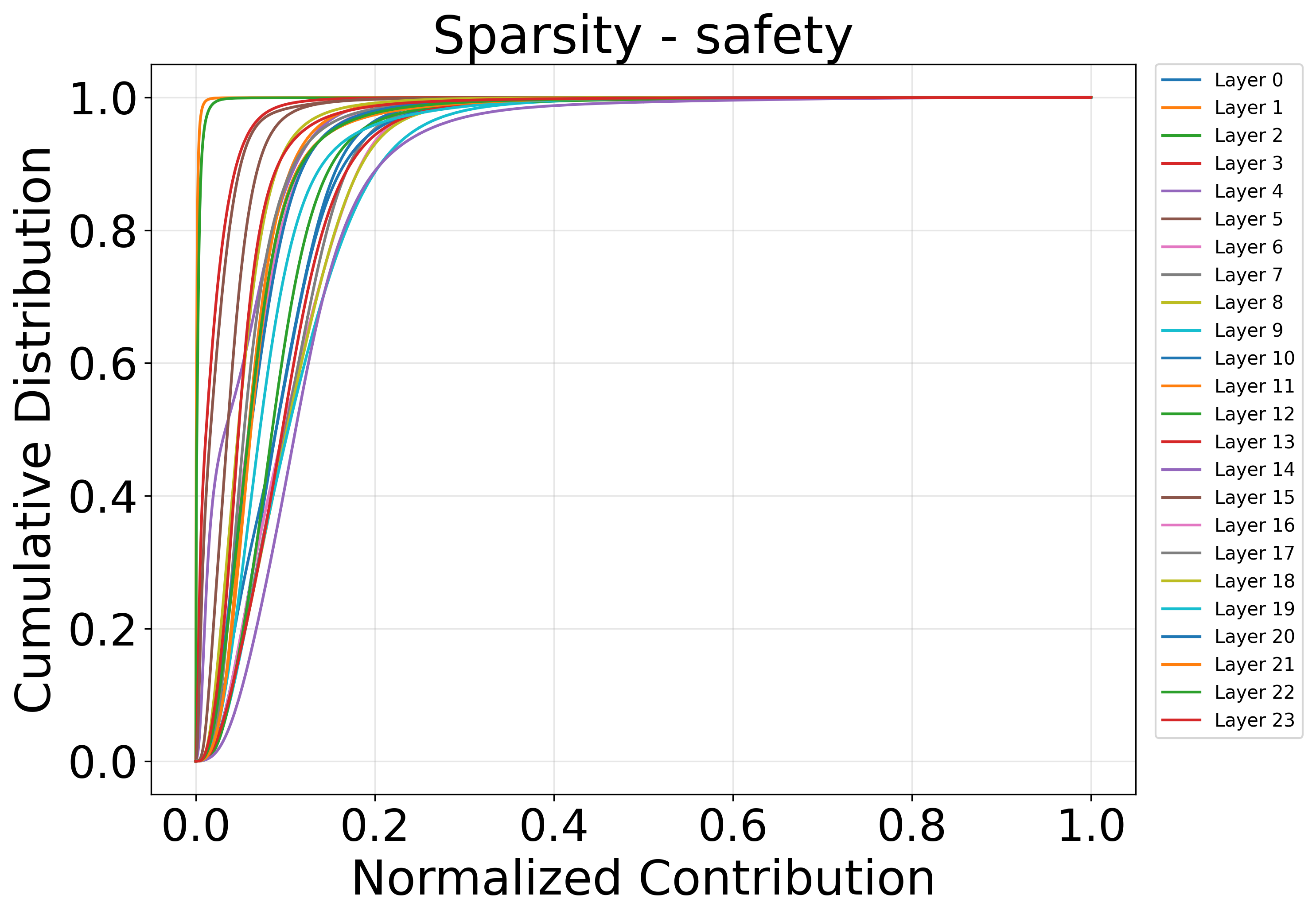}
        \caption{Safety}
    \end{subfigure}
    \caption{\textbf{CDF of FFN neuron contribution scores across tasks.}
    The CDF curves exhibit a sharp rise near zero across all four task domains, indicating that only a small subset of FFN neurons contributes prominently to each task. This suggests that task-relevant FFN activity is highly concentrated rather than uniformly distributed over all neurons.}
    \label{fig:app_cdf_all}
\end{figure}

Figure~\ref{fig:app_cdf_all} shows that the contribution distributions are highly skewed across all four domains. In each case, most neurons have near-zero contribution scores, while a small fraction of neurons forms a high-contribution tail. This observation supports the design choice of CASS: instead of retaining or pruning individual parameters based only on static weight magnitude, we identify task-relevant functional units by their activation-weighted contribution to the residual stream.

\subsection{Layer-wise Separation of Active FFN Neurons}
\label{app:ffn_layerwise_overlap}

We next analyze whether the active FFN neurons selected for different tasks overlap substantially. For each task and each layer, we select the top-20\% neurons according to their contribution scores and compute the pairwise Jaccard similarity between task-specific active sets:
\begin{equation}
    \mathrm{Jaccard}(t_1,t_2)
    =
    \frac{
    |\mathcal{N}_{t_1}^{(\ell)} \cap \mathcal{N}_{t_2}^{(\ell)}|
    }{
    |\mathcal{N}_{t_1}^{(\ell)} \cup \mathcal{N}_{t_2}^{(\ell)}|
    },
\end{equation}
where $\mathcal{N}_{t}^{(\ell)}$ denotes the selected active FFN neuron set for task $t$ in layer $\ell$.

\begin{figure*}[t]
    \centering
    \setlength{\tabcolsep}{2pt}
    \renewcommand{\arraystretch}{0.25}

    \begin{tabular}{cccc}
        \includegraphics[width=0.235\textwidth]{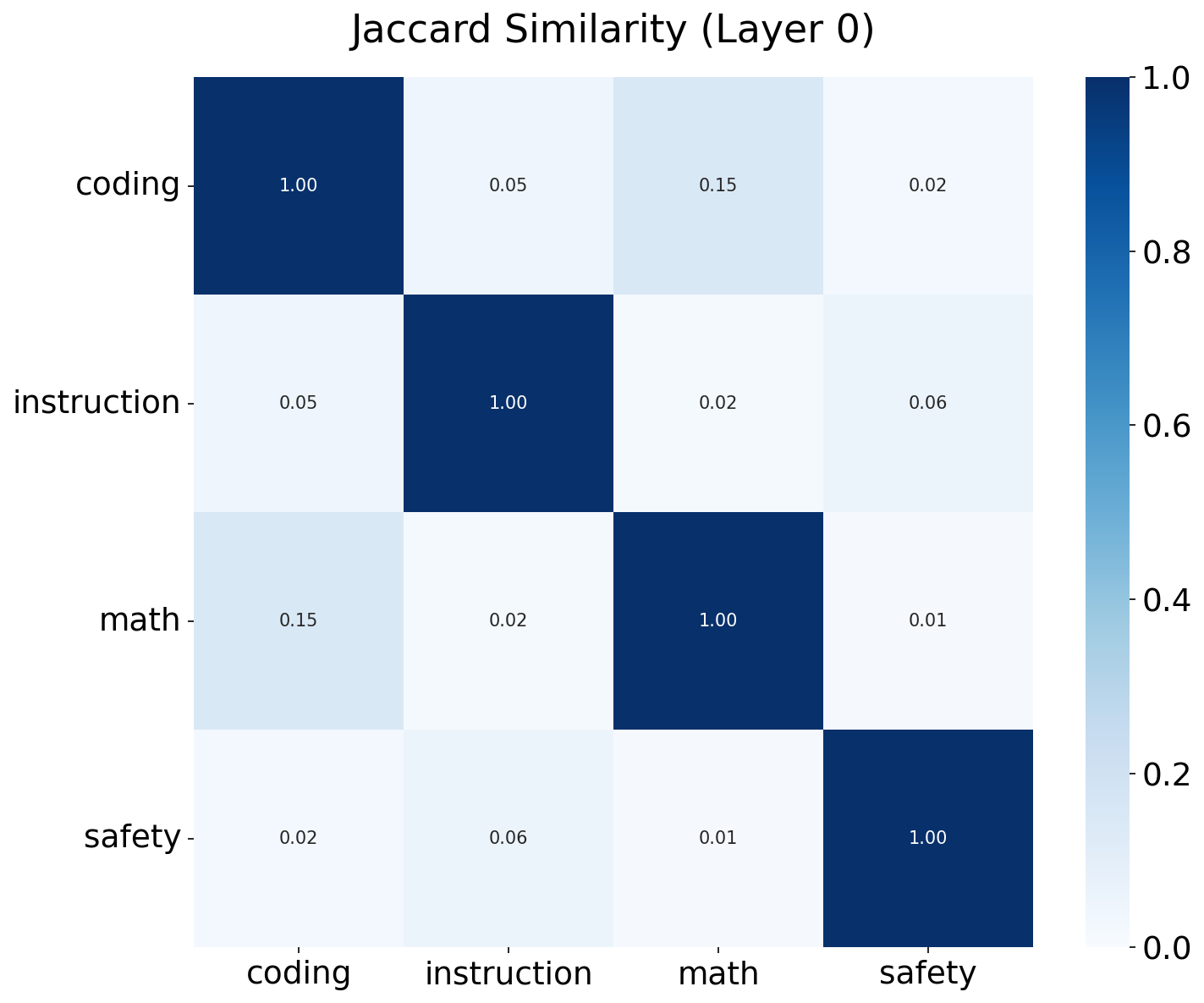} &
        \includegraphics[width=0.235\textwidth]{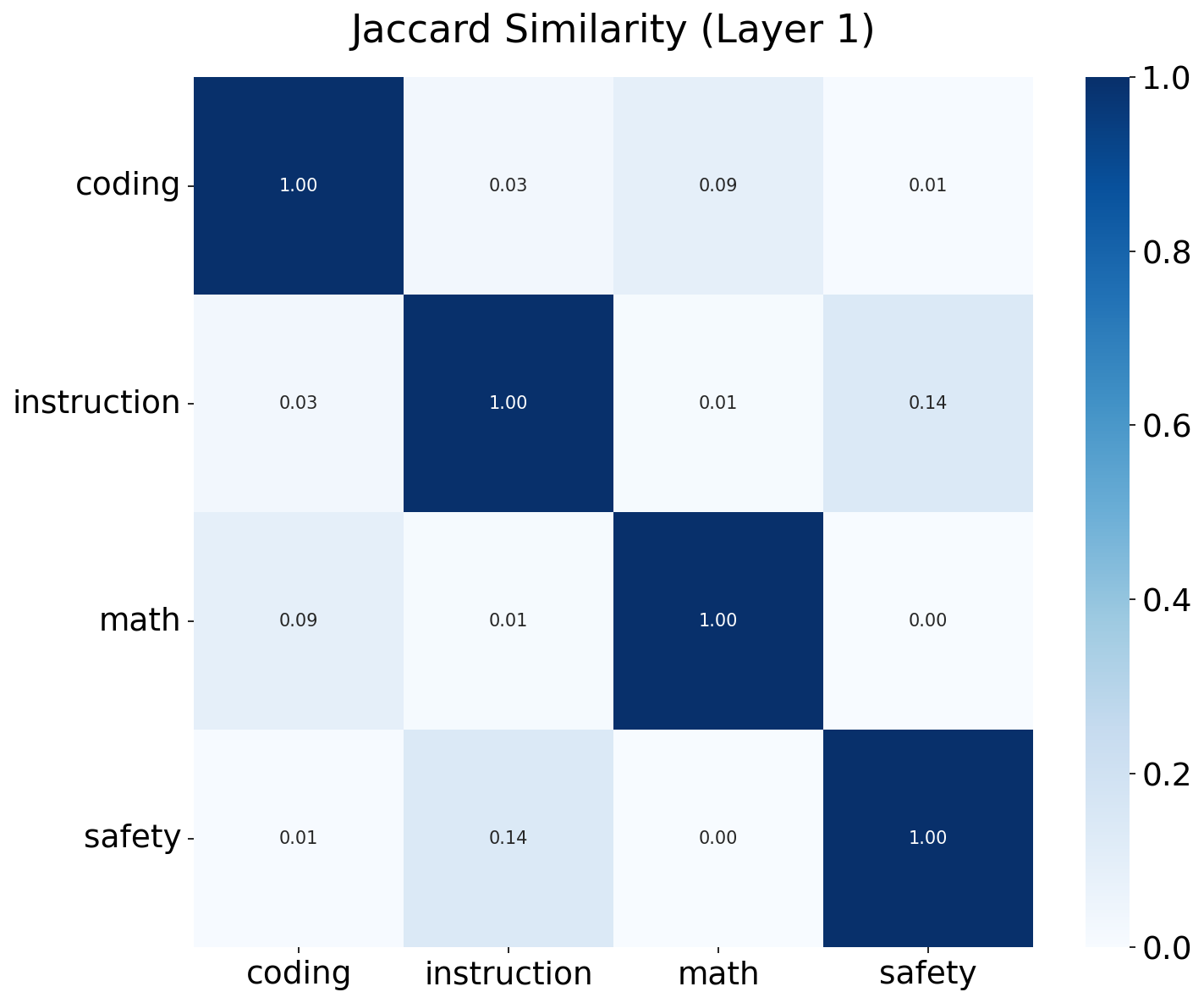} &
        \includegraphics[width=0.235\textwidth]{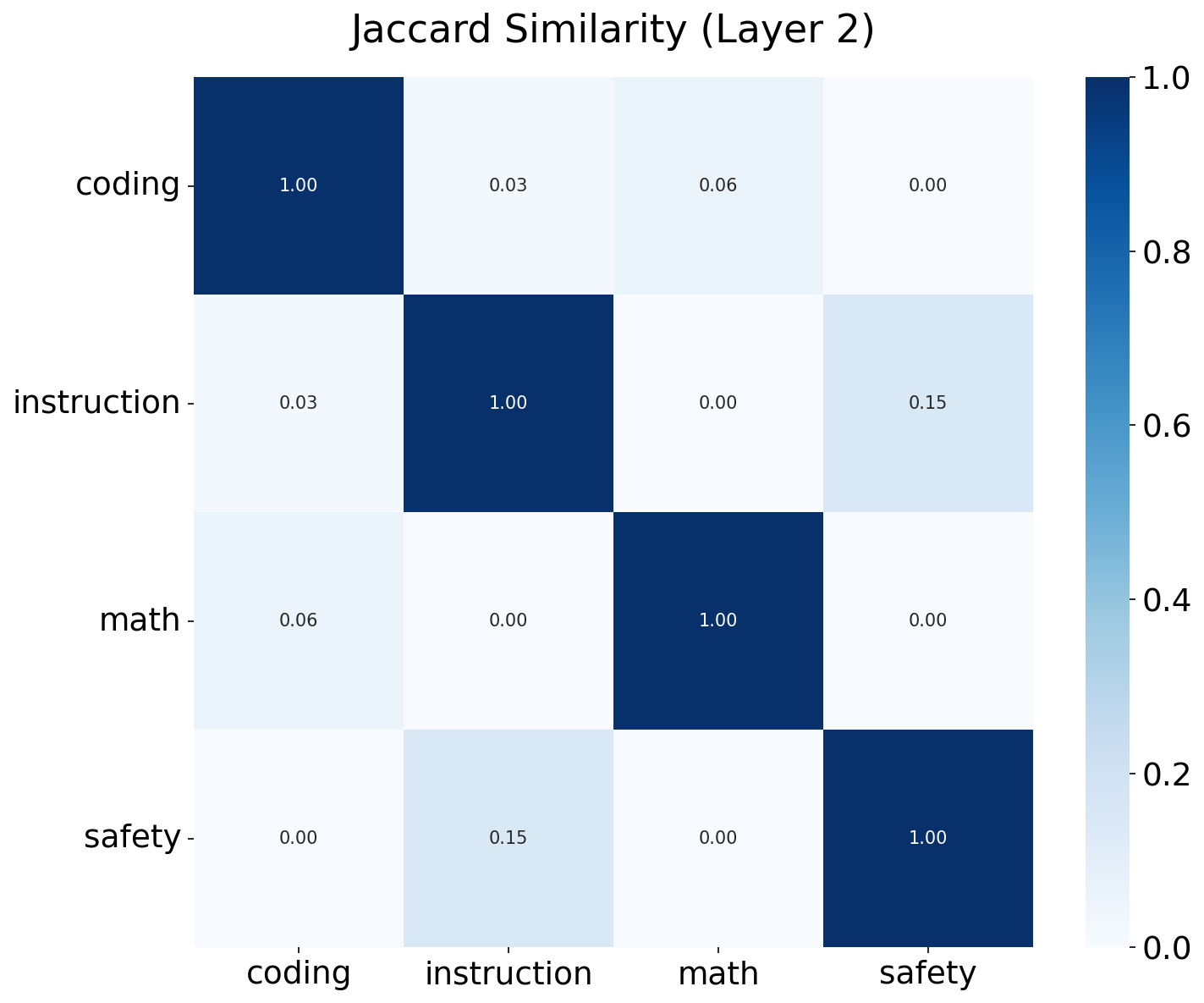} &
        \includegraphics[width=0.235\textwidth]{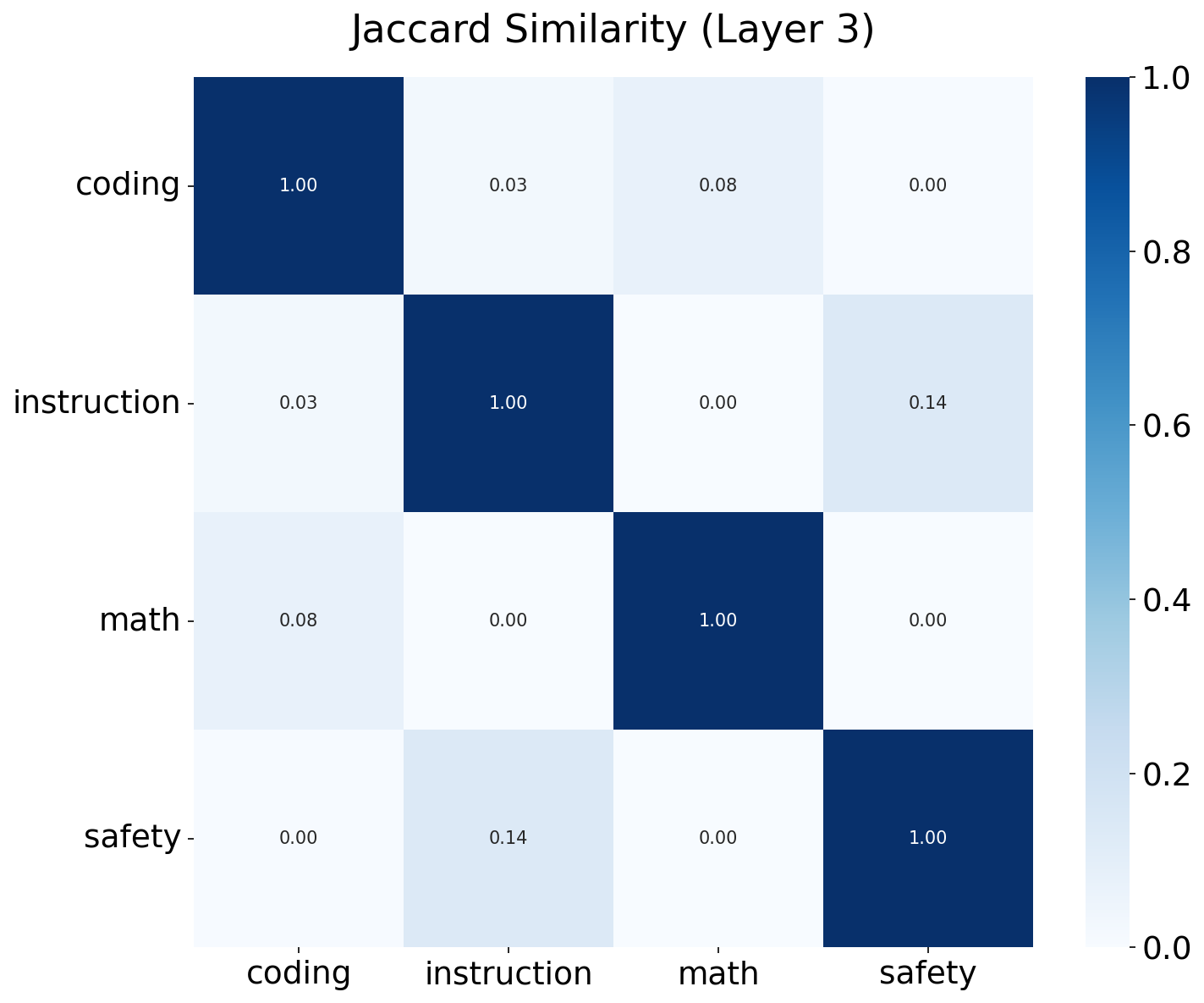} \\

        \includegraphics[width=0.235\textwidth]{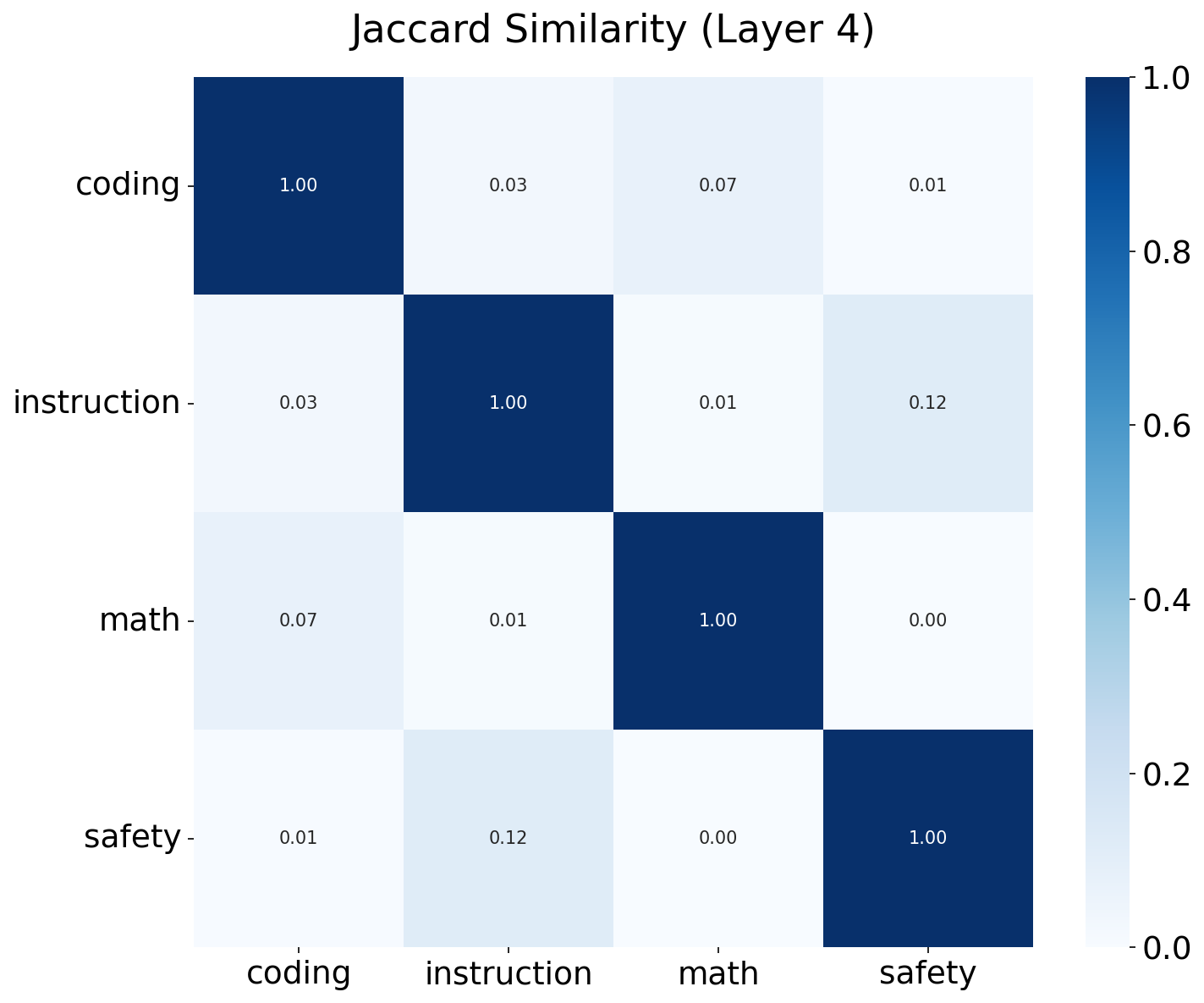} &
        \includegraphics[width=0.235\textwidth]{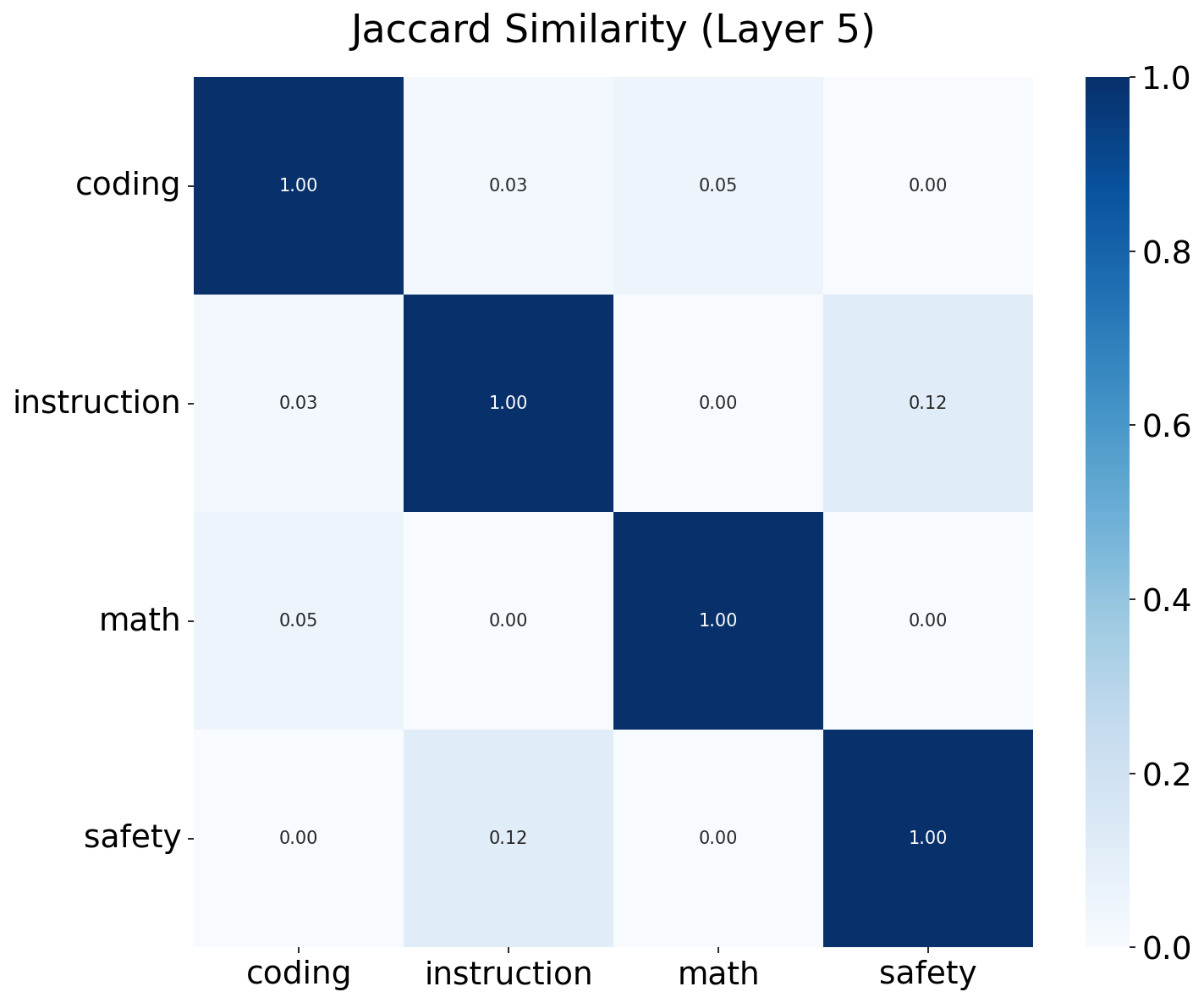} &
        \includegraphics[width=0.235\textwidth]{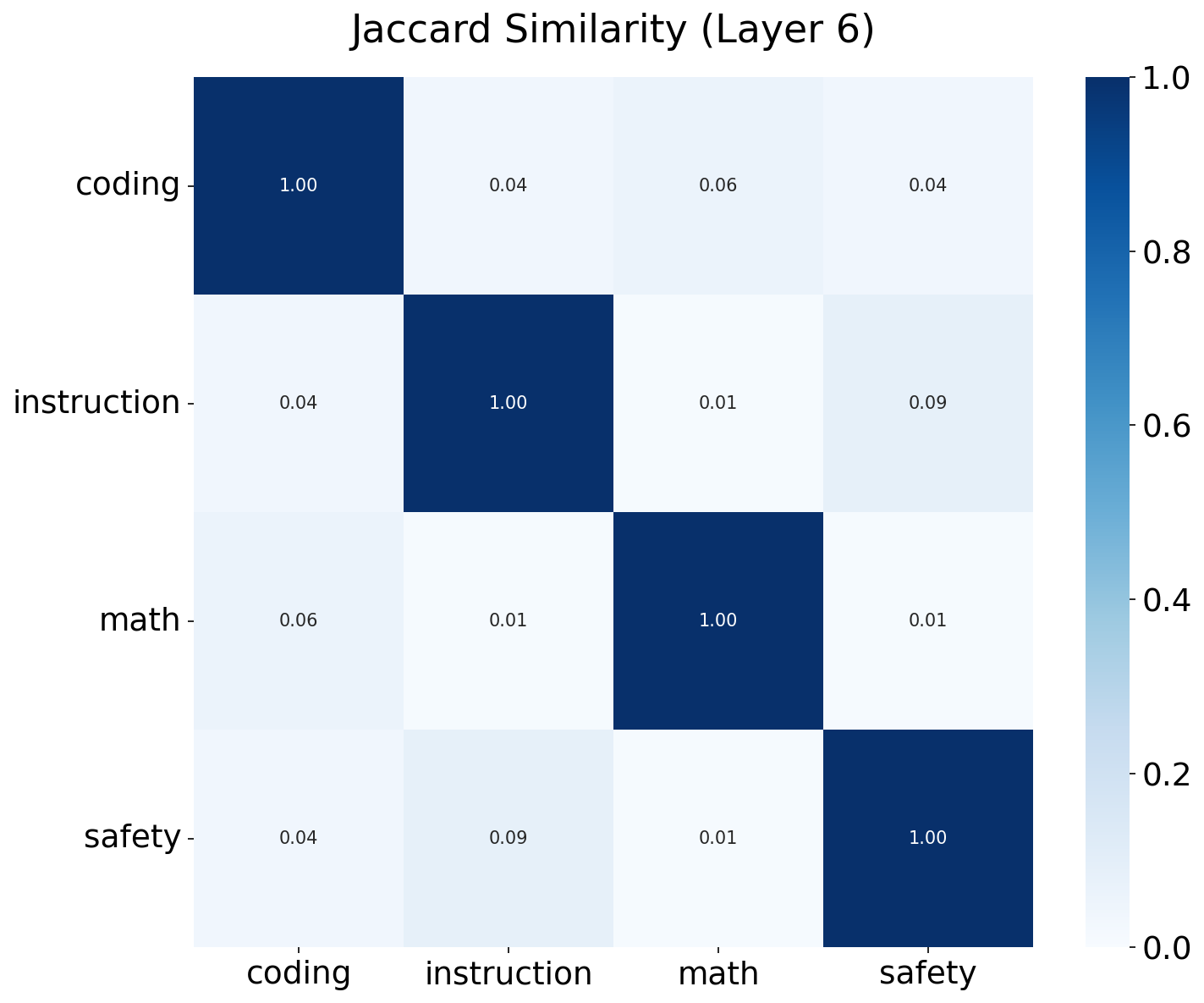} &
        \includegraphics[width=0.235\textwidth]{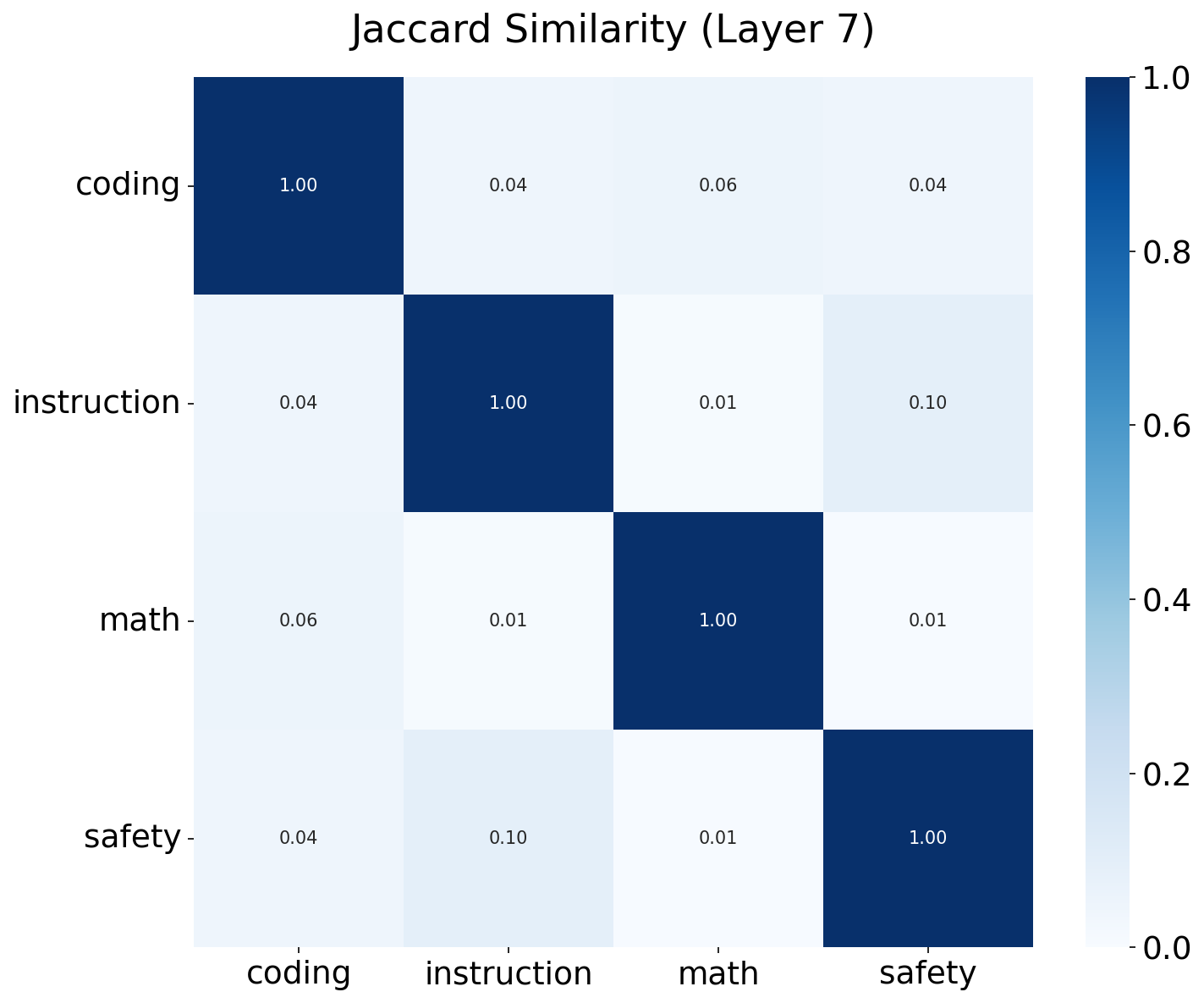} \\

        \includegraphics[width=0.235\textwidth]{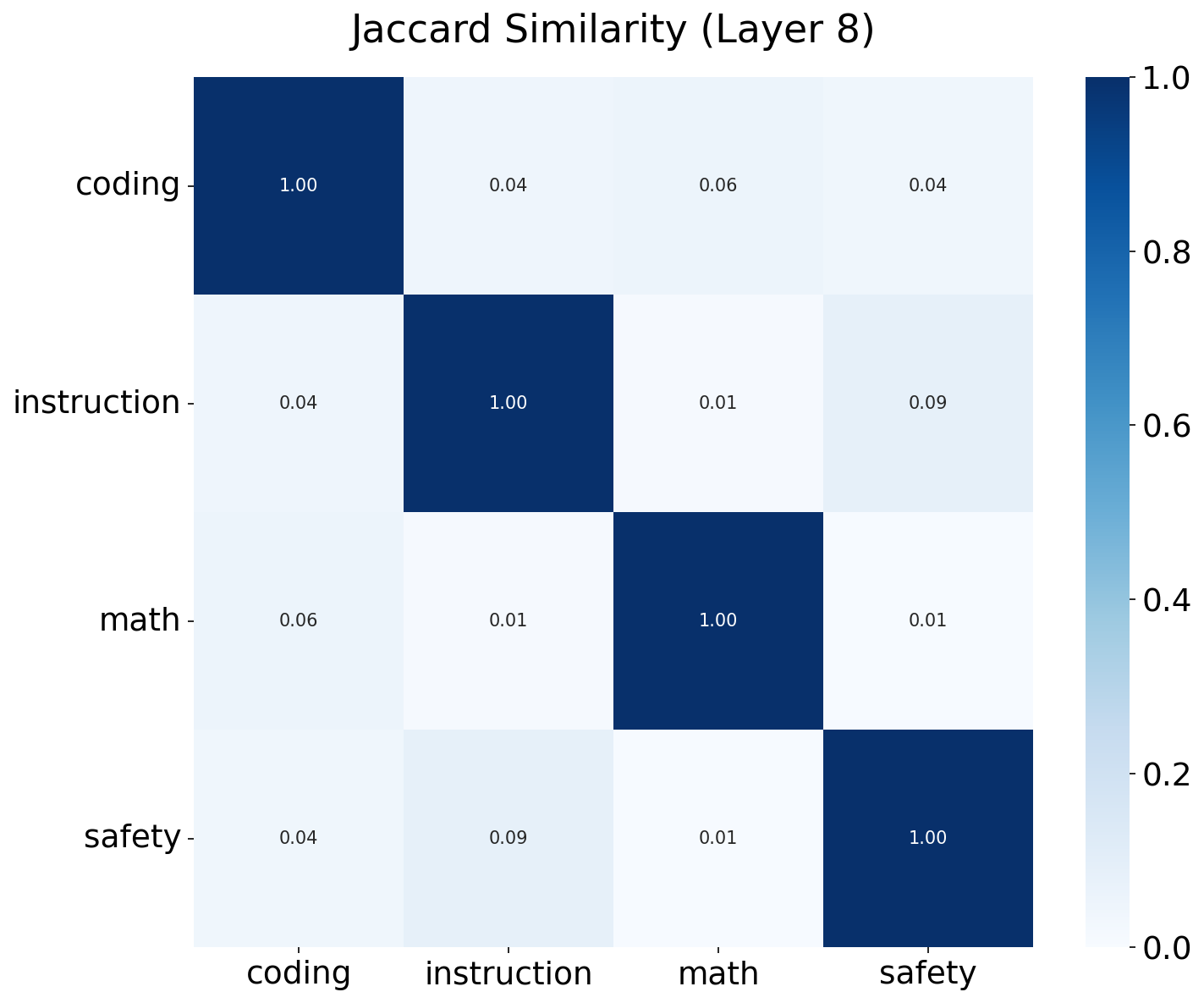} &
        \includegraphics[width=0.235\textwidth]{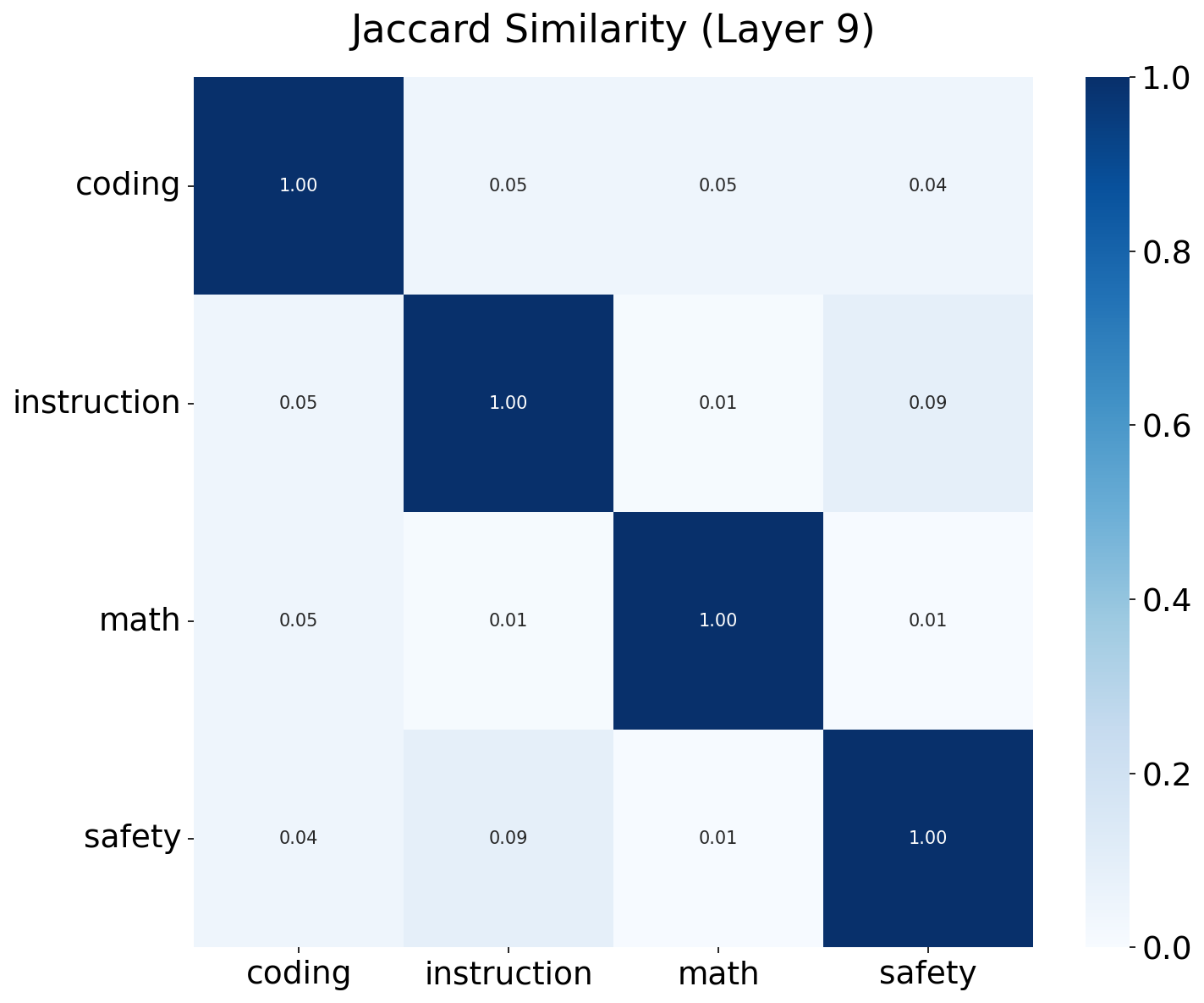} &
        \includegraphics[width=0.235\textwidth]{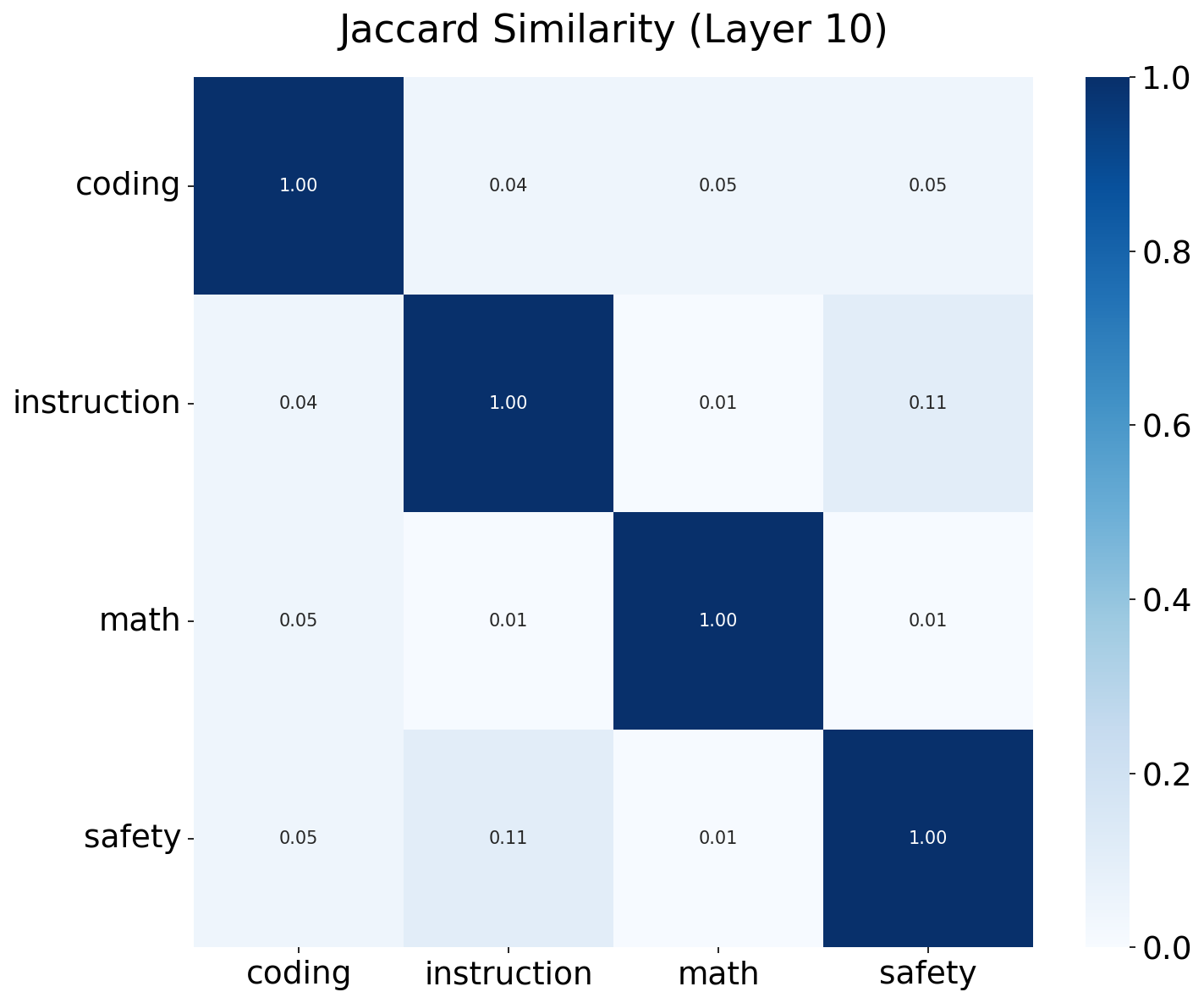} &
        \includegraphics[width=0.235\textwidth]{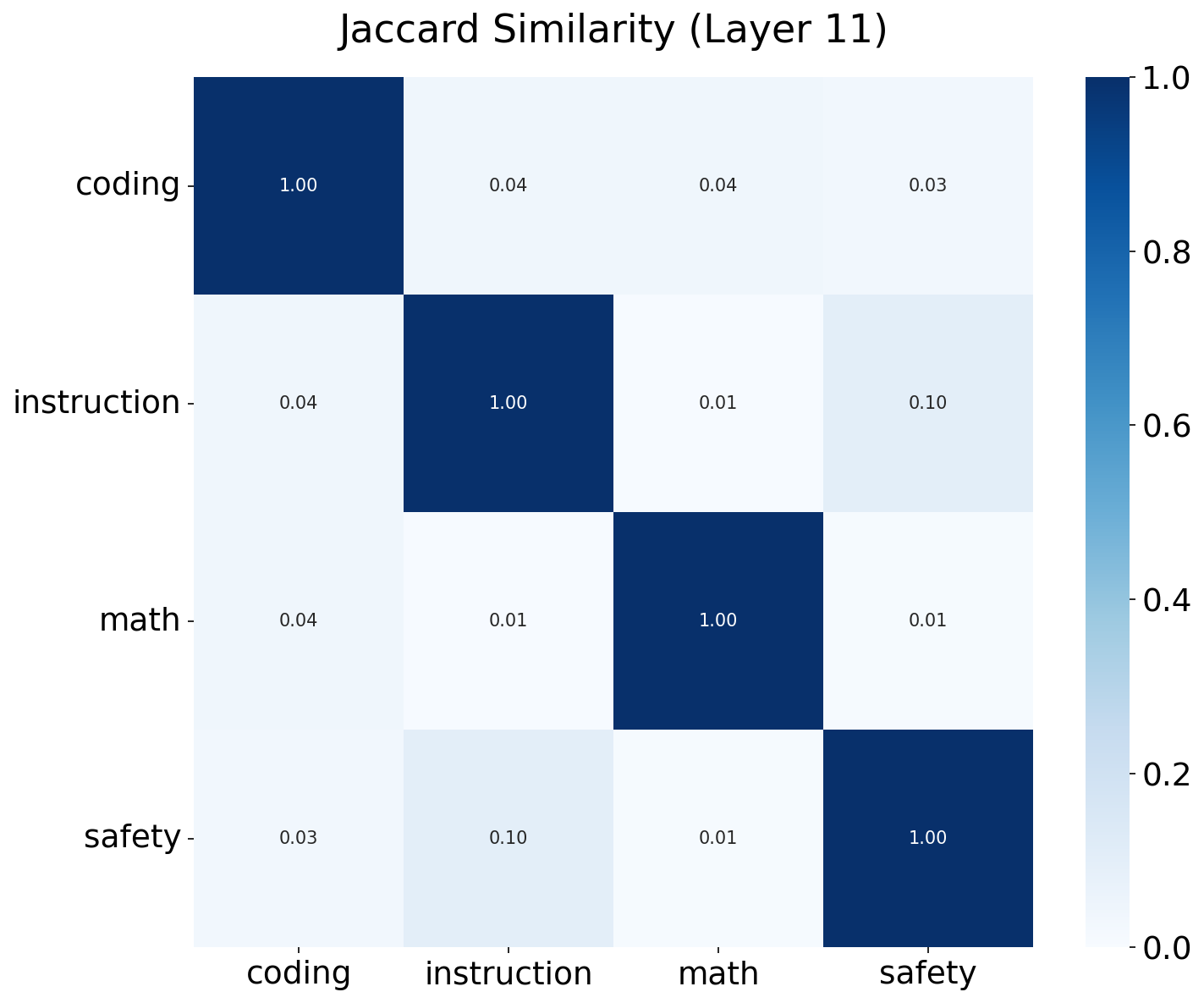} \\

        \includegraphics[width=0.235\textwidth]{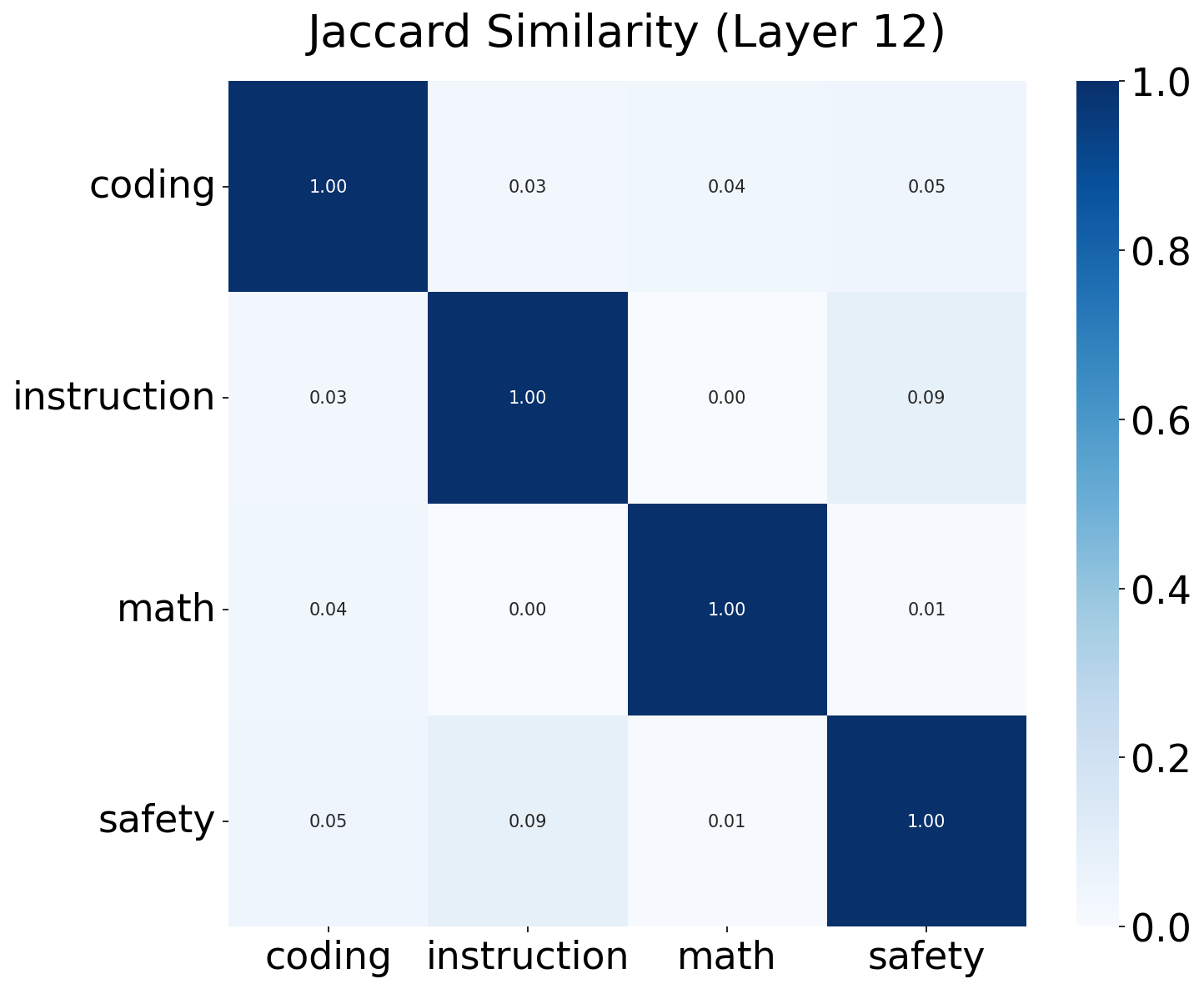} &
        \includegraphics[width=0.235\textwidth]{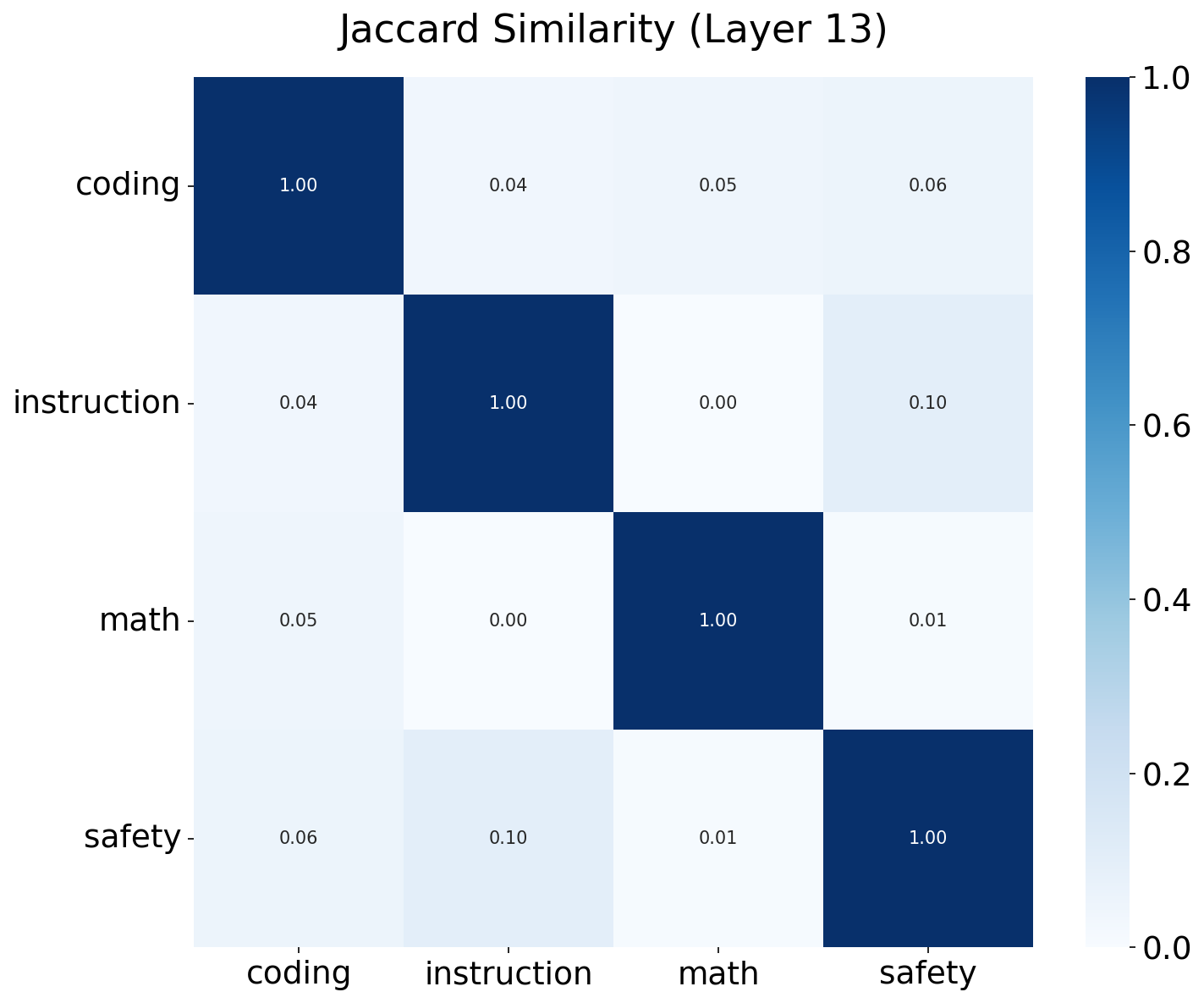} &
        \includegraphics[width=0.235\textwidth]{Pic/neuron_jaccard_20/layer_14.png} &
        \includegraphics[width=0.235\textwidth]{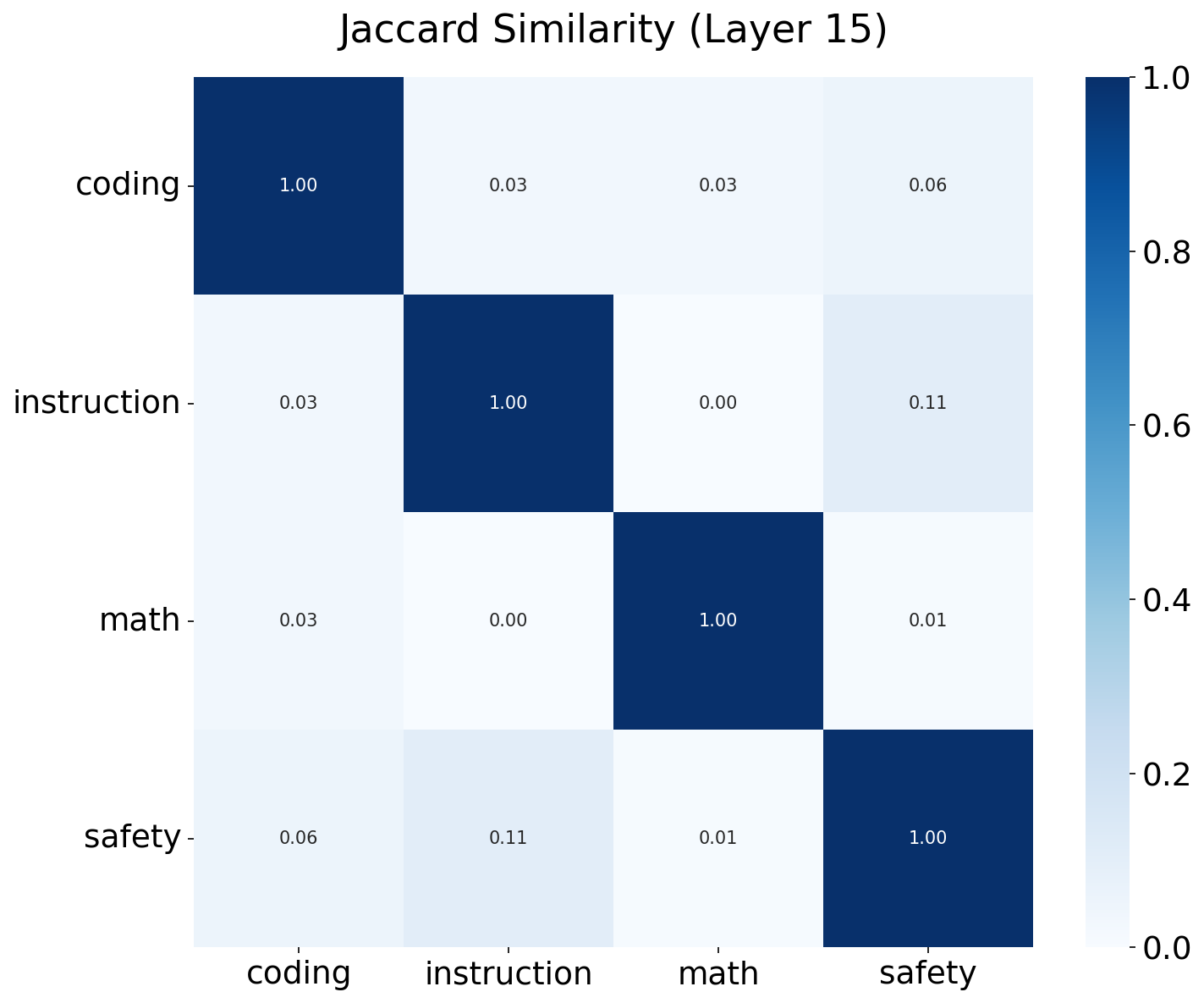} \\

        \includegraphics[width=0.235\textwidth]{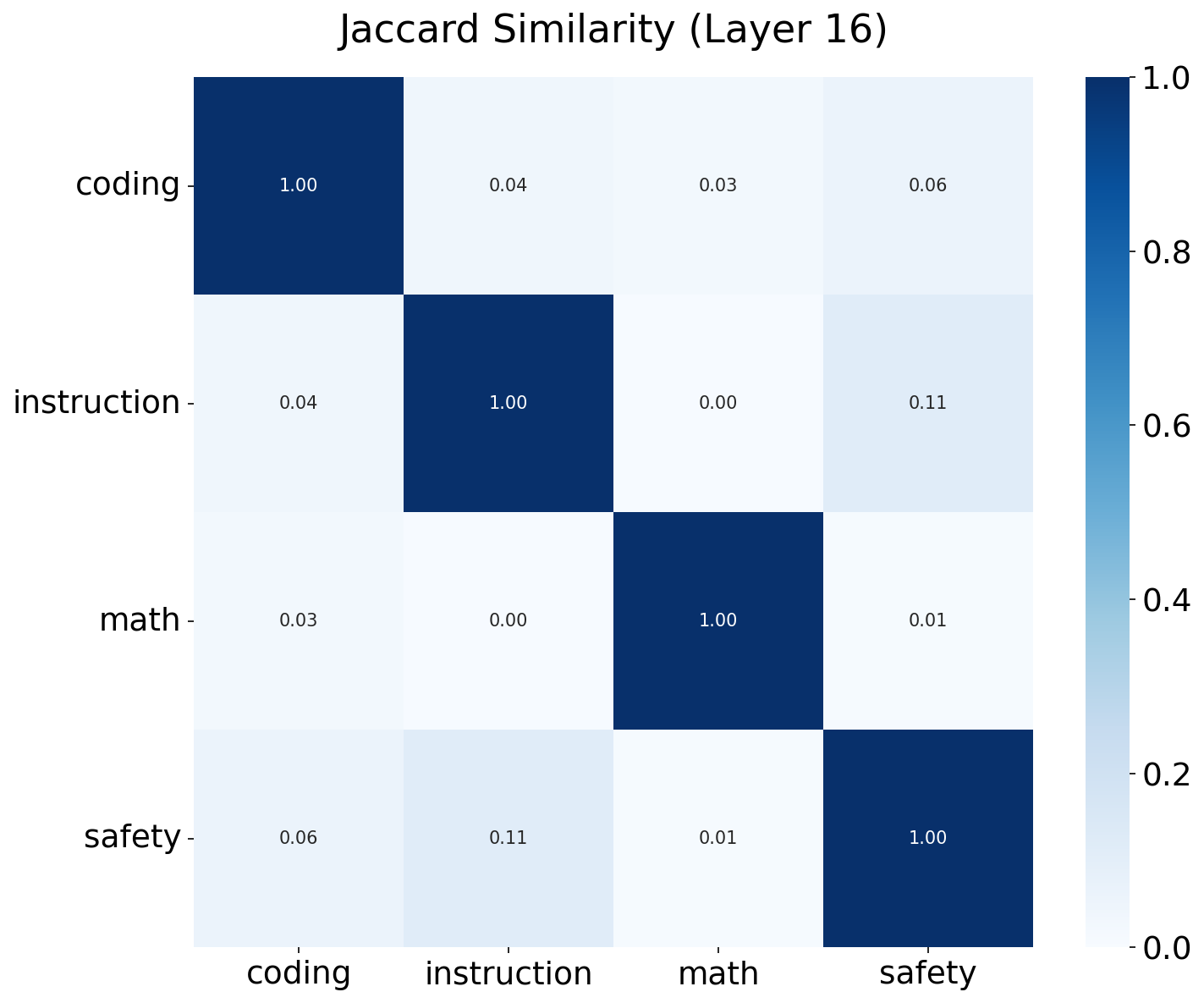} &
        \includegraphics[width=0.235\textwidth]{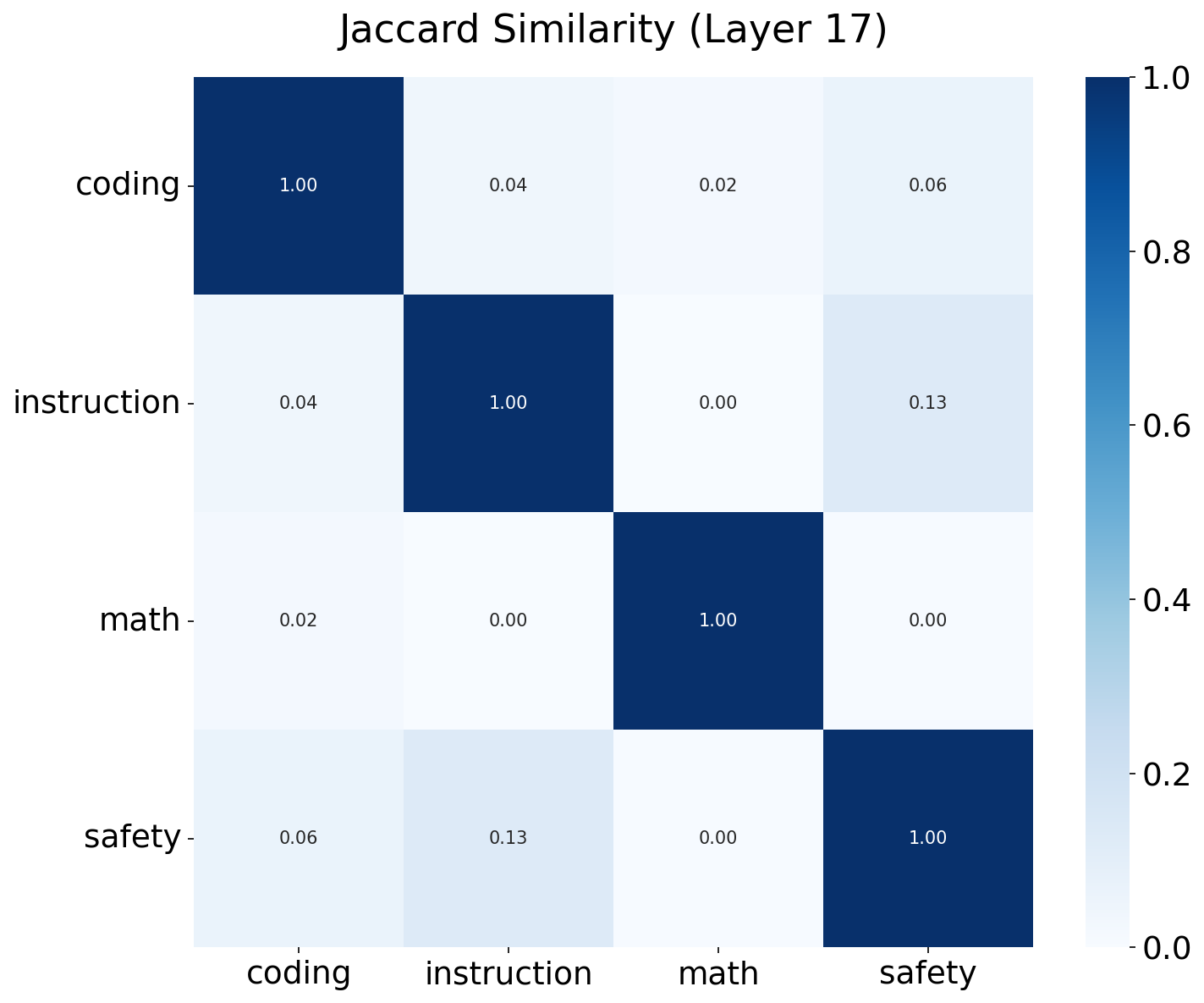} &
        \includegraphics[width=0.235\textwidth]{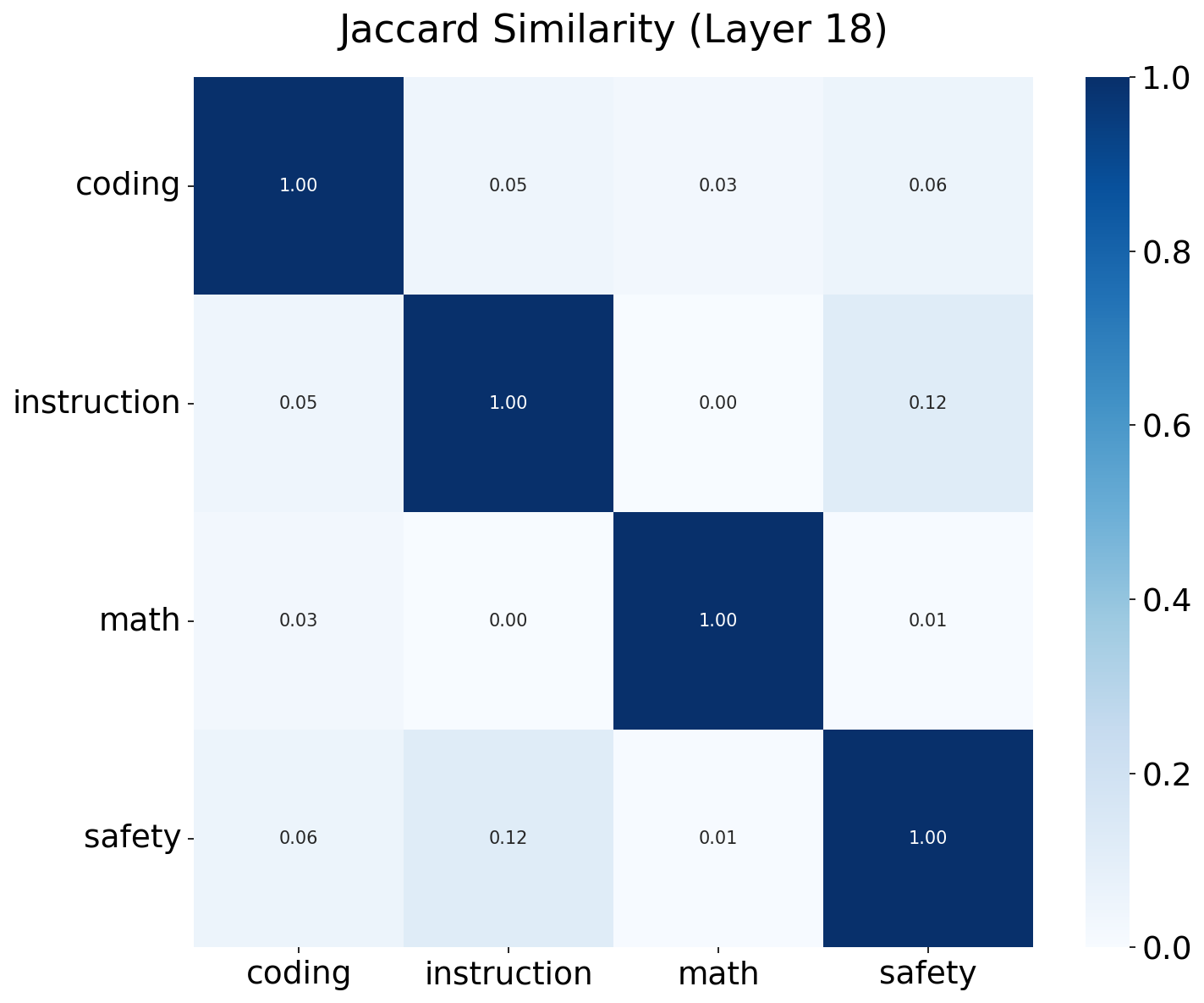} &
        \includegraphics[width=0.235\textwidth]{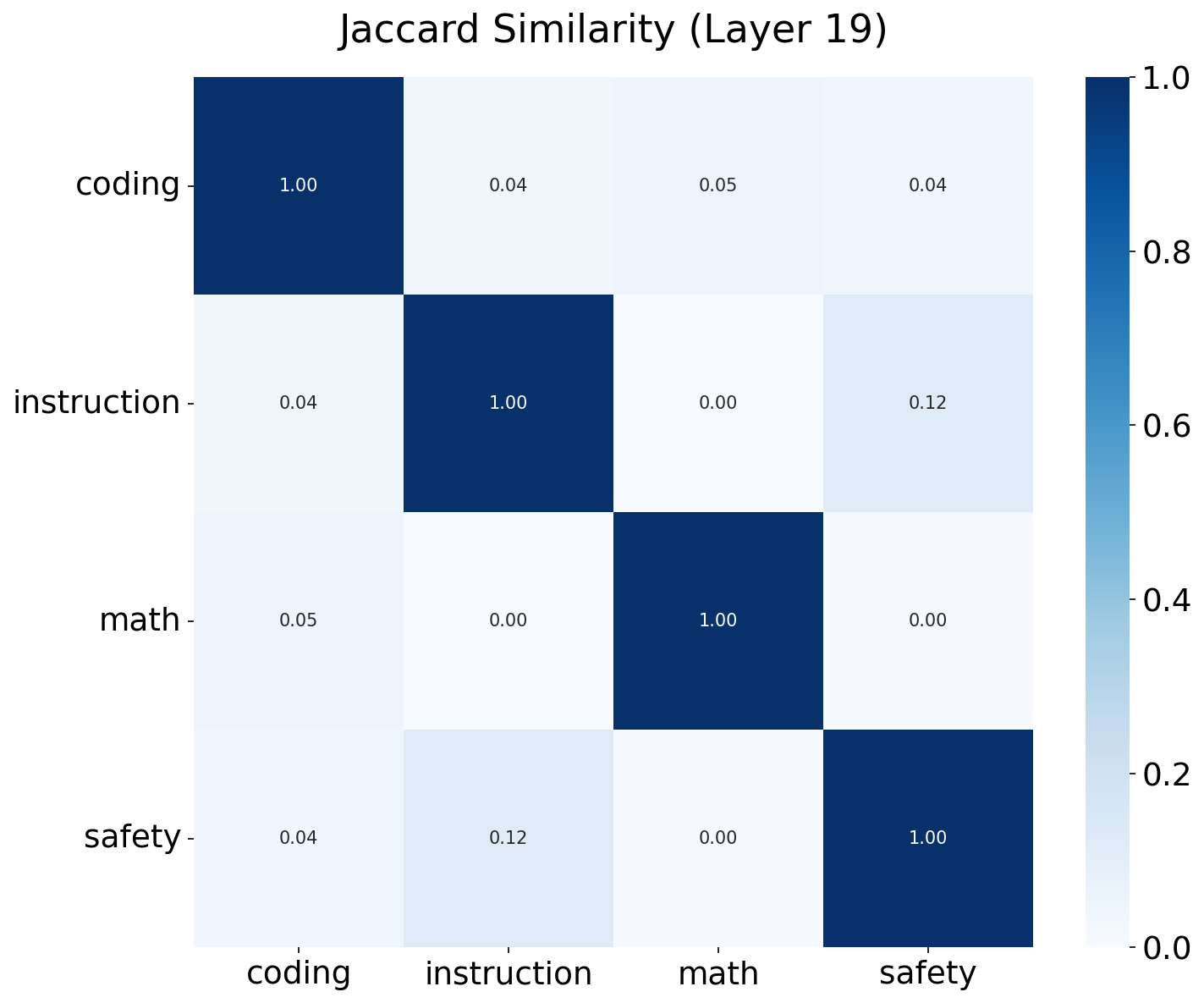} \\

        \includegraphics[width=0.235\textwidth]{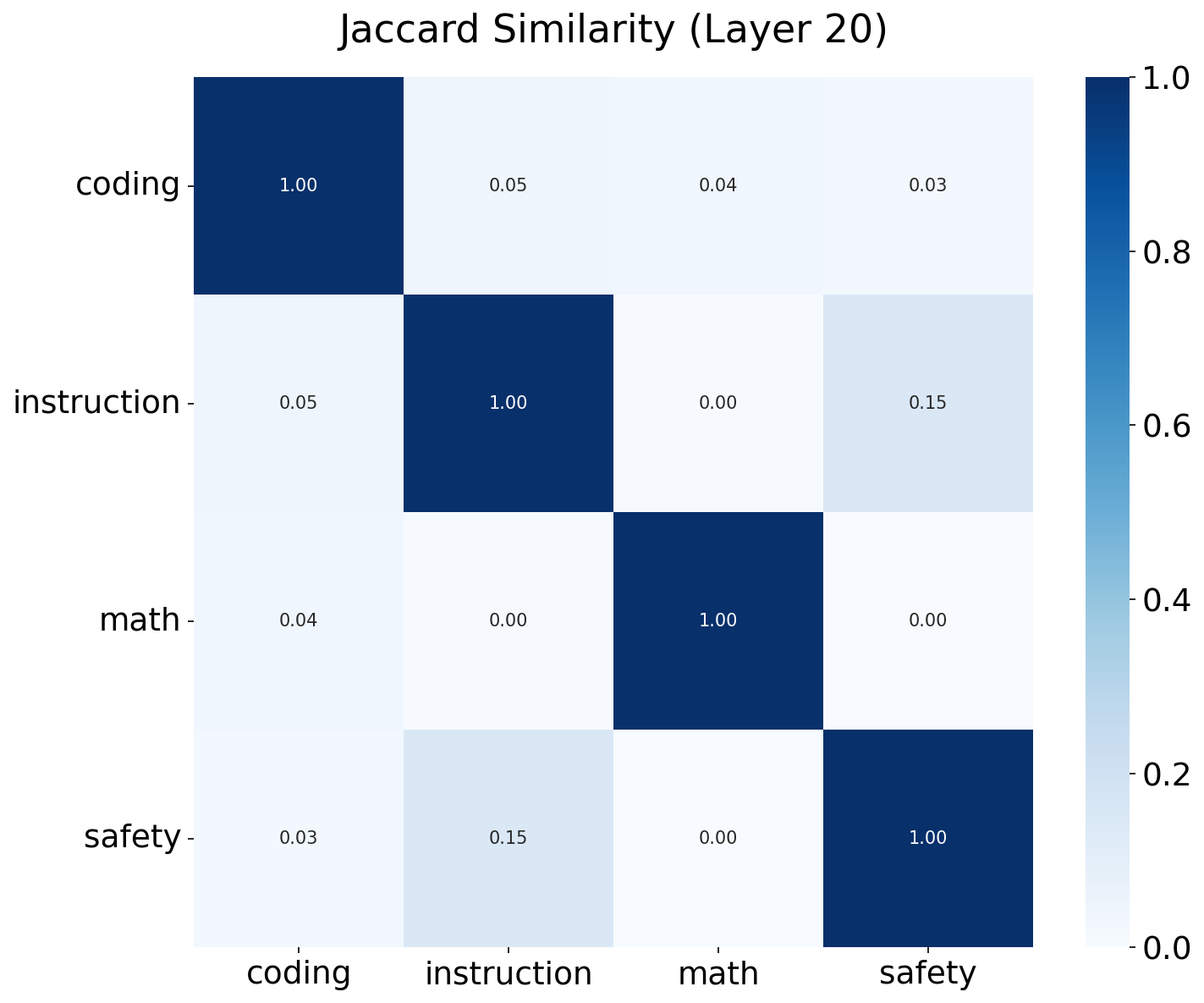} &
        \includegraphics[width=0.235\textwidth]{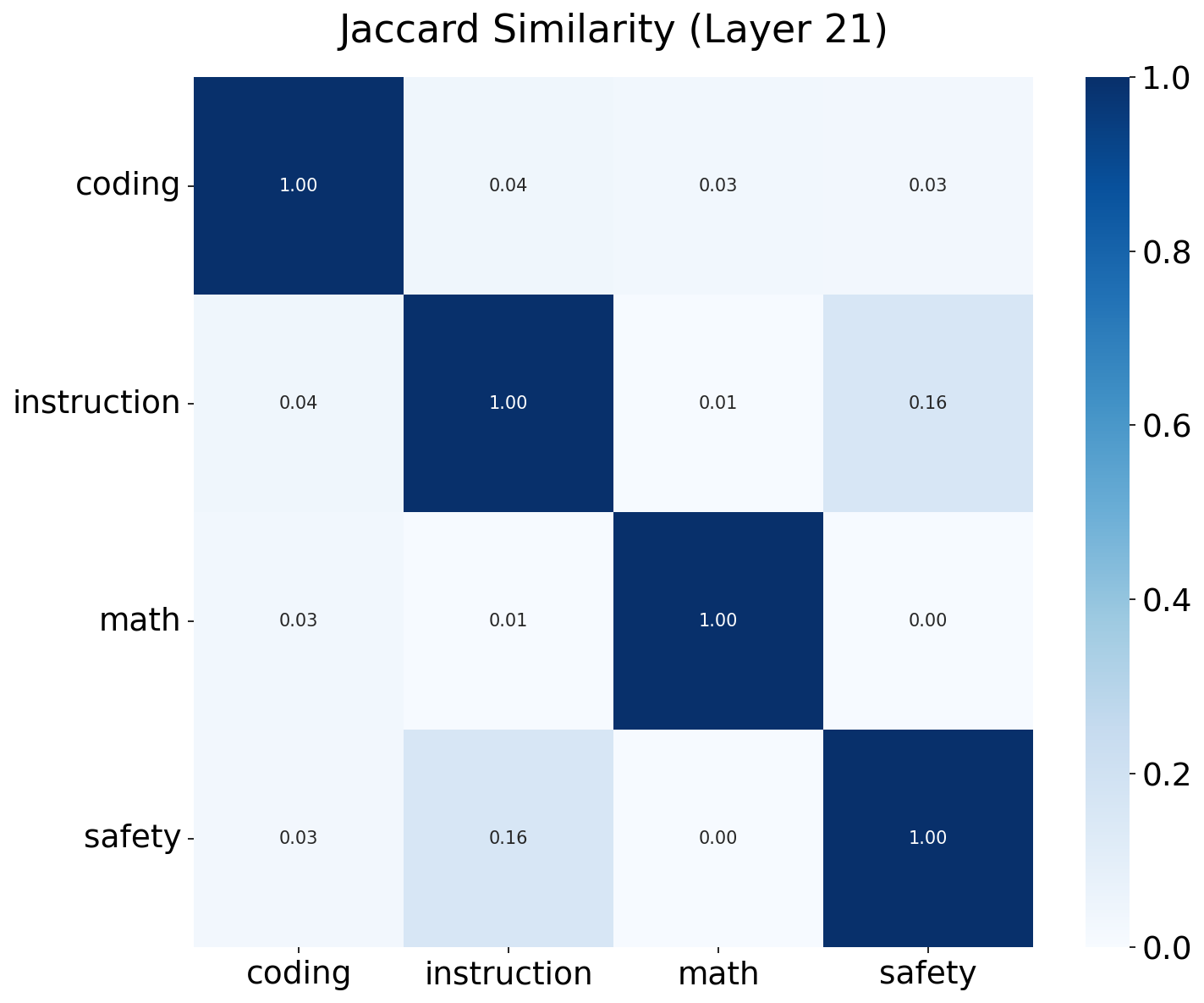} &
        \includegraphics[width=0.235\textwidth]{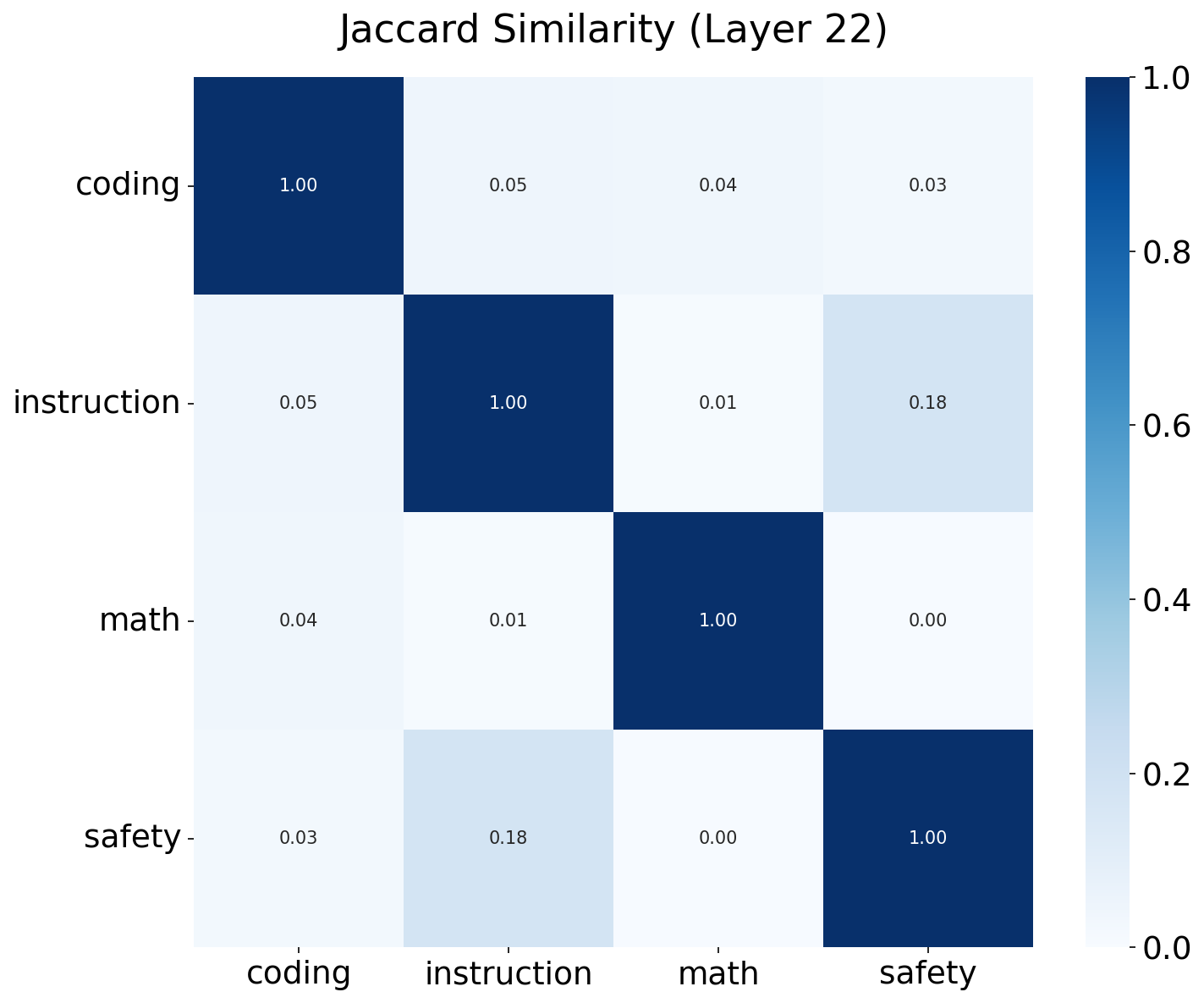} &
        \includegraphics[width=0.235\textwidth]{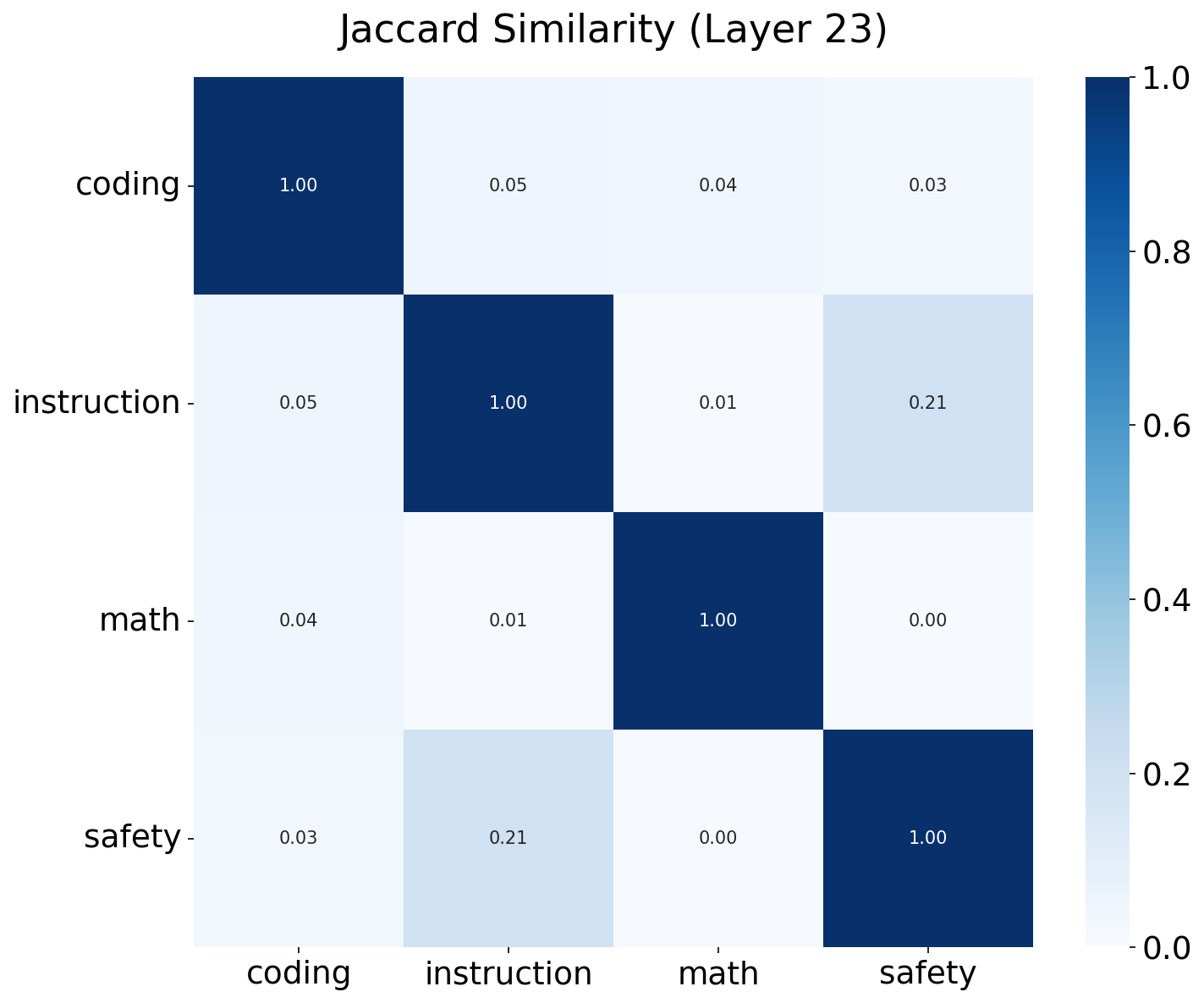} \\

        \includegraphics[width=0.235\textwidth]{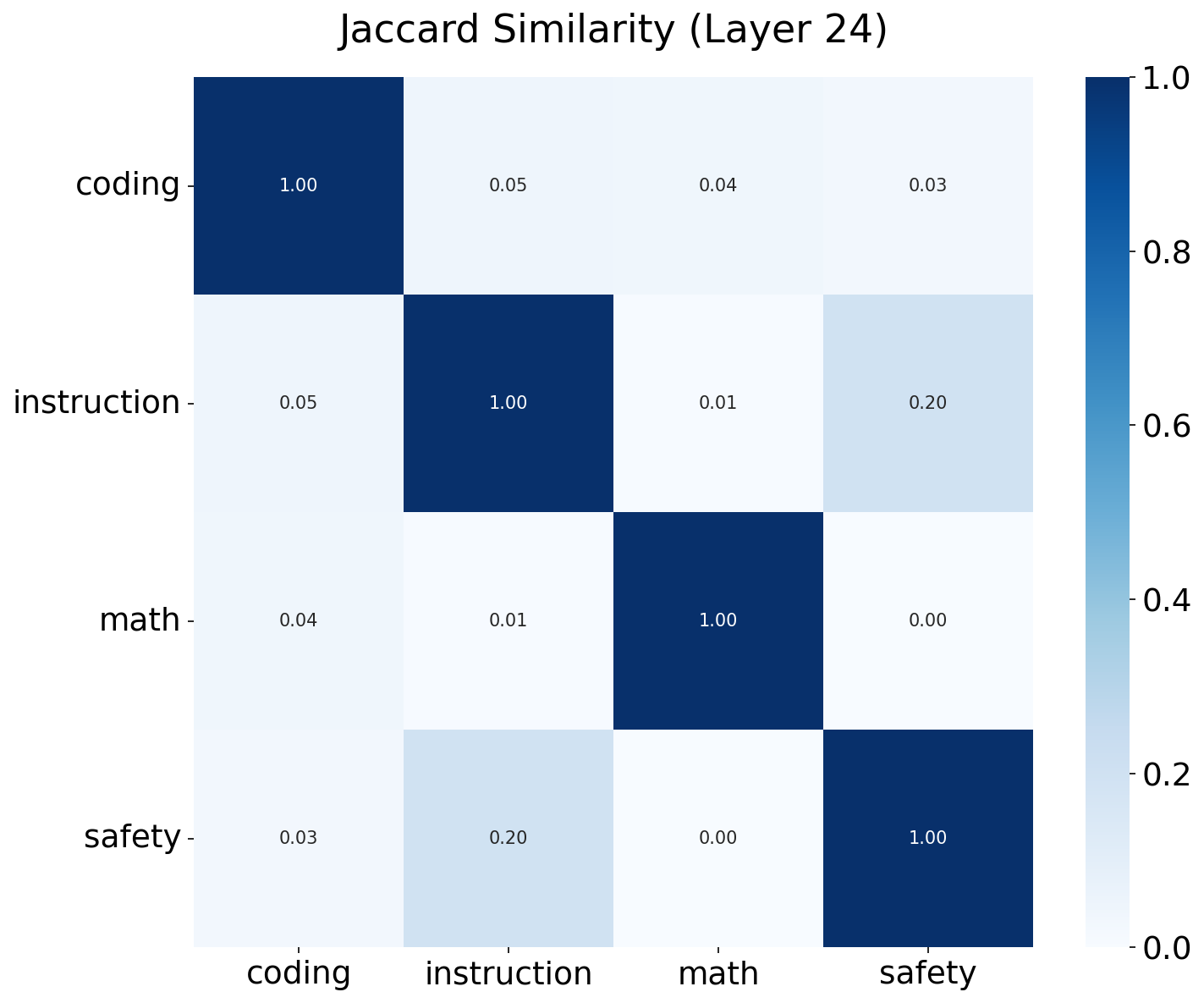} &
        \includegraphics[width=0.235\textwidth]{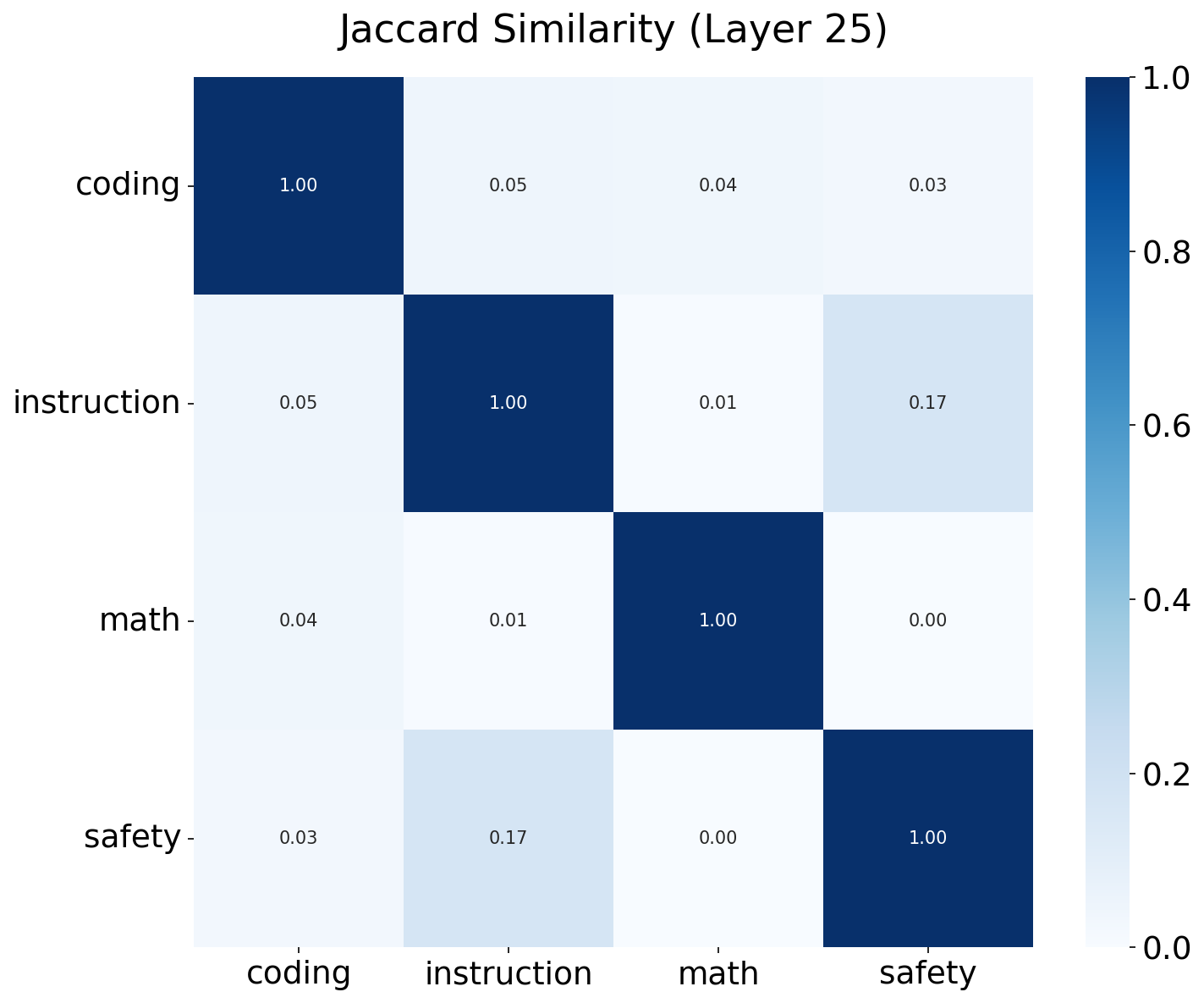} &
        \includegraphics[width=0.235\textwidth]{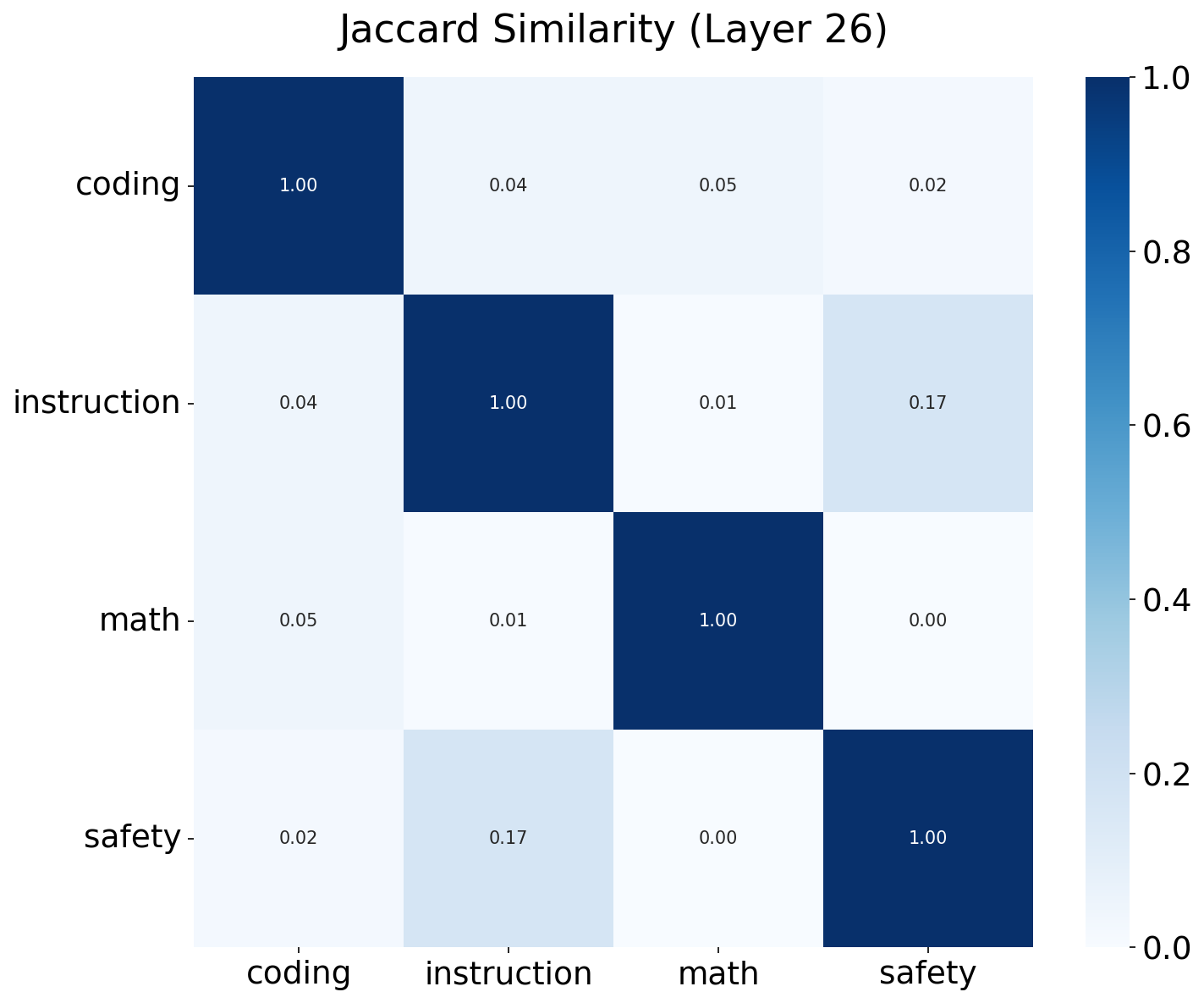} &
        \includegraphics[width=0.235\textwidth]{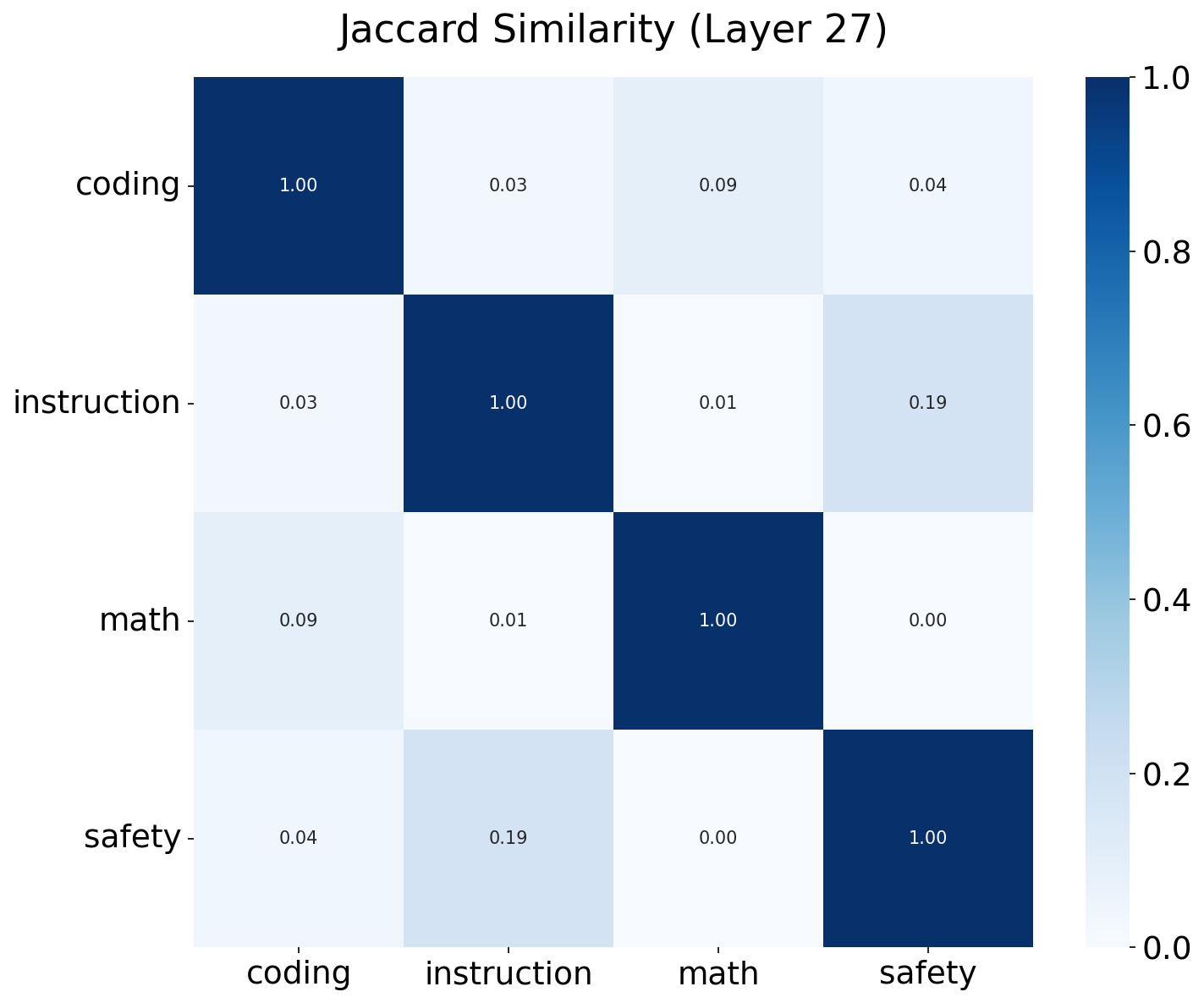}
    \end{tabular}

    \caption{\textbf{Layer-wise Jaccard similarity of active FFN neurons across all 28 layers.}
    Each heatmap shows the pairwise overlap between task-specific active neuron sets in one Transformer layer of Qwen2.5-1.5B-Instruct. The heatmaps are ordered from Layer 0 to Layer 27 in row-major order, from left to right and top to bottom. For each task, we select the top-20\% FFN neurons according to their contribution scores and compute the Jaccard similarity between the selected sets. Darker off-diagonal entries indicate lower cross-task overlap. The consistently low off-diagonal values across many layers suggest that task-relevant FFN activity tends to be concentrated in partially separated neuron subsets, supporting the use of neuron-level structured masks in CASS.}
    \label{fig:app_heatmap_grid_all}
\end{figure*}

Figure~\ref{fig:app_heatmap_grid_all} shows Jaccard heatmaps from shallow to deep layers. The off-diagonal entries are generally low in the selected layers, indicating limited overlap among the most active FFN neurons of different tasks. We do not interpret this as strict disjointness across the entire model; rather, the visualization suggests that task-relevant FFN activity is often concentrated in partially separated neuron subsets.

Overall, the FFN analysis supports the use of neuron-level structured masks in CASS. Since task-relevant neuron contributions are sparse and the most active neurons of different tasks exhibit limited overlap in representative layers, filtering task-vector updates at the neuron level can suppress task-irrelevant updates while preserving the components most associated with each task.

\clearpage
\section{Extended Analysis on MHA}
\label{app:mha_analysis}

In addition to FFN neurons, CASS also constructs structured masks over attention heads. Unlike FFN neurons, many attention heads may serve general-purpose functions and remain active across multiple tasks. Therefore, selecting attention heads solely by absolute contribution magnitude may overemphasize generally active heads rather than task-specialized ones. To address this issue, we analyze attention heads using a task-selectivity metric. Similar patterns are also observed on ViT and RoBERTa. To keep the appendix concise, we focus the detailed breakdown on Qwen2.5-1.5B-Instruct results.

\subsection{Task-Selectivity Metric for Attention Heads}
\label{app:mha_selectivity_metric}

For a head $h$, we define its absolute contribution on task $t$ as
\begin{equation}
    \mathcal{A}_{h,t}
    =
    \mathbb{E}_{\mathbf{X}\sim\mathcal{D}_t}
    \left[
    \left\|
    \mathbf{H}_{h}(\mathbf{X})
    \mathbf{W}_{O}^{(h)}
    \right\|_F
    \right],
    \label{eq:app_head_contribution}
\end{equation}
where $\mathbf{H}_{h}(\mathbf{X})$ denotes the output of head $h$ and $\mathbf{W}_{O}^{(h)}$ denotes the corresponding output-projection block. The Frobenius norm aggregates the contribution over the sequence dimension.

We then compute the global average contribution of head $h$ across all task domains:
\begin{equation}
    \mu_{h,\mathrm{global}}
    =
    \frac{1}{K}
    \sum_{t'=1}^{K}
    \mathcal{A}_{h,t'}.
    \label{eq:app_head_global_mean}
\end{equation}
The task-selectivity score is defined as the relative deviation from this global average:
\begin{equation}
    \mathcal{S}_{h,t}
    =
    \frac{
    \mathcal{A}_{h,t}
    -
    \mu_{h,\mathrm{global}}
    }{
    \mu_{h,\mathrm{global}} + \epsilon
    },
    \label{eq:app_head_selectivity}
\end{equation}
where $\epsilon$ is a small constant for numerical stability.

A positive score $\mathcal{S}_{h,t}>0$ indicates that head $h$ contributes more strongly to task $t$ than its average contribution across tasks. A negative score indicates that the head is less active than its global average under the current task distribution. Therefore, when constructing task-specific head sets for analysis, we rank heads by their positive selectivity scores and select the heads with the largest positive values. We do not use the absolute value of selectivity, since strongly negative scores correspond to suppressed rather than activated task-specific heads.

\subsection{Head-level Sparsity and Cross-task Overlap}
\label{app:mha_overlap}

\begin{figure}[h]
    \centering
    \begin{subfigure}{0.48\textwidth}
        \centering
        \includegraphics[width=0.5\linewidth]{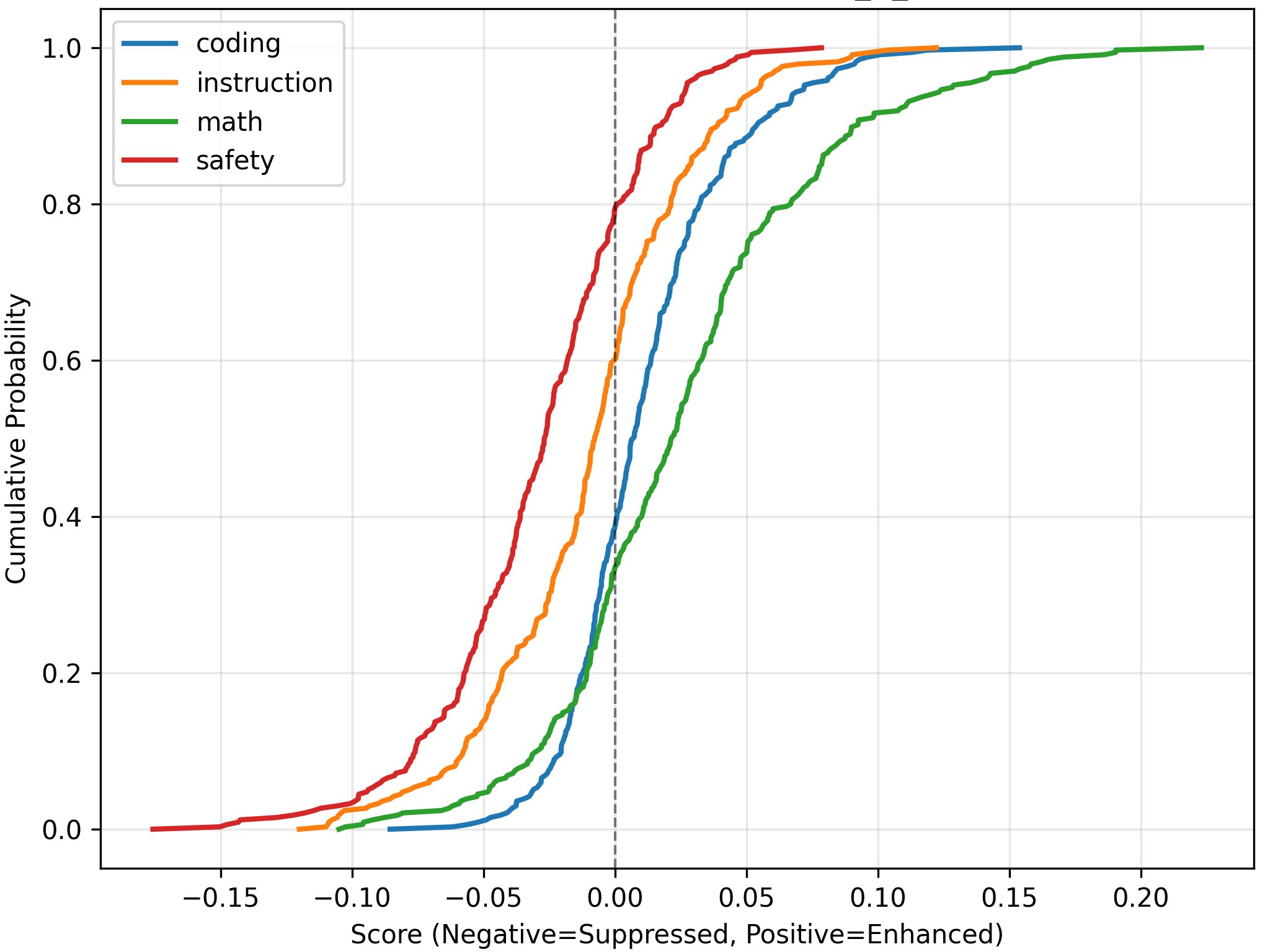}
        \caption{\textbf{CDF of task-selectivity scores.}
        Negative scores indicate heads whose contribution is below their global average for a task, while the positive tail corresponds to task-selective heads.}
        \label{fig:head_cdf}
    \end{subfigure}
    \hfill
    \begin{subfigure}{0.48\textwidth}
        \centering
        \includegraphics[width=0.5\linewidth]{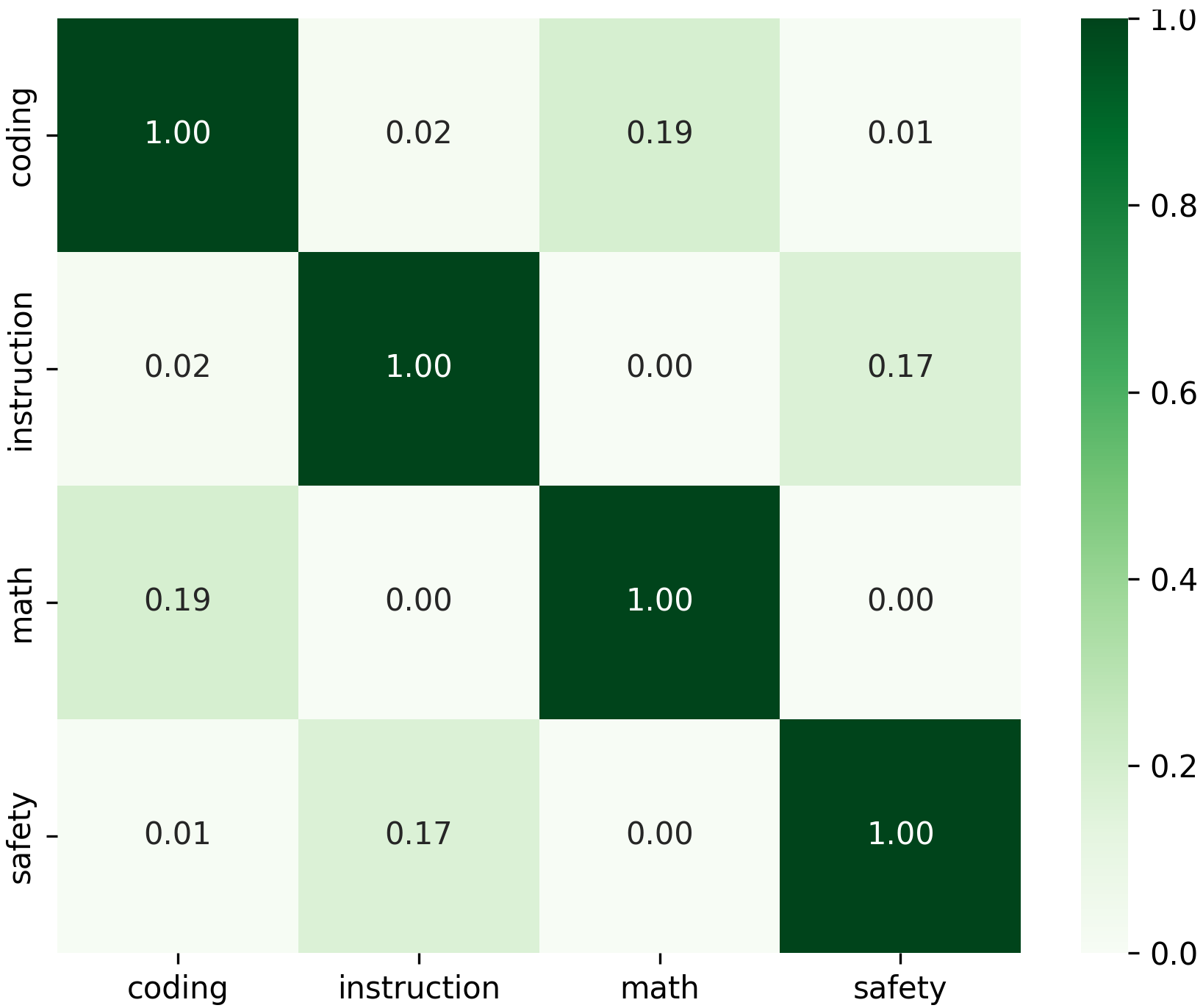}
        \caption{\textbf{Global Jaccard overlap of top-20\% heads.}
        For each task, we select the top-20\% heads with the largest positive selectivity scores and compute cross-task overlap.}
        \label{fig:head_jaccard}
    \end{subfigure}
    \caption{\textbf{Attention head specialization across tasks.}
    Task-selectivity scores reveal that only a subset of heads is strongly specialized for each task, and the selected top heads exhibit limited overlap across different task domains.}
    \label{fig:app_head_stats}
\end{figure}

Figure~\ref{fig:head_cdf} shows the CDF of task-selectivity scores. The presence of negative values indicates that many heads are below their global average contribution for a given task. Meanwhile, the right tail shows that only a subset of heads exhibits strong positive selectivity. This pattern motivates using task-selectivity rather than raw contribution magnitude for attention head selection.

Figure~\ref{fig:head_jaccard} reports the Jaccard similarity between the selected top-20\% task-selective heads for each task. The overlap is low for many task pairs, suggesting that the most task-selective attention heads are not uniformly shared across domains. We also observe a moderate overlap between Coding and Mathematics, which is consistent with the fact that both tasks involve structured reasoning and may rely on partially shared attention mechanisms. We treat this as evidence of partial sharing rather than strict head-level independence.

\subsection{Depth Distribution of Task-selective Heads}
\label{app:mha_depth_distribution}

We further examine where task-selective heads are located across the model depth. For each task, we visualize the layer-wise distribution of the top-20\% heads selected by positive task-selectivity.

\begin{figure}[h]
    \centering
    \includegraphics[width=0.85\linewidth]{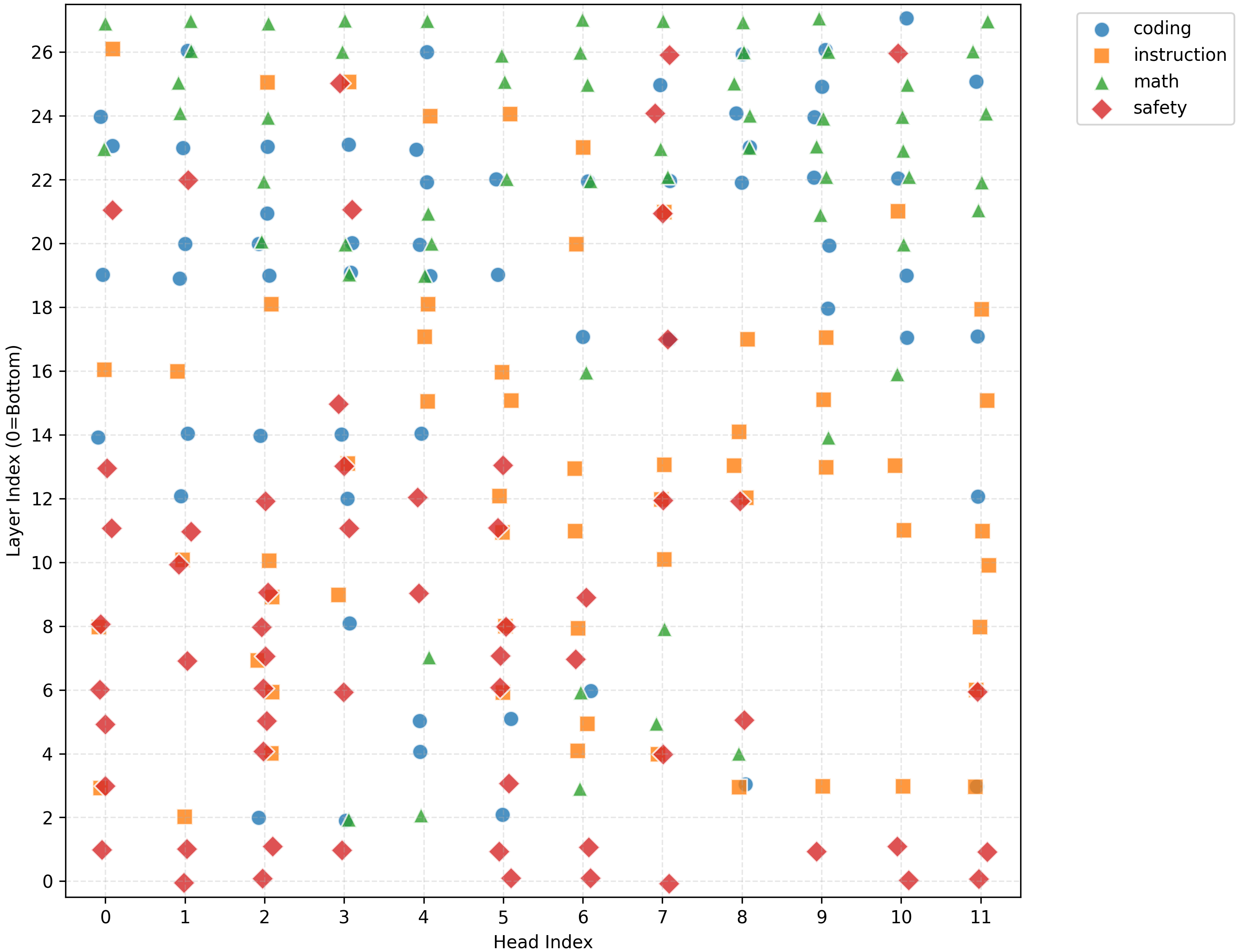}
    \caption{\textbf{Layer-wise distribution of top-20\% task-selective heads.}
    Different tasks exhibit different depth preferences. Coding and Mathematics heads are more concentrated in deeper layers, while Safety heads appear more frequently in relatively shallower layers. Instruction-following heads are distributed more broadly across layers.}
    \label{fig:head_location}
\end{figure}

Figure~\ref{fig:head_location} suggests that task-selective attention heads are not uniformly distributed across depth. Coding and Mathematics show stronger concentration in deeper layers, while Safety-related heads appear more frequently in shallower layers. Instruction-following heads are more broadly distributed. These observations are consistent with the intuition that different tasks may rely on different information-routing patterns, but we avoid interpreting them as strict mechanistic localization.

Overall, the MHA analysis supports the use of task-selectivity-based head masks in CASS. Since many attention heads may be generally active across tasks, raw contribution magnitude alone can overselect generic heads. Positive task-selectivity provides a more targeted criterion for identifying heads whose contributions are relatively amplified under a specific task distribution, enabling CASS to preserve task-relevant head updates while suppressing less relevant ones.

\clearpage
\section{Experimental Settings}
\label{app:settings}

This section describes the experimental settings used in our vision, encoder-only language, and decoder-only language experiments. We summarize the task collections, evaluation protocols, fine-tuning setup, normalization metrics, and random seed configuration.

\subsection{Vision Benchmarks}
\label{app:settings_vision}

For vision experiments, we evaluate CASS on CLIP-based ViT-B/32 and ViT-B/16 models. The benchmark contains 20 image classification tasks:
\begin{quote}
\small
DTD~\cite{cimpoi2014describing}, EuroSAT~\cite{helber2019eurosat}, GTSRB~\cite{stallkamp2011german}, MNIST~\cite{lecun2002gradient}, RESISC45~\cite{cheng2017remote}, Stanford-Cars~\cite{krause20133d}, SUN397~\cite{xiao2016sun}, SVHN~\cite{netzer2011reading}, Oxford-Flowers102~\cite{nilsback2008automated}, PCAM~\cite{veeling2018rotation}, FER2013~\cite{goodfellow2013challenges}, Oxford-IIIT-Pet~\cite{parkhi2012cats}, STL10~\cite{coates2011analysis}, CIFAR100~\cite{krizhevsky2009learning}, CIFAR10~\cite{krizhevsky2009learning}, Food101~\cite{bossard2014food}, Fashion-MNIST~\cite{xiao2017fashion}, EMNIST-Letters~\cite{cohen2017emnist}, KMNIST~\cite{clanuwat2018deep}, and Rendered-SST2~\cite{socher2013recursive}.
\end{quote}

For the main vision experiments, we merge task-specific experts from all 20 tasks and report the average classification accuracy across these tasks following previous study~\cite{gargiulo2025task}.

For the vision OOD experiment, we split the 20 tasks into 12 in-distribution tasks and 8 held-out OOD tasks. The 12 in-distribution tasks are:
\begin{quote}
\small
DTD, EuroSAT, GTSRB, MNIST, RESISC45, Stanford-Cars, SUN397, SVHN, Oxford-Flowers102, PCAM, FER2013, and Oxford-IIIT-Pet.
\end{quote}
The 8 OOD tasks are:
\begin{quote}
\small
STL10, CIFAR100, CIFAR10, Food101, Fashion-MNIST, EMNIST-Letters, KMNIST, and Rendered-SST2.
\end{quote}
In this setting, models are merged using only the 12 in-distribution task experts and are then evaluated on both the 12 in-distribution tasks and the 8 held-out tasks.

\subsection{RoBERTa Benchmarks}
\label{app:settings_roberta}

For encoder-only language experiments, we evaluate RoBERTa on 8 tasks from the GLUE benchmark~\cite{wang2018glue}:
\begin{quote}
\small
CoLA, MNLI, MRPC, QNLI, QQP, RTE, SST-2, and STS-B.
\end{quote}
We report the average task performance over the 8 tasks. Each task is evaluated using its standard benchmark metric~\cite{matena2022merging}. Since these tasks are evaluated under the same benchmark protocol, we report the direct average of task-level scores without additional normalization.

\subsection{Qwen2.5 Benchmarks}
\label{app:settings_qwen}

For decoder-only language experiments, we evaluate Qwen2.5 models on four capabilities: Mathematics, Instruction Following, Coding, and Safety. The datasets and metrics are summarized in Table~\ref{tab:qwen_datasets}.

\begin{table}[h]
    \centering
    \caption{\textbf{Qwen2.5 tasks, datasets, and metrics.}
    We evaluate four decoder-only capabilities covering mathematical reasoning, instruction following, code generation, and safety.}
    \label{tab:qwen_datasets}
    \setlength{\tabcolsep}{4pt}
    \begin{tabular}{l|lll}
        \toprule
        \textbf{Capability} & \textbf{Fine-tuning Dataset} & \textbf{Evaluation Dataset} & \textbf{Metric} \\
        \midrule
        Math & DART-Math~\cite{tong2024dart} & minerva\_math500~\cite{hendrycks2021measuring} & math\_verify accuracy \\
        Instruction & TULU-3 persona IF~\cite{lambert2024tulu} & ifeval~\cite{zhou2023instruction} & prompt-level strict accuracy \\
        Coding & Magicoder~\cite{wei2023magicoder} & mbppplus~\cite{austin2021program} & pass@1 \\
        Safety & WildGuardMix~\cite{han2024wildguard} & wildguardtest~\cite{han2024wildguard} & Micro Harm Rate ($\downarrow$) \\
        \bottomrule
    \end{tabular}
\end{table}

Since the four Qwen2.5 capabilities use different evaluation metrics and scales, we report average normalized performance for Qwen2.5 experiments following previous study~\cite{he2025mergebench}. The normalization protocol is described in Appendix~\ref{app:settings_normalization}.

\subsection{Fine-tuning Hyperparameters}
\label{app:settings_finetuning}

For Qwen2.5 task-specific fine-tuning, we use LoRA-based parameter-efficient fine-tuning. The hyperparameters are summarized in Table~\ref{tab:lora_params}. The same configuration is used across the four decoder-only capabilities unless otherwise specified.

\begin{table}[h]
    \centering
    \caption{\textbf{Fine-tuning hyperparameters for Qwen2.5.}}
    \label{tab:lora_params}
    \setlength{\tabcolsep}{6pt}
    \begin{tabular}{l|c}
        \toprule
        \textbf{Hyperparameter} & \textbf{Value} \\
        \midrule
        Fine-tuning Method & LoRA \\
        LoRA Rank ($r$) & 32 \\
        LoRA Alpha & 64 \\
        LoRA Dropout & 0.1 \\
        \midrule
        Learning Rate & $2 \times 10^{-4}$ \\
        LR Scheduler & Cosine \\
        Warmup Ratio & 0.05 \\
        Epochs & 1 \\
        Batch Size & 16 \\
        \bottomrule
    \end{tabular}
\end{table}

\subsection{Evaluation and Normalization Metrics}
\label{app:settings_normalization}

For vision experiments, we report the average classification accuracy across the evaluated tasks~\cite{gargiulo2025task}. For RoBERTa experiments, we report the average task performance over the 8 GLUE tasks using their standard evaluation metrics~\cite{matena2022merging}.

For Qwen2.5 experiments, the four capabilities use heterogeneous metrics. Mathematics, Instruction Following, and Coding are higher-is-better metrics, while Safety is measured by Micro Harm Rate, where lower is better. To aggregate these capabilities into a single score, we normalize each task score relative to the corresponding single-task fine-tuned expert~\cite{he2025mergebench}.

Let $S_t(\theta)$ denote the raw evaluation score of model $\theta$ on task $t$, and let $\theta_{ft}^{(t)}$ denote the dedicated fine-tuned expert for task $t$. For higher-is-better metrics, the normalized score is defined as
\begin{equation}
    P_t(\theta)
    =
    \frac{
    S_t(\theta)
    }{
    S_t(\theta_{ft}^{(t)})
    }.
    \label{eq:app_norm_higher}
\end{equation}

For the Safety task, we denote the Micro Harm Rate of model $\theta$ by $H_{\mathrm{safety}}(\theta)$. Since lower harm rate indicates better safety performance, we convert it into a higher-is-better normalized score:
\begin{equation}
    P_{\mathrm{safety}}(\theta)
    =
    \frac{
    1 - H_{\mathrm{safety}}(\theta)
    }{
    1 - H_{\mathrm{safety}}(\theta_{ft}^{(\mathrm{safety})})
    }.
    \label{eq:app_norm_safety}
\end{equation}

The final normalized performance for Qwen2.5 is computed as the arithmetic mean over the four capabilities:
\begin{equation}
    P_{\mathrm{avg}}(\theta)
    =
    \frac{1}{4}
    \sum_{t=1}^{4}
    P_t(\theta).
    \label{eq:app_norm_average}
\end{equation}
Under this definition, the dedicated fine-tuned experts have normalized performance equal to 1.0 for their corresponding tasks. This allows models to be compared across heterogeneous capabilities using a unified aggregate score.

\subsection{Random Seeds}
\label{app:settings_seed}

All experiments use a fixed random seed of 42, including mask construction, fine-tuning, and evaluation procedures whenever randomness is involved.

\subsection{Compute Resources.}

All experiments were conducted on NVIDIA GPUs and Ascend NPUs. The computational cost varies substantially across merging baselines. Simple arithmetic or sparsification-based methods such as TA and DARE require only a few seconds per merging run after the single-task checkpoints are available, whereas optimization- or subspace-based methods such as Iso-C may require approximately tens of minutes per run depending on the number of tasks and model size. The additional cost introduced by CASS is small: estimating and applying the CASS mask typically takes only a few seconds per run, and the main overhead comes from collecting activation statistics on a small calibration set. 

\clearpage
\section{Detailed Qwen2.5 Results}
\label{app:qwen_detail}

This section provides detailed per-capability results on Qwen2.5-0.5B-Instruct and Qwen2.5-1.5B-Instruct. While the main text reports average normalized performance, here we further decompose the results into Coding, Safety, Mathematics, and Instruction Following. These results provide a more fine-grained view of how CASS affects different capabilities under both post-hoc merging and mask-guided fine-tuning.

\subsection{Detailed Results of CASS-Merging}
\label{app:qwen_detail_merging}

\begin{table*}[htbp]
\centering
\caption{\textbf{Detailed CASS-Merging results on Qwen2.5.}
Per-capability normalized performance on Qwen2.5-0.5B-Instruct and Qwen2.5-1.5B-Instruct. CASS-M is applied as a post-hoc structured task-vector filter before each merge operator.}
\label{tab:qwen_detail_merging}
\begin{tabular}{l|ccccc|ccccc}
\toprule
\multirow{2.5}{*}{\textbf{Method}} & \multicolumn{5}{c|}{\textbf{Qwen2.5-0.5B-Instruct}} & \multicolumn{5}{c}{\textbf{Qwen2.5-1.5B-Instruct}} \\
\cmidrule(lr){2-6} \cmidrule(lr){7-11}
 & \textbf{Code} & \textbf{Safe} & \textbf{Math} & \textbf{Instr.} & \textbf{Avg.} & \textbf{Code} & \textbf{Safe} & \textbf{Math} & \textbf{Instr.} & \textbf{Avg.} \\
\midrule
Base Model & 0.676 & 0.708 & 0.765 & 0.731 & 0.720 & 0.687 & 0.775 & 0.855 & 0.754 & 0.768 \\
Fine-Tuned & 1.000 & 1.000 & 1.000 & 1.000 & 1.000 & 1.000 & 1.000 & 1.000 & 1.000 & 1.000 \\
Post-Mask & 0.964 & 0.836 & 0.961 & 0.878 & 0.910 & 0.953 & 1.015 & 1.051 & 0.882 & 0.975 \\
\midrule
TA & 0.801 & 0.687 & 0.765 & 0.718 & 0.743 & 0.735 & 0.778 & 0.942 & 0.703 & 0.789 \\
\rowcolor{Gray}
\quad w/ CASS-M & 0.820 & 0.684 & 0.814 & 0.788 & \textbf{0.776} & 0.808 & 0.777 & 1.065 & 0.698 & \textbf{0.837} \\
\cmidrule{1-11}
TIES & 0.671 & 0.659 & 0.873 & 0.577 & 0.695 & 0.829 & 0.752 & 0.978 & 0.651 & 0.803 \\
\rowcolor{Gray}
\quad w/ CASS-M & 0.753 & 0.731 & 0.863 & 0.743 & \textbf{0.773} & 0.846 & 0.767 & 1.094 & 0.610 & \textbf{0.829} \\
\cmidrule{1-11}
DARE & 0.857 & 0.697 & 0.892 & 0.743 & 0.797 & 0.933 & 0.722 & 1.072 & 0.641 & 0.842 \\
\rowcolor{Gray}
\quad w/ CASS-M & 0.903 & 0.684 & 0.853 & 0.788 & \textbf{0.807} & 0.924 & 0.782 & 1.087 & 0.692 & \textbf{0.871} \\
\cmidrule{1-11}
TSV & 0.757 & 0.717 & 0.853 & 0.776 & 0.776 & 0.806 & 0.763 & 1.058 & 0.698 & 0.831 \\
\rowcolor{Gray}
\quad w/ CASS-M & 0.813 & 0.712 & 0.775 & 0.808 & \textbf{0.777} & 0.838 & 0.756 & 1.065 & 0.728 & \textbf{0.847} \\
\cmidrule{1-11}
Iso-C & 0.704 & 0.701 & 0.725 & 0.820 & \textbf{0.738} & 0.743 & 0.782 & 0.935 & 0.774 & 0.809 \\
\rowcolor{Gray}
\quad w/ CASS-M & 0.702 & 0.698 & 0.755 & 0.763 & 0.729 & 0.698 & 0.787 & 1.014 & 0.754 & \textbf{0.814} \\
\cmidrule{1-11}
WUDI & 0.589 & 0.674 & 0.843 & 0.763 & 0.717 & 0.971 & 0.713 & 1.058 & 0.708 & 0.862 \\
\rowcolor{Gray}
\quad w/ CASS-M & 0.832 & 0.692 & 0.843 & 0.737 & \textbf{0.776} & 0.927 & 0.737 & 1.094 & 0.733 & \textbf{0.873} \\
\bottomrule
\end{tabular}
\end{table*}

Table~\ref{tab:qwen_detail_merging} reports the detailed results of CASS-Merging. The ``Post-Mask'' row serves as a diagnostic reference by evaluating task-specific experts after applying CASS masks. Its strong average performance on both Qwen2.5-0.5B-Instruct and Qwen2.5-1.5B-Instruct indicates that the selected components preserve a large fraction of task-specific capability, despite suppressing updates outside the identified active subspace. This supports the premise that task-relevant updates are concentrated in a sparse subset of activation-prominent components.

Notably, on Qwen2.5-1.5B-Instruct, the Post-Mask scores on Safety and Mathematics slightly exceed 1.0. This does not imply that Post-Mask is evaluated against a different reference; rather, the normalized Fine-Tuned score of 1.0 is an empirical single-task baseline rather than a theoretical upper bound. Removing updates outside the selected active subspace can act as a mild denoising or regularization effect, so the masked expert may occasionally outperform the original fine-tuned expert on individual metrics. This observation further suggests that some fine-tuning updates outside the active components can be redundant or mildly detrimental, although Post-Mask is used only as a diagnostic reference rather than as a competing merging method.

Across merge operators, CASS-Merging improves most average scores on both model scales. On Qwen2.5-0.5B-Instruct, the largest gains appear for TIES and WUDI, improving the average normalized performance from 0.695 to 0.773 and from 0.717 to 0.776, respectively. Task Arithmetic also benefits substantially, improving from 0.743 to 0.776. These improvements suggest that structured component-level filtering is particularly effective when the original merge operator does not explicitly distinguish task-relevant functional units from task-irrelevant updates.

On Qwen2.5-1.5B-Instruct, CASS-Merging improves the average performance of all evaluated merge operators. The gains are especially visible for Task Arithmetic, DARE, TIES, and TSV. For example, Task Arithmetic improves from 0.789 to 0.837, while DARE improves from 0.842 to 0.871. The improvements on the larger model indicate that the structured masks remain effective when the model has more capacity and a richer set of functional components.

The per-capability results also show that CASS-Merging does not uniformly improve every individual capability. Some merge operators exhibit trade-offs, where improvements in Mathematics or Coding are accompanied by small decreases in Safety or Instruction Following. This is expected because CASS is applied before a merge operator and does not directly optimize the final multi-task objective. Instead, it filters each task vector according to task-specific component relevance. The consistent average gains indicate that this filtering generally reduces harmful interference, even though individual capabilities may still depend on the downstream merge operator.

The only average-performance degradation occurs for Iso-C on Qwen2.5-0.5B-Instruct, where CASS-Merging changes the average score from 0.738 to 0.729. This suggests that the interaction between structured task-vector filtering and spectrum-based subspace regularization may be sensitive to model scale or hyperparameter choices. In contrast, Iso-C combined with CASS-Merging improves on Qwen2.5-1.5B-Instruct, indicating that this interaction is not uniformly negative but may depend on the capacity and geometry of the underlying model.

Overall, the detailed CASS-Merging results support the conclusion that CASS is broadly complementary to existing merge operators. The benefits are strongest for methods that rely on simple arithmetic, sign conflict resolution, random dropping, or optimization in the original task-vector space, while gains are more moderate for methods that already impose strong subspace or spectrum constraints.

\subsection{Detailed Results of CASS-Tuning}
\label{app:qwen_detail_tuning}

\begin{table*}[htbp]
\centering
\caption{\textbf{Detailed CASS-Tuning results on Qwen2.5.}
Per-capability normalized performance on Qwen2.5-0.5B-Instruct and Qwen2.5-1.5B-Instruct. CASS-T constrains task-specific fine-tuning with structured masks before merging.}
\label{tab:qwen_detail_tuning}
\begin{tabular}{l|ccccc|ccccc}
\toprule
\multirow{2.5}{*}{\textbf{Method}} & \multicolumn{5}{c|}{\textbf{Qwen2.5-0.5B-Instruct}} & \multicolumn{5}{c}{\textbf{Qwen2.5-1.5B-Instruct}} \\
\cmidrule(lr){2-6} \cmidrule(lr){7-11}
 & \textbf{Code} & \textbf{Safe} & \textbf{Math} & \textbf{Instr.} & \textbf{Avg.} & \textbf{Code} & \textbf{Safe} & \textbf{Math} & \textbf{Instr.} & \textbf{Avg.} \\
\midrule
Base Model & 0.676 & 0.708 & 0.765 & 0.731 & 0.720 & 0.687 & 0.775 & 0.855 & 0.754 & 0.768 \\
Fine-Tuned & 1.000 & 1.000 & 1.000 & 1.000 & 1.000 & 1.000 & 1.000 & 1.000 & 1.000 & 1.000 \\
Masked FT & 0.966 & 0.988 & 0.902 & 0.962 & 0.954 & 0.885 & 1.008 & 0.978 & 0.826 & 0.924 \\
\midrule
TA & 0.801 & 0.687 & 0.765 & 0.718 & 0.743 & 0.735 & 0.778 & 0.942 & 0.703 & 0.789 \\
\rowcolor{Gray}
\quad w/ CASS-T & 0.889 & 0.705 & 0.863 & 0.788 & \textbf{0.811} & 0.798 & 0.772 & 1.007 & 0.703 & \textbf{0.820} \\
\cmidrule{1-11}
TIES & 0.671 & 0.659 & 0.873 & 0.577 & 0.695 & 0.829 & 0.752 & 0.978 & 0.651 & 0.803 \\
\rowcolor{Gray}
\quad w/ CASS-T & 0.788 & 0.715 & 1.000 & 0.763 & \textbf{0.816} & 0.969 & 0.748 & 1.073 & 0.677 & \textbf{0.867} \\
\cmidrule{1-11}
DARE & 0.857 & 0.697 & 0.892 & 0.743 & 0.797 & 0.933 & 0.722 & 1.073 & 0.641 & 0.842 \\
\rowcolor{Gray}
\quad w/ CASS-T & 0.853 & 0.690 & 0.922 & 0.776 & \textbf{0.810} & 0.933 & 0.743 & 1.022 & 0.703 & \textbf{0.850} \\
\cmidrule{1-11}
TSV & 0.757 & 0.717 & 0.853 & 0.776 & 0.776 & 0.806 & 0.763 & 1.058 & 0.698 & 0.831 \\
\rowcolor{Gray}
\quad w/ CASS-T & 0.813 & 0.710 & 0.902 & 0.801 & \textbf{0.806} & 0.800 & 0.783 & 1.130 & 0.723 & \textbf{0.859} \\
\cmidrule{1-11}
Iso-C & 0.704 & 0.701 & 0.725 & 0.820 & 0.738 & 0.743 & 0.782 & 0.935 & 0.774 & 0.809 \\
\rowcolor{Gray}
\quad w/ CASS-T & 0.784 & 0.708 & 0.775 & 0.763 & \textbf{0.757} & 0.743 & 0.787 & 1.022 & 0.749 & \textbf{0.825} \\
\cmidrule{1-11}
WUDI & 0.589 & 0.674 & 0.843 & 0.763 & 0.717 & 0.971 & 0.713 & 1.058 & 0.708 & 0.862 \\
\rowcolor{Gray}
\quad w/ CASS-T & 0.901 & 0.695 & 0.794 & 0.769 & \textbf{0.790} & 1.003 & 0.764 & 1.036 & 0.728 & \textbf{0.883} \\
\bottomrule
\end{tabular}
\end{table*}

Table~\ref{tab:qwen_detail_tuning} reports the detailed results of CASS-Tuning. Unlike CASS-Merging, which filters task vectors after conventional fine-tuning, CASS-Tuning applies structured masks during the fine-tuning process itself. The ``Masked FT'' row reports the single-task performance of experts fine-tuned under the CASS masks. The high average scores of 0.954 on Qwen2.5-0.5B-Instruct and 0.924 on Qwen2.5-1.5B-Instruct show that restricting updates to the selected active subspace still preserves most task-specific capability. This confirms that the masked subspace is sufficiently expressive for task adaptation.

CASS-Tuning improves all evaluated merge operators on both Qwen2.5 model scales. On Qwen2.5-0.5B-Instruct, the largest improvement is observed for TIES, which increases from 0.695 to 0.816. WUDI and Task Arithmetic also show strong gains, improving from 0.717 to 0.790 and from 0.743 to 0.811, respectively. These results indicate that constraining fine-tuning to task-relevant components can produce cleaner task vectors before any merge operation is applied.

On Qwen2.5-1.5B-Instruct, CASS-Tuning also consistently improves average performance. TIES improves from 0.803 to 0.867, TSV improves from 0.831 to 0.859, and WUDI improves from 0.862 to 0.883. Compared with the 0.5B model, the gains on the 1.5B model are generally smaller in magnitude but more stable across merge operators. This suggests that larger models may already provide more separable task representations, while CASS-Tuning further regularizes the update subspaces to reduce residual interference.

The per-capability results reveal that the improvements from CASS-Tuning are particularly pronounced for Coding and Mathematics in several merge operators. This pattern is consistent with the intuition that structured reasoning capabilities are sensitive to destructive task-vector interference and can benefit from restricting updates to task-relevant components. Safety and Instruction Following show more mixed per-task changes, but the average results remain consistently positive, indicating that CASS-Tuning improves the overall balance among the four capabilities.

Compared with CASS-Merging, CASS-Tuning provides a stronger form of structural control because it constrains the optimization trajectory rather than filtering task vectors only after fine-tuning. As a result, the generated task vectors are already concentrated in the selected component subspaces before merging. This explains why CASS-Tuning yields consistent average improvements across all evaluated merge operators, including methods such as Iso-C and WUDI, whose interaction with post-hoc filtering can be more method-dependent.

Taken together, the detailed results show that CASS-Merging and CASS-Tuning provide different ways to reduce interference in decoder-only model merging. CASS-Merging acts as a plug-and-play post-hoc filter for existing task vectors, while CASS-Tuning proactively shapes the task vectors during fine-tuning. Both variants improve average normalized performance on Qwen2.5, and CASS-Tuning achieves the most consistent improvements across merge operators.

\clearpage
\section{Limitations}
\label{app:limitations}

We acknowledge two limitations of the current study.

\textbf{First}, CASS uses a fixed sparsity ratio for selecting task-relevant components. Although the adopted ratio (20\% in our experiments) is motivated by the observed concentration of contribution scores, a uniform threshold may not be optimal for all tasks, layers, or model architectures. Tasks with different levels of complexity may require different numbers of active heads and FFN neurons. Developing adaptive selection strategies, such as task-dependent or layer-dependent sparsity ratios, is an important direction for future work.

\textbf{Second}, our decoder-only experiments are conducted on Qwen2.5 models up to 1.5B parameters due to computational resource constraints. Although the consistent improvements across ViT, RoBERTa, and Qwen2.5 suggest that contribution-aware structured sparsity is broadly applicable, validating CASS on larger foundation models, such as 7B-scale or larger decoder-only models, remains future work.

\section{Broader Impacts}
This work aims to improve the efficiency and effectiveness of model merging and task-aware adaptation. Its potential positive impacts include reducing the computational cost of constructing multi-task models, improving the accessibility of model adaptation, and enabling more efficient reuse of existing task-specific models. These benefits may reduce redundant training costs and lower the barrier for researchers and practitioners with limited compute resources. In future work, we plan to further validate the effectiveness and generalizability of our method across a broader range of tasks and application scenarios~\cite{xu2026expected,li2024aves}.

Potential negative impacts may arise from the same efficiency gains. More efficient model merging and adaptation could make it easier to combine or specialize models for harmful applications, or to unintentionally preserve and transfer undesirable biases or unsafe behaviors from task-specific models into a merged model. In addition, if the merged model is deployed without sufficient validation, performance degradation on under-evaluated tasks or safety-sensitive inputs could lead to unreliable behavior. We therefore recommend careful evaluation of merged models, including task-specific robustness and safety checks, before deployment in real-world or high-risk settings.


\end{document}